\pdfoutput=1
\documentclass{article}
\ifdefined\pdfinclusioncopyfonts \pdfinclusioncopyfonts=1 \fi

\usepackage{microtype}
\usepackage{graphicx}
\usepackage{subfigure}
\usepackage{booktabs}
\usepackage{amsmath,amssymb}
\usepackage{multirow}
\usepackage{xcolor}
\usepackage{hyperref}

\usepackage[accepted]{mlsys2025}
\makeatletter
\renewcommand{\printAffiliationsAndNotice}[1]{%
\stepcounter{@affiliationcounter}%
{\let\thefootnote\relax\footnotetext{\hspace*{-\footnotesep}#1%
\forloop{@affilnum}{1}{\value{@affilnum} < \value{@affiliationcounter}}{%
\textsuperscript{\arabic{@affilnum}}\csname @affilname\the@affilnum\endcsname}.
Correspondence to: \mlsyscorrespondingauthor@text.}}}
\makeatother
\hypersetup{pdftitle={TriTile: Three-Tier Attention for Video Diffusion Transformers},
            pdfauthor={Yeonjae Kim, Omin Kwon, Seong Hoon Seo, Jae W. Lee}, pdfsubject={}}

\usepackage{xspace}
\usepackage{pifont}
\usepackage{xurl}  %

\newcommand{\sys}{\textsc{TriTile}\xspace}

\newif\ifcomments
\commentsfalse

\newcommand{\newauthortag}[3]{%
  \ifcomments
    \newcommand{#1}[1]{{\color{#3}\textbf{[#2:}~##1\textbf{]}}}%
  \else
    \newcommand{#1}[1]{}%
  \fi}

\newauthortag{\yj}{YJ}{blue}
\newauthortag{\jwl}{JWL}{red}
\newauthortag{\todo}{TODO}{orange}
\newauthortag{\fixme}{FIXME}{magenta}

\ifcomments
\else

\fi
\newauthortag{\ask}{ASK}{red}   %

\ifcomments
  
\else
  
\fi

\definecolor{chgpurple}{RGB}{133,0,200}
\ifcomments
  \newcommand{\chgnote}[1]{{\color{chgpurple}\footnotesize\itshape[#1]}}
\else
  
  \newcommand{\chgnote}[1]{}
\fi
\newcommand{\chgdel}[1]{}  %
\ifcomments
  
\else
  
\fi
\usepackage[normalem]{ulem}
\ifcomments

\else

\fi
\ifcomments
  \DeclareRobustCommand{\chgold}[1]{{\color{gray}\sout{#1}}}
\else
  \DeclareRobustCommand{\chgold}[1]{\ignorespaces}
\fi
\ifcomments
  
  \DeclareRobustCommand{\rchgold}[1]{{\color{chgpurple}\sout{#1}}}
\else
  
  \DeclareRobustCommand{\rchgold}[1]{\ignorespaces}
\fi
\newcommand{\Nt}{N_{\mathrm{t}}}  %

\newcommand{\x}{$\times$\xspace}
\newcommand{\figref}[1]{Figure~\ref{#1}}
\newcommand{\tabref}[1]{Table~\ref{#1}}
\newcommand{\secref}[1]{\S\ref{#1}}
\newcommand{\algref}[1]{Algorithm~\ref{#1}}
\newcommand{\appref}[1]{Appendix~\ref{#1}}

\usepackage{pgfplots}
\pgfplotsset{compat=1.18}
\usepgfplotslibrary{groupplots}
\usetikzlibrary{calc,patterns}
\definecolor{cFP}{HTML}{36597E}    %
\definecolor{cSP}{HTML}{0F907B}    %
\definecolor{cRef}{HTML}{9A9EA8}   %
\definecolor{cGrid}{HTML}{E4E4E2}
\pgfplotsset{
  tritile/.style={
    font=\scriptsize,
    width=0.48\columnwidth, height=3.3cm,
    scale only axis=false,
    axis line style={black!60, line width=0.4pt},
    tick style={black!60, line width=0.4pt},
    major tick length=2pt,
    tick label style={font=\footnotesize},
    label style={font=\footnotesize},
    title style={font=\footnotesize\bfseries, yshift=-1.5pt},
    ymajorgrids, grid style={cGrid, line width=0.4pt},
    axis x line*=bottom, axis y line*=left,
    every axis plot/.append style={line width=1pt, mark size=1.6pt},
    legend style={draw=none, fill=none, font=\scriptsize, column sep=3pt},
    legend cell align=left,
    y tick label style={/pgf/number format/fixed, /pgf/number format/precision=2},
    scaled y ticks=false,
  },
}

\newif\ifwithappendix
\withappendixtrue       %
\graphicspath{{figures/}}

\mlsystitlerunning{TriTile: Three-Tier Attention for Video Diffusion Transformers}

\begin{document}

\twocolumn[
\mlsystitle{TriTile: Three-Tier Attention for Video Diffusion Transformers}

\begin{mlsysauthorlist}
\mlsysauthor{Yeonjae Kim}{snu}
\mlsysauthor{Omin Kwon}{snu}
\mlsysauthor{Seong Hoon Seo}{snu}
\mlsysauthor{Jae W. Lee}{snu}
\end{mlsysauthorlist}

\mlsysaffiliation{snu}{Department of Computer Science and Engineering, Seoul National University, Seoul, Republic of Korea}
\mlsyscorrespondingauthor{Jae W. Lee}{jaewlee@snu.ac.kr}

\mlsyskeywords{Video Diffusion Transformers, Attention, Quantization, Sparsity, Efficient Inference}

\vskip 0.3in

\begin{abstract}
Attention over tens of thousands of visual tokens dominates the inference of video diffusion transformers (DiTs).
Existing methods reduce this cost through sparsity, low precision, or both, but use a single precision for all kept tiles.
We observe that the output error introduced by approximation varies widely across tiles.
Many tiles can be skipped with little error, others must be computed but gain little beyond low precision, and a few retain substantial error even at low precision.
We present \sys, a training-free attention method that assigns each tile one of three approximation tiers: \emph{skipping} it by replacing it with its block means, \emph{skimming} it in NVFP4, or \emph{refining} it with NVFP4 residual correction.
A lightweight router predicts each tile’s error under each tier from block statistics and spends extra compute on a tile only where it removes enough error; a single knob, a price on compute, controls the whole speed--quality trade-off.
A fused kernel executes all three tiers in a single pass on the FP4 tensor cores. %
Over dense FlashAttention, \sys speeds up generation by 1.38--1.83\x on Wan2.2 at 23.0--27.6\,dB PSNR, 2.28--2.32\x on HunyuanVideo at 34.5--36.0\,dB, and 1.84\x on LTX-2.5 at 34.1\,dB, and in every setting it Pareto-dominates every baseline in speed and PSNR.

\end{abstract}
]

\printAffiliationsAndNotice{}

\section{Introduction}
\label{sec:intro}

Diffusion transformers (DiTs)~\cite{dit} have become the backbone of video generation~\cite{hunyuanvideo,wan,ltx} and world simulation~\cite{cosmos1,cosmos3}.
Recent video DiTs apply 3D full attention, in which every visual token attends to every other across all frames~\cite{cogvideox,hunyuanvideo,wan}, and a video yields tens of thousands of these tokens.
This attention has quadratic cost in the number of tokens and accounts for a large fraction of inference time, a share that grows as models move to higher resolutions and longer videos.

\begin{figure}[t]
  \centering
  \includegraphics[width=\columnwidth]{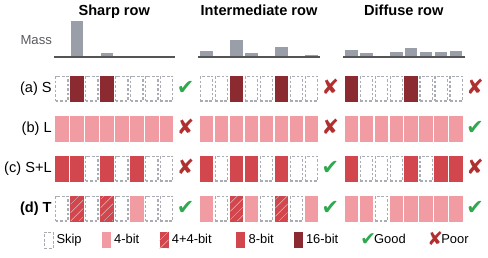}  %
  \caption{Per-tile precision on sharp, intermediate, and diffuse rows, each with one query block and eight tiles, under the same compute budget. (a)~Sparse (S), (b)~low-bit (L), and (c)~combined (S+L) attention favor different attention distributions; (d)~\sys (T) adapts to each distribution by skipping, skimming, or refining individual tiles.}
  \label{fig:policy}
\end{figure}

\begin{figure*}[t]
  \centering
  \includegraphics[width=\textwidth]{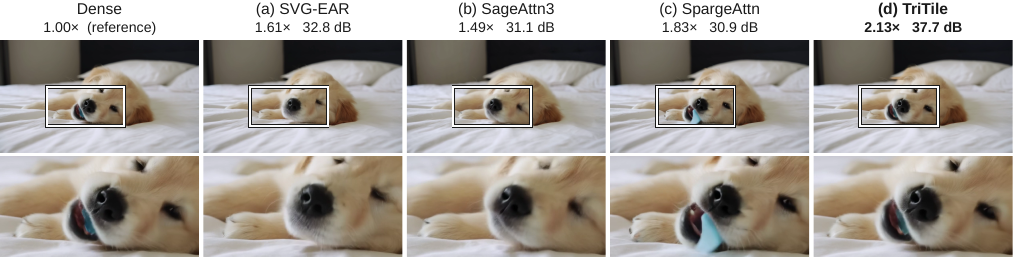}
  \caption{The 21st of 129 frames of the video that HunyuanVideo generates at 720p for the prompt ``On the bed in the bedroom, a puppy is playing with a slipper in its mouth.'' The bottom row zooms into the box; labels give the end-to-end speedup over dense and the PSNR against the dense video. (a)~SVG-EAR~\cite{svg-ear}, (b)~SageAttention3~\cite{sageattention3}, and (c)~SpargeAttn~\cite{sparge1} follow the policies of \figref{fig:policy}a--c; (d)~\sys.}
  \label{fig:intro-qual}
\end{figure*}

Sparse attention and low-bit attention are the two main approaches to reducing this cost.
Both operate on \emph{tiles}, each pairing a block of queries with a block of keys.
Sparse attention selects the tiles to compute and skips the rest, dropping them~\cite{sparge1,svg2} or approximating each by its mean key and value~\cite{sol-attn,svg-ear} (\figref{fig:policy}a).
Low-bit attention instead computes every tile in a low-bit format such as FP8 or NVFP4~\cite{sageattention2,sageattention3} (\figref{fig:policy}b).
Methods that combine the two~\cite{sparge1,paroattention} skip tiles and compute the kept ones in a low-bit format (\figref{fig:policy}c).

These approaches give a tile at most two options, skipping it or computing it at a single precision shared by every kept tile.
Which approach is most accurate depends on the \emph{row}, the tiles that one query block attends to (\figref{fig:policy}).
A row is \emph{sharp} when its query block puts almost all of its attention on a few tiles, and \emph{diffuse} when it spreads its attention over many.
At the same compute cost, sparse attention tends to be the most accurate on sharp rows, low-bit attention on diffuse rows, and their combination on rows in between.

Our key observation is that the output error from skipping a tile or computing it at low precision varies widely across tiles, even within the same row.
Because this error differs so much from tile to tile, extra computation should be judged by its \emph{marginal gain}, the output error it removes per unit of additional compute.
Existing approaches, which share one precision across all computed tiles, cannot fully exploit these differences: they spend more compute than some tiles need while leaving substantial error on others that would benefit from additional computation.

We present \sys, which assigns each tile one of three approximation tiers, namely \emph{skip} (block means), \emph{skim} (NVFP4), and \emph{refine} (NVFP4 with residual correction) (\figref{fig:policy}d).
Before each attention call, a \emph{router} estimates each tile's output error under each tier from block statistics and upgrades a tile only when its marginal gain is at least a price $\pi$.
Under our error model, this rule reduces to comparing one score per tile against two thresholds, making the decision inexpensive and leaving $\pi$ as the single user-controlled parameter for the speed--quality trade-off.
A periodically rebuilt clustering tree groups similar queries and keys into blocks, keeping the block statistics representative and the cheaper approximations accurate.

\sys then executes all three tiers in one fused kernel and refines selected tiles with additional NVFP4 residual products, which give nearly the same error as using BF16 while keeping every tier on the FP4 tensor cores.
Skipped tiles join the same online softmax as virtual tiles of their block means.
In \figref{fig:intro-qual}d, \sys is both faster and more faithful to the dense video than all three methods, and across Wan2.2, HunyuanVideo, and LTX-2.5 it accelerates each attention call by up to 8.56\x and generation by up to 2.32\x over FlashAttention.

In summary, this paper makes the following contributions:
\begin{itemize}
  \item We show that existing approaches leave accuracy unexploited because approximation error varies widely across tiles, even within a row, motivating a per-tile choice by marginal gain (\secref{sec:motivation}).
  \item We propose three approximation tiers with a router that estimates each tile's error under each tier and trades it against compute, and a clustering tree that groups similar tokens into blocks, lowering the error of skipping and skimming (\secref{sec:design}).
  \item We implement all three tiers in one fused kernel on FP4 tensor cores, representing skipped tiles as virtual tiles of block means and refining selected tiles with NVFP4 residual products (\secref{sec:impl}).
  \item We evaluate \sys on four video DiTs in five settings and show that it achieves a better trade-off between speed and quality than all evaluated baselines (\secref{sec:eval}).
\end{itemize}

\section{Background}
\label{sec:background}

\subsection{Video Diffusion Transformers}
\label{sec:bg-dit}
Video diffusion transformers (DiTs) iteratively denoise latent representations, which encode videos compressed spatially and temporally by a 3D VAE.
These latents are divided into patches that serve as visual tokens.
Recent video DiTs apply 3D full attention, in which every token attends to every other across all frames~\cite{cogvideox,hunyuanvideo,wan}.
This improves quality but makes attention cost quadratic in the total number of visual tokens.

\subsection{Block-Sparse Attention}
\label{sec:bg-sparse}
Attention kernels such as FlashAttention~\cite{flashattention,flashattention2,flashinfer} partition queries, keys, and values into blocks and process one \emph{tile} at a time, where a tile pairs a query block with a key block.
An online softmax merges tiles one at a time, so a kernel can skip a tile or compute it differently without changing the others.
We call the set of tiles attended to by one query block a \emph{row}.
For a row with query block $\mathcal{I}$ and key blocks $\mathcal{K}_1, \dots, \mathcal{K}_{\Nt}$, the \emph{mass} of tile $t$ is the share of the row's attention on it,
\begin{equation}
  m_t = \frac{1}{|\mathcal{I}|} \sum_{i \in \mathcal{I}} \sum_{j \in \mathcal{K}_t} P_{ij},
  \label{eq:mass}
\end{equation}
where $P$ is the softmax attention matrix, so a row's masses sum to one.
Block-sparse attention exploits the concentration of attention mass on a few keys~\cite{svg1,sparge1,xattention,svg2}: since each tile's contribution to the output is weighted by its mass, it keeps the tiles with the largest estimated mass.
Because each query block's output depends only on its own row, these methods select tiles separately for each row.
Each row computes either a fixed number of tiles or enough tiles to cover a target share of its mass.

Some methods fine-tune models to use sparse attention~\cite{vsa,sla,sparge2,fpsattn}, but the additional training cost for each model motivates our focus on training-free methods.
Recent training-free methods use three techniques.
First, they select tiles dynamically for each input~\cite{sparge1,svg2,adaspa,sol-attn}, since static masks~\cite{svg1,sta,radial-attention} such as the sliding windows of STA cannot adapt to attention patterns that vary across prompts, layers, and denoising steps.
To avoid computing all attention scores, most of these methods estimate tile masses using block averages of queries and keys.
Second, token clustering groups similar tokens into blocks to improve the accuracy of these estimates~\cite{svg2,svg-ear,svoo,hypervattention}, a reordering that 3D full attention permits.
Applying $k$-means to each input adapts the grouping to its tokens, unlike fixed arrangements such as 3D tiling~\cite{sta}, a Hilbert curve~\cite{sparge1}, or per-head axis permutations~\cite{paroattention}.
Third, mean-based approximations represent skipped tiles by their block means rather than discarding their contributions entirely~\cite{sol-attn,svg-ear,pisa}.

\subsection{Low-Bit Attention and NVFP4}
\label{sec:bg-lowbit}
Low-bit attention reduces the cost of each computed tile by quantizing queries, keys, values, and probabilities for low-precision tensor-core operations.
NVFP4 represents each value in the 4-bit E2M1 format~\cite{nvfp4}.
Each group of 16 consecutive values shares an E4M3 scale, and each tensor an FP32 scale.
On Blackwell GPUs, the FP4 tensor cores run 4\x as fast as BF16 ones with FP32 accumulation on the RTX PRO 6000, and 8\x on the RTX 5090~\cite{nvidia-blackwell,nvidia-rtx-pro-blackwell}.

Outliers can inflate quantization scales and increase rounding error.
SageAttention2 mitigates this through query and key smoothing, which SageAttention3 also adopts~\cite{sageattention2,sageattention3}.
We describe this computation below using the NVFP4 format of SageAttention3.
Let $\bar{q}$ be the mean query of a query block and $\bar{k}$ the mean key of a head, and let $\hat{Q} = Q - \bar{q}$ and $\hat{K} = K - \bar{k}$ be the \emph{smoothed operands}.
Subtracting $\bar{k}$ shifts each query's attention scores by a constant, leaving the softmax unchanged.
Centering the queries additionally requires a correction term.
Omitting the scaling factor, we approximate the smoothed scores as
\begin{equation}
  S \approx \mathcal{Q}(\hat{Q})\,\mathcal{Q}(\hat{K})^\top + C,
  \qquad C = \bar{q}\,\hat{K}^\top,
  \label{eq:smooth}
\end{equation}
where $\mathcal{Q}(X)$ denotes the scaled values represented by NVFP4 quantization.
The correction $C$ is computed in full precision.
The value product also uses quantized probabilities and values, while accumulation and the online softmax remain in FP32.

\subsection{Orthogonal Acceleration}
\label{sec:bg-orthogonal}
Video DiTs are also accelerated outside attention: quantizing the linear layers~\cite{ptq4dit,vidit-q,svdquant} speeds up the rest of each block, step distillation~\cite{consistency-models,dmd,rcm} reduces the number of denoising steps, and feature caching~\cite{pab,teacache,taylorseer} reuses block outputs across steps. 
Each computed step still runs full attention, so these methods are orthogonal to attention acceleration.

\section{Motivation}
\label{sec:motivation}
Existing sparse, low-bit, and combined attention methods typically compute all retained tiles at a single precision.
We analyze how approximation error varies, first across rows and then among tiles within a row, to identify opportunities for choosing sparsity and precision per tile.

\paragraph{Methodology.}
We capture queries, keys, and values from dense runs of Wan2.2-TI2V-5B and HunyuanVideo, using four prompts and four denoising steps, and sampling 10 and 15 evenly spaced layers, respectively (160 and 240 attention calls).
We measure the relative L2 error of each row's output against dense attention.
We compare three allocations at 25\% of the modeled cost of dense BF16 attention, assuming for simplicity that the cost of an operation scales with its bit width, so that 8-bit and 4-bit operations cost half and a quarter as much as BF16 ones.
Sparse attention (S) computes the heaviest 25\% of each row's tiles in BF16.
Combined attention (S+L) computes the heaviest 50\% using the INT8 query--key and FP8 probability--value products of SageAttention2~\cite{sageattention2}.
Low-bit attention (L) computes all tiles in NVFP4 using SageAttention3~\cite{sageattention3}.
S and S+L select tiles by exact attention mass from dense attention and drop the rest, isolating allocation effects from mass-estimation error.

\subsection{Different Approximations Across Rows}
\label{sec:motiv-complementary}

\setcounter{topnumber}{1}
\begin{figure}[t]
  \centering
  \includegraphics[width=\columnwidth]{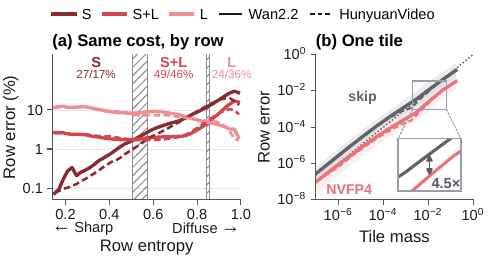}
  \caption{Approximation error across rows and tiles. (a)~Median row error versus row entropy at equal modeled compute; percentages: share of rows on which each approach is most accurate (Wan2.2/HunyuanVideo). (b)~Row error from approximating a single tile versus its attention mass.}
  \label{fig:motiv-entropy}
\end{figure}

At a fixed compute cost, which of S, S+L, and L achieves the lowest error depends on the row's attention distribution.
We summarize this distribution by the normalized entropy of its tile masses, $H = -\sum_t m_t \ln m_t / \ln \Nt$.
$H$ is 0 when one tile holds all the mass (a sharp row) and 1 when all tiles hold equal mass (a diffuse row).
\figref{fig:motiv-entropy}a plots the median error of each approach against $H$.

\paragraph{Each approach wins on different rows.}
S is most accurate on sharp rows, S+L on intermediate rows, and L on diffuse rows.
These allocations trade the error from omitting tiles against the error from computing retained tiles at lower precision.
To explain why the balance between these errors changes with the row's attention distribution, we examine how approximating one tile changes the row's output.
\begin{figure}[t]
  \centering
  \includegraphics[width=\columnwidth]{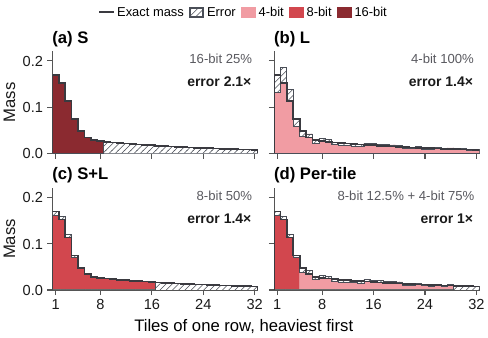}  %
  \caption{Illustrative allocations at equal compute. (a)--(c)~S, L, and S+L use one precision for kept tiles. (d)~A per-tile allocation reserves more accurate computation for heavy tiles. Hatched areas show modeled errors; labels give their sum relative to (d).}
  \label{fig:motiv-row}
\end{figure}
\paragraph{Error scales with tile mass.}

We approximate one tile at a time, keeping all other tiles exact, and measure the row's relative L2 error.
The tile is either replaced by its block-mean key and value or computed in NVFP4.
\figref{fig:motiv-entropy}b shows that both errors grow roughly in proportion to tile mass in each model, with NVFP4 at about a fifth to a quarter of the block-mean error.

\begin{figure*}[!t]
  \centering
  \includegraphics[width=\textwidth]{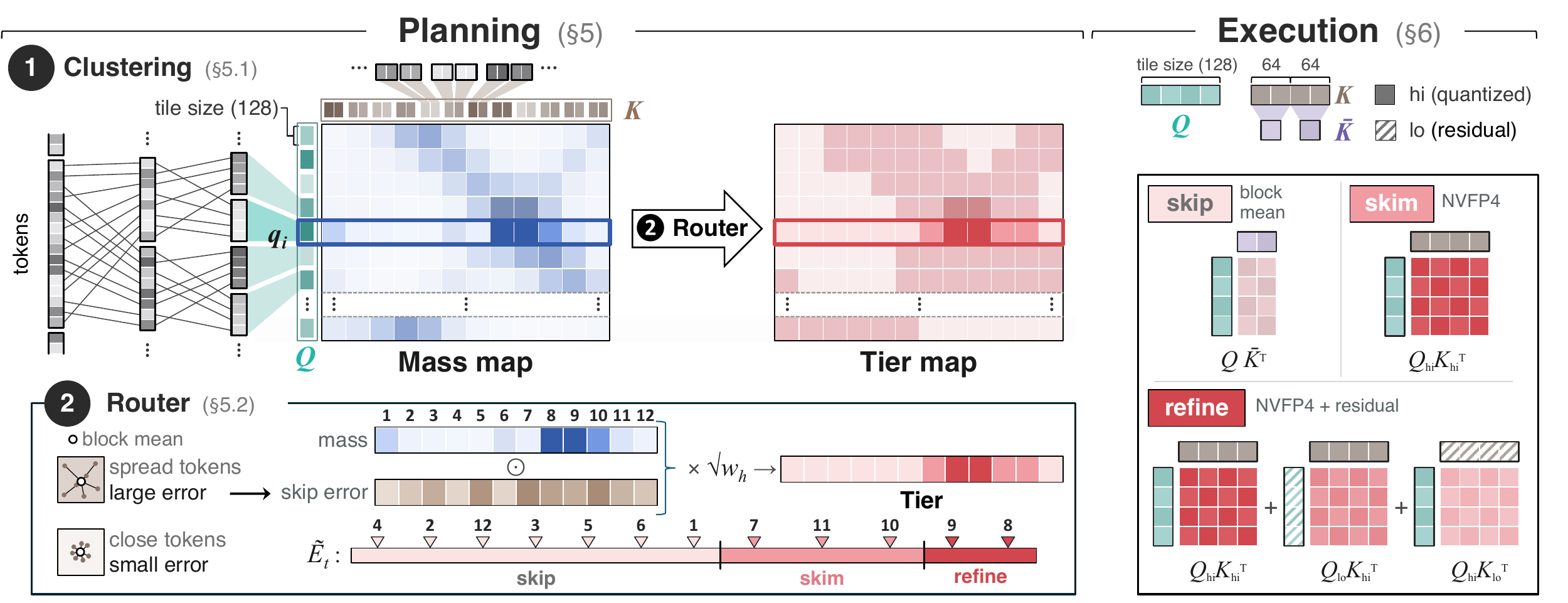}
  \caption{Overview of \sys. Planning: \ding{202}~clustering trees group similar tokens into blocks; \ding{203}~the router picks a tier for each tile. Execution: one pass runs all three tiers on the FP4 tensor cores.}
  \label{fig:overview}
\end{figure*}

When one tile holds almost all a row's mass, skipping the rest can introduce little error, while NVFP4 rounding error on that tile can remain large.
In diffuse rows, skipping loses more mass, favoring more tiles at lower precision.
This helps explain the different winners in \figref{fig:motiv-entropy}a.

\subsection{Different Precisions Within a Row}
\label{sec:motiv-concentration}

Within a row, increasing precision costs about the same across all equal-sized tiles, but can remove more output error from heavy tiles, those with large attention mass.
The mass--error relationship therefore motivates choosing precision per tile by its marginal gain, the output error removed per unit of additional compute.

A row can contain a few heavy tiles with large NVFP4 error and many moderate tiles that together hold too much mass to skip with little error.
Using NVFP4 for all kept tiles leaves error on the heavy ones, while raising precision for all also increases the cost of keeping the moderate tiles.

\figref{fig:motiv-row} illustrates this trade-off.
S leaves error on skipped tiles (a), and L leaves rounding error on heavy tiles (b).
S+L reduces rounding error on every kept tile but must skip more tiles at the same compute (c). A per-tile allocation instead matches the computation of each tile to its mass (d), reducing the summed modeled error at equal compute.

\section{Overview}
\label{sec:overview}
\setcounter{topnumber}{2}  %

\sys chooses one of three tiers for each tile.
\emph{Skip} replaces a light tile with the means of its keys and values, \emph{skim} computes a moderate tile in NVFP4, and \emph{refine} computes a heavy tile at higher precision.
Making this choice by marginal gain sets two goals.
The first is to predict the error of each tier with low overhead, without the full score matrix.
Exact masses require the scores that skipping avoids, and the tokens within a tile vary, so cheap summaries such as block means can miss both a tile's mass and its approximation error.
The second is to run the chosen tiers without losing the speed of low-bit computation.
Running the tiers in separate kernels requires merging their outputs, and higher precision usually comes from BF16, which runs at a fraction of the FP4 throughput.
\figref{fig:overview} shows how the two stages of \sys meet these goals.

\paragraph{Planning.}
Planning~(\secref{sec:design}) first groups similar queries and keys into fixed-size blocks with a clustering tree~\ding{202}~(\secref{sec:design-cluster}) that repeatedly splits tokens along their principal direction, reducing the variation within each tile and thus the approximation error.
It then collects block statistics to estimate the \emph{mass} of each tile and its \emph{skip error}. 
The skip error is the error that replacing the tile by its block mean would leave and grows with the spread of its tokens. 
The router~\ding{203}~(\secref{sec:design-router}) multiplies the two and weights the product by the head's relative output energy $\sqrt{w_h}$, since heads with larger outputs contribute more to the error of the layer, to obtain one score $\tilde E_t$ per tile. 
Two thresholds on $\tilde E_t$ then pick skip, skim, or refine.
Statistics collection and routing are fused with smoothing and quantization into a few preprocessing kernels that run at every call, while the tree is rebuilt periodically, keeping the planning overhead small.
Planning outputs a tier map together with the block means and quantized operands that execution needs.

\paragraph{Execution.}
Execution (\secref{sec:impl}) combines all three tiers within one online softmax on the FP4 tensor cores.
Skipped tiles contribute through their block means, and skimmed tiles use one NVFP4 product of the \emph{hi} parts.
Refined tiles add the products with the \emph{lo} parts, the NVFP4 quantization residuals, which nearly match the accuracy of BF16.

\section{Planning: Choosing Tiers}
\label{sec:design}

\subsection{Tile-Aligned Clustering}
\label{sec:design-cluster}

\sys clusters similar queries and keys to improve both routing estimates and approximate computation.
Grouping dissimilar tokens makes block means less representative and can increase errors from mean replacement and shared quantization scales.
Unlike fixed token orders, $k$-means~\cite{svg2,svg-ear} adapts to the input, but it produces clusters of varying size, which fixed-size tiles must pad or split.
\sys therefore uses a divisive clustering tree that groups similar tokens while controlling block size.

\sys builds separate query and key trees for each head using a size-balanced variant of principal direction divisive partitioning (PDDP)~\cite{pddp}.
Each node represents a group of tokens, and the root contains all queries or all keys of a head.
At each node, \sys estimates the principal direction and sorts the tokens by their projections onto it.
It then splits the sorted tokens near the midpoint, at a multiple of the target block size.
Splitting continues until each leaf holds one block, and only the final block may be partially filled.
The trees are rebuilt every five steps.
\appref{sec:app-tree} gives the full algorithm, including dimensionality reduction and principal-direction estimation.

The attention kernel processes tiles of 128 queries by 128 keys.
Query clusters contain 128 tokens, while key clusters contain 64 so that routing can use statistics from smaller key groups.
The router combines the statistics of two consecutive key clusters to assign one tier to each 128-key tile.
A skipped tile is represented by the mean keys and mean values of its two clusters.

\subsection{Router}
\label{sec:design-router}
The router assigns each tile a tier $k_t \in \{\text{skip}, \text{skim}, \text{refine}\}$ by its marginal gain, which requires predicting the error and compute of each tier.
Predicting the error exactly is costly, because approximating one tile changes the effect of the others through the shared softmax sum of a row, and evaluating every combination of tiers grows exponentially with the number of tiles.
Under a first-order approximation in the change of this sum, however, approximating one tile does not change the effect of the others, so the router uses the sum of the errors of approximating each tile alone as a surrogate for the row's error.
The compute of a tile also depends only on its own tier, so the router can choose tiers tile by tile.
Let $E_t(k)$ be the error of approximating tile $t$ alone with tier $k$, relative to the row's output, and let $c(k)$ be the compute of tier $k$ relative to skimming: 0 for skip, 1 for skim, and $c_{\mathrm{rf}}$ for refine. The router minimizes
\begin{equation}
  \sum_t \Big( E_t(k_t) + \pi\, \frac{c(k_t)}{\Nt} \Big),
  \label{eq:objective}
\end{equation}
where dividing by the row's tile count $\Nt$ keeps the price $\pi$ comparable across sequence lengths, and $\pi$ is the smallest marginal gain that justifies extra compute.

\paragraph{Skipping and skimming errors.}
Skipping replaces the keys and values of each of the tile's two 64-key clusters with their means, which changes the tile's contribution in two ways.
First, because the exponential is convex, a mean key contributes less to the softmax normalizer than the keys it replaces, so the tile's contribution to it shrinks by a fraction $J_t$.
Second, a mean value weights all values equally, whereas attention favors the values of high-scoring keys, so the tile's value shifts by a relative amount $R_t$.
A first-order model of these two effects gives
\begin{equation}
  E_t(\text{skip}) \approx m_t(J_t + R_t).
  \label{eq:skip-model}
\end{equation}
If we assume Gaussian scores and a uniformly random query direction, both factors follow from block statistics:
\begin{align*}
  J_t &= 1 - e^{-\|q\|^2 \langle \operatorname{tr}\Sigma_K \rangle_t/(2D^2)}, \\
  R_t &= \frac{\|q\|}{D}\Big\langle \frac{\|\Sigma_{KV}\|_F}{\|\bar v\|} \Big\rangle_t,
\end{align*}
where $D$ is the head dimension, $\Sigma_K$, $\Sigma_{KV}$, and $\bar v$ are the key covariance, key--value covariance, and mean value of a cluster, $\langle\cdot\rangle_t$ averages a per-cluster quantity over the tile's two clusters weighted by key count, and $\|q\|$ is the root-mean-square query norm of the query block.
Skimming error also grows roughly with mass, at about a fifth of the skipping error (\figref{fig:motiv-entropy}b), so as a simplification the router models it as a fixed fraction $\alpha$ of the skipping error, $E_t(\text{skim}) \approx \alpha E_t(\text{skip})$.
The router approximates refinement as exact and sets its error to zero.

\begin{figure}[t]
  \centering
  \includegraphics[width=\columnwidth]{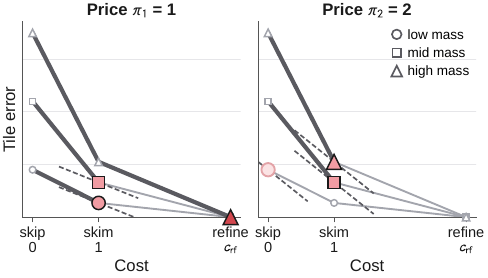}
  \caption{Choosing tiers with one price. Each tile takes the tier where a line of slope $-\pi/\Nt$ touches its error--compute curve, and a higher price moves tiles to cheaper tiers. Numbers are illustrative.}
  \label{fig:router-lambda}
\end{figure}

\paragraph{Estimating mass.}
The router estimates the mass in two ways from block means.
The first scores the mean query $\bar q$ of the query block $I$ against every key, reusing the scores that smoothing computes in \eqref{eq:smooth}.
It keeps the differences among keys but hides those among queries, which the query variance $\sigma_q^2$ around $\bar q$ restores,
\begin{equation*}
  \tilde{a}_t = \Big[\operatorname{softmax}\Big(\frac{\bar{q} K^\top}{\sqrt{D}} + \frac{\sigma_q^2}{2D^2}\|K\|^2\Big)\Big]_t,
\end{equation*}
where $K$ holds the row's keys, $\|K\|^2$ the squared norm of each key, and $[\cdot]_t$ sums the entries that belong to tile $t$.
The second scores every query against the block means $\bar K$ of the row.
It keeps the differences among queries but hides those among keys, and the mean keys lose the same fraction $J_t$ of their contribution as in skipping,
\begin{equation*}
  \tilde{b}_t \propto \frac{1}{1-J_t}\Big[\frac{1}{|I|} \sum_{i \in I} \operatorname{softmax}\Big(\frac{q_i \bar{K}^\top}{\sqrt{D}}\Big)\Big]_t,
\end{equation*}
normalized over the row.
The router averages the two, $\tilde{m}_t = \tfrac{1}{2}(\tilde{a}_t + \tilde{b}_t)$.
Because $E_t$ is relative to each head's output, the router also weights each head by $\sqrt{w_h}$, the square root of its relative output energy in the layer's previous attention call, which gives its estimate of the skipping error,
\begin{equation}
  \tilde{E}_t = \tilde{m}_t\,(J_t + R_t)\,\sqrt{w_h}.
  \label{eq:skip-error}
\end{equation}
\appref{sec:app-routing} derives the decoupling, the skipping error, and both mass corrections.

\paragraph{Choosing tiers by price.}
Because the objective \eqref{eq:objective} is a sum of per-tile terms, each tile independently takes the tier where a line of slope $-\pi/\Nt$ touches its error--compute curve (\figref{fig:router-lambda}).
We set $\alpha = 0.22$, the median ratio of skimming to skipping error in \figref{fig:motiv-entropy}b, and $c_{\mathrm{rf}} = 3$, since refining a tile takes three times the MMAs of skimming~(\secref{sec:design-fp4}).
We fix both for every model and setting, so $\pi$ is the only parameter we tune.
With these values, skimming has a larger marginal gain than refining, so a tile is  skimmed and then refined as long as each step's marginal gain is at least $\pi$,
\begin{equation}
  k_t =
  \begin{cases}
    \text{refine} & \text{if } \dfrac{\alpha \Nt \tilde{E}_t}{c_{\mathrm{rf}}-1} \ge \pi, \\[4pt]
    \text{skim} & \text{else if } (1-\alpha) \Nt \tilde{E}_t \ge \pi, \\[2pt]
    \text{skip} & \text{otherwise}.
  \end{cases}
  \label{eq:thresholds}
\end{equation}

\subsection{Preprocessing Kernels}
\label{sec:impl-preprocess}
Estimating $J_t$, $R_t$, and $\tilde m_t$ from block statistics lets planning share the preprocessing needed for quantization.
\sys fuses statistics collection, mass estimation, and tier assignment with smoothing and quantization in a few Triton~\cite{triton} passes grouped by operand.
The scores $\bar q K^\top$ in $\tilde a_t$ come from the BF16 operands of smoothing, while the scores $q_i \bar K^\top$ in $\tilde b_t$ are computed from the NVFP4 operands on the FP4 tensor cores for speed.
These operations are memory-bound, so fusion avoids repeated reads of $Q$, $K$, and $V$; tokens are accessed through the clustering tree's index rather than a permuted copy.

\section{Execution: Running All Tiers in One Pass}
\label{sec:impl}

To preserve the speed of low-bit computation, \sys runs all three tiers in a single kernel on the FP4 tensor cores.
Skimmed tiles follow the NVFP4 computation of SageAttention3 (\secref{sec:bg-lowbit}), and every tile adds its NVFP4 product to the full-precision smoothing correction of \eqref{eq:smooth}.

\subsection{Skipping with Virtual Tiles}
\label{sec:impl-virtual}
A skipped tile replaces the keys and values of each of its key clusters $\beta$ with their means $\bar k_\beta$ and $\bar v_\beta$.
For a query $q$ with $s_\beta = q \cdot \bar k_\beta / \sqrt{D}$, the $n_\beta$ keys of the cluster, each replaced by the mean, add to the softmax normalizer and to the weighted sum of values exactly what one key with an additional logit of $\ln n_\beta$ adds,
\begin{equation}
  \big(n_\beta e^{s_\beta},\; n_\beta e^{s_\beta}\,\bar v_\beta\big) = \big(e^{s_\beta + \ln n_\beta},\; e^{s_\beta + \ln n_\beta}\,\bar v_\beta\big).
  \label{eq:virtual}
\end{equation}
The kernel therefore appends the block means $\bar K$ and $\bar V$ to the keys as virtual tiles, which every query block visits as skimmed tiles.
As with real keys, the correction $\bar q(\bar K - \bar k)^\top$ completes each query's score.
The correction also adds $\ln n_\beta$ to a cluster whose tile is skipped, so that its mean represents all $n_\beta$ keys, and $-\infty$ to a cluster whose tile is kept, so that the tile is not counted twice.

\subsection{Refining Tiles on FP4 Tensor Cores}
\label{sec:design-fp4}
A refined tile needs higher accuracy than a skimmed one, which could come from BF16 in two ways.
A second BF16 kernel over the refined tiles would need an extra launch, load $Q$ and write the output a second time, and add a merge pass.
BF16 products inside the same kernel would avoid these costs, but BF16 tensor cores are several times slower than the FP4 ones (\secref{sec:bg-lowbit}).
\sys instead refines tiles on the FP4 tensor cores, where the three NVFP4 products of a refined tile ($c_{\mathrm{rf}} = 3$) still take less time than one BF16 product at peak rates.

To do so, \sys quantizes not only each operand $X \in \{\hat{Q}, \hat{K}, P, V\}$ but also its quantization residual:
\begin{equation}
  X \approx g\,(X_{\mathrm{hi}} + X_{\mathrm{lo}}),
  \label{eq:hilo}
\end{equation}
where $X_{\mathrm{hi}} = \mathcal{Q}(X/g)$ and $X_{\mathrm{lo}} = \mathcal{Q}(X/g - X_{\mathrm{hi}})$, $\mathcal{Q}$ is NVFP4 quantization, and $g$ is a power of two shared by both terms, so dividing by it is exact and the three products below accumulate without rescaling in between.
A skimmed tile uses only the $\mathrm{hi}$ terms.
A refined tile adds the products with one residual and drops the product of two residuals, which is second order:
\begin{align}
  S &\approx g_Q g_K \big(\hat{Q}_{\mathrm{hi}} \hat{K}_{\mathrm{hi}}^\top + \hat{Q}_{\mathrm{lo}} \hat{K}_{\mathrm{hi}}^\top + \hat{Q}_{\mathrm{hi}} \hat{K}_{\mathrm{lo}}^\top\big) + C, \\
  O &\mathrel{+}= g_P g_V \big(P_{\mathrm{hi}} V_{\mathrm{hi}} + P_{\mathrm{lo}} V_{\mathrm{hi}} + P_{\mathrm{hi}} V_{\mathrm{lo}}\big),
\end{align}
where $\hat{Q}$, $\hat{K}$, and $C$ are the smoothed operands and their correction of \eqref{eq:smooth}, the $1/\sqrt{D}$ scaling is omitted, and $P_{\mathrm{lo}}$ is the NVFP4 residual of the probabilities, formed in registers.

\subsection{Attention Kernel}
\label{sec:impl-kernel}

We implement the kernel for sm\_120, the architecture of the RTX Blackwell GPUs that we evaluate, in the CuTe DSL of CUTLASS~\cite{cutlass}.
It extends a warp-specialized NVFP4 attention kernel on 128-query by 128-key tiles, in which one producer warp loads operands through the Tensor Memory Accelerator (TMA) and eight MMA warps add each NVFP4 product to the full-precision correction $C$.

\paragraph{Loading by tier.}
Besides the virtual tiles, the producer loads the key blocks of the tiles the router kept, with $K_{\mathrm{lo}}$ and $V_{\mathrm{lo}}$ for refined tiles, and the MMA warps run the residual products only on refined tiles.

\paragraph{Balancing uneven query blocks.}
Query blocks differ in cost, which the kernel estimates from their kept and refined tiles, so the persistent kernel deals them to cooperative thread arrays (CTAs) in a snake order, largest first, to balance load.

Together with the preprocessing kernels, the attention kernel forms a drop-in backend that replaces the self-attention call of each transformer block, so \sys plugs into pipelines such as SGLang Diffusion~\cite{sglang,sglang-diffusion} without changing the model.

\section{Evaluation}
\label{sec:eval}

\subsection{Setup}
\label{sec:eval-setup}

\paragraph{Models and hardware.}
We evaluate four video DiTs from three families in five settings, each on a single Blackwell GPU (\tabref{tab:e2e}).
The RTX 5090 (32\,GB) runs Wan2.2-TI2V-5B~\cite{wan} at 720p and HunyuanVideo~\cite{hunyuanvideo} at 540p, HunyuanVideo with FP8 weights to fit in memory.
The RTX PRO 6000 (96\,GB) runs Wan2.2-T2V-A14B, HunyuanVideo at 720p, and LTX-2.5~\cite{ltx} (image-to-video), all in BF16 at their official settings.
The text-to-video models run 50 denoising steps (40 for Wan2.2-T2V-A14B), and LTX-2.5 eight stage-1 and three stage-2 steps.
Every method runs in SGLang Diffusion~0.5.20~\cite{sglang-diffusion} with PyTorch~2.13 and CUDA~13.0 on Ubuntu~24.04, and replaces only the self-attention over the visual tokens; the dense reference runs FlashAttention-2~\cite{flashattention2}.

\paragraph{Workloads and metrics.}
The text-to-video models use 50 prompts from the Penguin Video Benchmark~\cite{hunyuanvideo}, and LTX-2.5 uses 50 image--prompt pairs from VBench~\cite{vbench,vbenchpp}, each with a fixed seed.
For fidelity, we report PSNR, SSIM, and LPIPS against the dense video per frame, and for quality the mean of five VBench dimensions.
For speed, we report the end-to-end and per-attention-call speedup over dense (\figref{fig:eval-stages}).

\begin{figure}[t]
  \centering
  \includegraphics[width=\columnwidth]{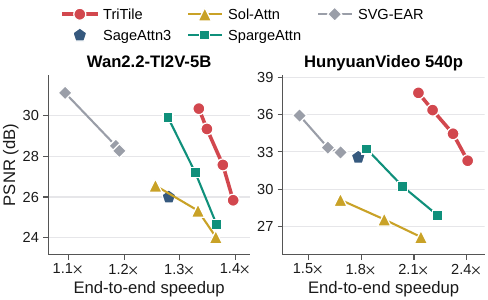}
  \caption{End-to-end PSNR versus speedup over dense on the RTX 5090 over the settings of \secref{sec:eval-setup}. Each point averages over the same 50 prompts.}
  \label{fig:eval-e2e}
\end{figure}

\paragraph{Baselines.}
We compare against training-free baselines of the three approaches in \figref{fig:policy}: sparse (S) Sol-Attn~\cite{sol-attn} and SVG-EAR~\cite{svg-ear}, low-bit (L) SageAttention3~\cite{sageattention3}, and combined (S+L) SpargeAttn~\cite{sparge1}.
They compute tiles in BF16 (S), NVFP4 (L), and INT8/FP8 (S+L).
Each baseline runs at its default setting in \tabref{tab:e2e}: Sol-Attn at $\tau = 1.0$ and SpargeAttn at top-$k$ 0.5, their defaults in SGLang Diffusion, and SVG-EAR at top-$p$ 0.85 with 300 query and 1000 key clusters on Wan2.2 and LTX-2.5 and 400 and 1000 on HunyuanVideo, halved in the two RTX 5090 settings; SageAttention3 has no setting.
In these two settings, \figref{fig:eval-e2e} also varies each knob around its default: $\tau \in \{0.5, 1.0, 1.5\}$, top-$k \in \{0.3, 0.5, 0.7\}$, top-$p \in \{0.75, 0.85\}$ and the halved and full cluster counts for SVG-EAR, and $\pi \in \{0.1, 0.15, 0.3, 0.6\}$ for \sys.
We exclude fine-tuned methods~\cite{vsa,sla,sparge2,fpsattn}.
Every method runs the first 20\% of the steps (the stage-1 steps on LTX-2.5) and the first transformer block densely.
\sys runs at $\pi = 0.15$ and $0.3$ (only $0.3$ on HunyuanVideo 720p and LTX-2.5).

\begin{table}[!t]
\centering
\footnotesize
\setlength{\tabcolsep}{1.1pt}
\caption{End-to-end results over 50 prompts: fidelity to the dense video (PSNR and SSIM, higher is better; LPIPS, lower is better), VBench quality (mean of five dimensions), and speedup over dense (E2E). Density: share of tiles computed, refined ones in parentheses. Tokens: visual tokens per sequence (stage 2 for LTX-2.5). \sys ($\pi$): price $\pi$. S, L, S+L: sparse, low-bit (4-bit), and combined (8-bit) attention. Bold: best method.}
\label{tab:e2e}
\vskip 0.1in %
\begin{tabular}{lcccccc}
\toprule
Method & PSNR & SSIM & LPIPS & VBench & Density & E2E \\
\midrule
\multicolumn{7}{c}{\textbf{RTX 5090 (32\,GB)}} \\
\midrule
\multicolumn{7}{l}{\emph{Wan2.2-TI2V-5B: 720p, 121f, 27K tokens}} \\
Dense & -- & -- & -- & 0.827 & 100\% & 1.00$\times$ \\
Sol-Attn\,{\scriptsize(S)} & 25.33 & 0.863 & 0.104 & 0.826 & 23\% & 1.33$\times$ \\
SVG-EAR\,{\scriptsize(S)} & 28.51 & 0.907 & 0.062 & 0.823 & 17\% & 1.19$\times$ \\
SageAttn3\,{\scriptsize(L)} & 26.00 & 0.863 & 0.096 & 0.820 & 100\% & 1.28$\times$ \\
SpargeAttn\,{\scriptsize(S+L)} & 27.19 & 0.895 & 0.074 & 0.826 & 50\% & 1.33$\times$ \\
\sys{} (0.15) & \textbf{29.35} & \textbf{0.919} & \textbf{0.048} & 0.826 & 29\%\,(10\%) & 1.35$\times$ \\
\sys{} (0.3) & 27.58 & 0.892 & 0.066 & 0.824 & 21\%\,(6\%) & \textbf{1.38$\times$} \\
\cmidrule(lr){1-7}
\multicolumn{7}{l}{\emph{HunyuanVideo (FP8): 540p, 129f, 67K tokens}} \\
Dense & -- & -- & -- & 0.826 & 100\% & 1.00$\times$ \\
Sol-Attn\,{\scriptsize(S)} & 27.57 & 0.875 & 0.100 & 0.825 & 23\% & 1.93$\times$ \\
SVG-EAR\,{\scriptsize(S)} & 33.35 & 0.938 & 0.041 & 0.822 & 22\% & 1.61$\times$ \\
SageAttn3\,{\scriptsize(L)} & 32.58 & 0.933 & 0.048 & 0.823 & 100\% & 1.78$\times$ \\
SpargeAttn\,{\scriptsize(S+L)} & 30.23 & 0.916 & 0.065 & 0.824 & 50\% & 2.04$\times$ \\
\sys{} (0.15) & \textbf{36.37} & \textbf{0.959} & \textbf{0.022} & 0.826 & 28\%\,(8\%) & 2.21$\times$ \\
\sys{} (0.3) & 34.46 & 0.948 & 0.030 & 0.826 & 19\%\,(5\%) & \textbf{2.32$\times$} \\
\midrule[\heavyrulewidth]
\multicolumn{7}{c}{\textbf{RTX PRO 6000 (96\,GB)}} \\
\midrule
\multicolumn{7}{l}{\emph{Wan2.2-T2V-A14B: 720p, 81f, 76K tokens}} \\
Dense & -- & -- & -- & 0.851 & 100\% & 1.00$\times$ \\
Sol-Attn\,{\scriptsize(S)} & 18.98 & 0.682 & 0.248 & 0.850 & 22\% & 1.69$\times$ \\
SVG-EAR\,{\scriptsize(S)} & 23.26 & 0.812 & 0.128 & 0.851 & 26\% & 1.35$\times$ \\
SageAttn3\,{\scriptsize(L)} & 22.53 & 0.783 & 0.147 & 0.847 & 100\% & 1.37$\times$ \\
SpargeAttn\,{\scriptsize(S+L)} & 20.24 & 0.732 & 0.196 & 0.849 & 50\% & 1.59$\times$ \\
\sys{} (0.15) & \textbf{24.40} & \textbf{0.839} & \textbf{0.104} & 0.850 & 35\%\,(8\%) & 1.72$\times$ \\
\sys{} (0.3) & 22.98 & 0.802 & 0.131 & 0.850 & 23\%\,(5\%) & \textbf{1.83$\times$} \\
\cmidrule(lr){1-7}
\multicolumn{7}{l}{\emph{HunyuanVideo: 720p, 129f, 119K tokens}} \\
Dense & -- & -- & -- & 0.834 & 100\% & 1.00$\times$ \\
Sol-Attn\,{\scriptsize(S)} & 28.02 & 0.888 & 0.102 & 0.835 & 21\% & 1.97$\times$ \\
SVG-EAR\,{\scriptsize(S)} & 35.73 & 0.957 & 0.028 & 0.832 & 21\% & 1.60$\times$ \\
SageAttn3\,{\scriptsize(L)} & 33.13 & 0.938 & 0.049 & 0.832 & 100\% & 1.49$\times$ \\
SpargeAttn\,{\scriptsize(S+L)} & 32.23 & 0.936 & 0.051 & 0.833 & 50\% & 1.83$\times$ \\
\sys{} (0.3) & \textbf{36.03} & \textbf{0.958} & \textbf{0.026} & 0.834 & 19\%\,(4\%) & \textbf{2.28$\times$} \\
\cmidrule(lr){1-7}
\multicolumn{7}{l}{\emph{LTX-2.5 (I2V, two-stage): 1088p, 721f, 186K tokens}} \\
Dense & -- & -- & -- & 0.844 & 100\% & 1.00$\times$ \\
Sol-Attn\,{\scriptsize(S)} & 27.91 & 0.880 & 0.081 & 0.846 & 19\% & 1.74$\times$ \\
SVG-EAR\,{\scriptsize(S)} & 33.30 & 0.942 & 0.036 & 0.844 & 44\% & 1.12$\times$ \\
SageAttn3\,{\scriptsize(L)} & 33.12 & 0.940 & 0.037 & 0.843 & 100\% & 1.36$\times$ \\
SpargeAttn\,{\scriptsize(S+L)} & 29.68 & 0.906 & 0.060 & 0.844 & 50\% & 1.60$\times$ \\
\sys{} (0.3) & \textbf{34.12} & \textbf{0.947} & \textbf{0.031} & 0.844 & 27\%\,(3\%) & \textbf{1.84$\times$} \\
\bottomrule
\end{tabular}

\end{table}
\suppressfloats[t]

\subsection{End-to-End Performance}
\label{sec:eval-e2e}

In each of the five settings of \tabref{tab:e2e}, one price of \sys is faster end to end than every baseline and also closer to the dense video in PSNR, SSIM, and LPIPS: $\pi = 0.15$ in the three settings where it runs, and $\pi = 0.3$ in the other two.
At $\pi = 0.15$, \sys computes more tiles than the sparse baselines, 28--35\%, but refines only 8--10\%, and its PSNR is 0.8--3.0\,dB above that of SVG-EAR, the most faithful baseline.
\sys preserves quality, with VBench within 0.003 of dense in every setting.
At $\pi = 0.3$, \sys computes about as many tiles as the sparse baselines (19--27\% versus 17--44\%) but in NVFP4 rather than BF16, and still exceeds the PSNR of SVG-EAR, the most faithful baseline, in three of the five settings and stays within 1\,dB of it in the other two, while running 1.16--1.64\x faster than it. It is the fastest method in every setting, at 1.38--2.32\x, with PSNR 0.4--8.0\,dB above SpargeAttn and Sol-Attn.
In the sweep of \figref{fig:eval-e2e}, \sys Pareto-dominates every baseline setting except SVG-EAR with the original cluster counts on Wan2.2-TI2V-5B: the original counts raise its PSNR by 2.6\,dB over the halved ones on both models, to 0.8\,dB above \sys at $\pi = 0.1$ on Wan2.2-TI2V-5B, but at 1.09\x it is too slow to be practical, below every setting of \sys; on HunyuanVideo 540p, \sys at $\pi = 0.1$ is both 1.8\,dB more faithful and 1.47\x faster than it.

The end-to-end gains are bounded because the dense steps and the work outside attention are unchanged, and they are larger where attention takes more of a dense step (\appref{sec:app-e2e}).
LTX-2.5 gains less because only its three stage-2 steps run sparsely.
\begin{figure}[!t]
  \centering
  \includegraphics[width=\columnwidth]{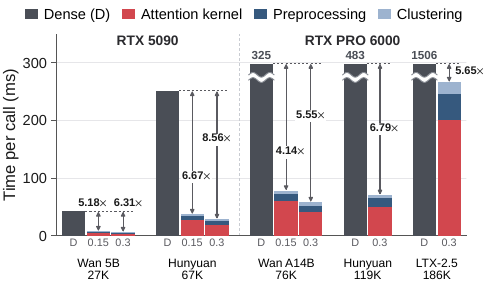}
  \caption{Time of one attention call, averaged over the sparse steps of the end-to-end runs, for dense attention (D, truncated) and \sys at $\pi = 0.15$ and $0.3$. Arrows: speedup over dense.}
  \label{fig:eval-stages}
\end{figure}

\begin{figure}[!t]
  \centering
  \includegraphics[width=\columnwidth]{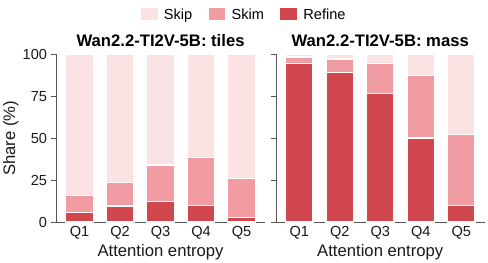}
  \caption{Share of a row's tiles and of its mass in each tier on Wan2.2-TI2V-5B ($\pi = 0.15$), per quintile of row entropy.}
  \label{fig:eval-tiers}
\end{figure}

\begin{figure*}[t]
  \centering
  \includegraphics[width=\textwidth]{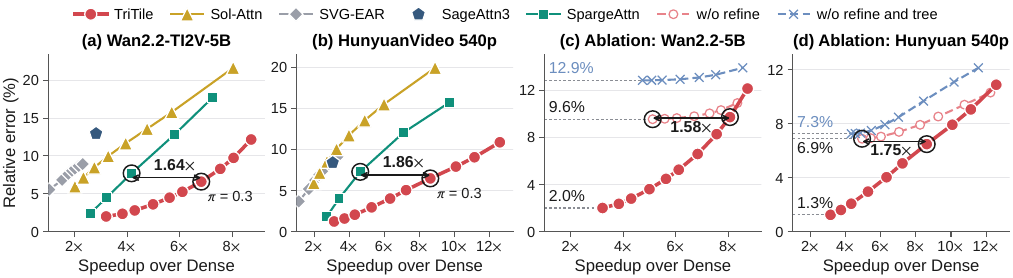}
  \caption{Relative error versus attention speedup over dense on the captured calls: (a,~b)~\sys against the baselines, (c,~d)~its ablations. Arrows: speedup of \sys over the circled point nearest in error ($\pi = 0.3$ in a,~b); dashed: lowest errors.}
  \label{fig:eval-layer}
\end{figure*}

\label{sec:eval-breakdown}
\label{sec:eval-overhead}

\figref{fig:eval-stages} breaks one attention call of the end-to-end runs into its stages.
The clustering and the preprocessing, which also collects the block statistics and routes the tiles, take 4.0--7.4\% of the time of dense attention, so a whole call still runs 4.14--8.56\x faster than dense.
Their share of a call, 23--46\%, is largest when the kernel is short, on short sequences with few computed tiles.
\appref{sec:app-e2e} reports every VBench dimension, and \figref{fig:intro-qual}, \appref{sec:app-qual-t2v}, and \appref{sec:app-qual-ltx} show generated frames.

\subsection{Attention-Level Analysis}
\label{sec:eval-layer}

We replay the attention calls captured in \secref{sec:motivation} (160 and 240 calls of Wan2.2-TI2V-5B and HunyuanVideo 540p on the RTX 5090) and sweep the knob of every method.
The time of every method includes everything it runs before attention, such as clustering and deciding which tiles to compute.

\sys reaches a better trade-off than every baseline on both models (\figref{fig:eval-layer}a,b).
At $\pi = 0.3$ (6.6\% and 6.5\% error, circled in \figref{fig:eval-layer}a,b), it runs 1.64\x and 1.86\x faster than the nearest measured point of SpargeAttn, the strongest baseline, whose error is higher (7.7\% and 7.3\%).

\figref{fig:eval-tiers} shows that \sys adapts its tiers well to the attention distribution of each row.
In the sharpest rows (Q1), refining only 3--12\% of the tiles covers 95\% of the mass, so the rest can be skipped at little cost. Toward more diffuse rows, the mass held by refined tiles falls and skimmed tiles take over; in Q5 every tile holds so little mass that its score rarely justifies refinement.
HunyuanVideo behaves the same way (\appref{sec:app-tiers}).

\begin{table}[!t]
\centering
\footnotesize
\setlength{\tabcolsep}{3pt}
\caption{Router score ablation at equal kept and refined tiles: error relative to the full score (geometric mean) and share of calls with a higher error, pooled over four budgets.}
\label{tab:eval-router}
\vskip 0.1in %
\begin{tabular}{lcccccc}
\toprule
 & \multicolumn{3}{c}{Error vs.\ full} & \multicolumn{3}{c}{Calls with higher error} \\
\cmidrule(lr){2-4} \cmidrule(lr){5-7}
Variant & Wan & Hun. & All & Wan & Hun. & All \\
\midrule
w/o $J_t + R_t$ & 1.28$\times$ & 1.12$\times$ & 1.18$\times$ & 100\% & 92\% & 95\% \\
w/o $\sqrt{w_h}$ & 1.08$\times$ & 1.07$\times$ & 1.07$\times$ & 97\% & 69\% & 80\% \\
w/o mass corr. & 1.03$\times$ & 1.04$\times$ & 1.03$\times$ & 89\% & 81\% & 84\% \\
\bottomrule
\end{tabular}
\end{table}

\subsection{Ablations}
\label{sec:eval-ablation}

We remove the refine tier and the clustering tree from \sys (\figref{fig:eval-layer}c,d).

\paragraph{Refinement.}
Without the refine tier, the error of \sys never falls below 9.6\% and 6.9\% on Wan2.2-TI2V-5B and HunyuanVideo 540p.
Refining the tiles that the router selects removes this floor, down to 2.0\% and 1.3\%, and at its lowest error \sys is 1.58\x and 1.75\x faster than it.

\paragraph{Clustering tree.}
Without the tree, tiles mix dissimilar tokens, so their means represent them poorly and \sys must compute more tiles to reach the same error.
Removing the tree from the variant without refinement raises its lowest error from 9.6\% to 12.9\% on Wan2.2-TI2V-5B and from 6.9\% to 7.3\% on HunyuanVideo (\figref{fig:eval-layer}c,d).

\subsection{Router Score}
\label{sec:eval-router}

We remove one factor of the router score $\tilde{E}_t = \tilde{m}_t (J_t + R_t) \sqrt{w_h}$ at a time and, so that only the ranking of the tiles changes, keep and refine as many tiles as the full score in every captured call, at four budgets (20\% or 30\% of the tiles kept, 10\% or 20\% of them refined) (\tabref{tab:eval-router}).
Every factor helps, in the same order at every budget.
Without $J_t + R_t$, the router ranks tiles by mass alone and skips tiles that their means represent poorly, which raises the error by 18\% and does so in 95\% of the calls.
Without the head weight, the router allows the same relative error in every head, although the heads with a large output dominate the error of the layer, and the error rises by 7\%. The mass corrections (the query-variance term in $\tilde a_t$ and the factor $1/(1-J_t)$ in $\tilde b_t$) reduce the error by a small but consistent 3\%.

\subsection{Refinement in NVFP4 versus BF16} %
\label{sec:eval-kernel}

\begin{table}[t]
\centering
\footnotesize
\setlength{\tabcolsep}{2.6pt}
\caption{Attention kernel alone on the RTX 5090: relative L2 error, time (ms), and speedup over dense. 2-pass BF16: the refined tiles computed in a separate BF16 pass instead.}
\label{tab:eval-kernel}
\vskip 0.1in %
\begin{tabular}{lrrrrrr}
\toprule
 & \multicolumn{3}{c}{Wan2.2-TI2V-5B} & \multicolumn{3}{c}{HunyuanVideo 540p} \\
\cmidrule(lr){2-4} \cmidrule(lr){5-7}
Method & Err. & Time & Speed. & Err. & Time & Speed. \\
\midrule
Dense & -- & 41.7 & 1.00$\times$ & -- & 251.7 & 1.00$\times$ \\
\sys{} (0.15) & 4.5\% & 4.3 & 9.73$\times$ & 4.0\% & 29.4 & 8.57$\times$ \\
\quad 2-pass BF16 & 4.4\% & 6.1 & 6.80$\times$ & 4.0\% & 40.5 & 6.21$\times$ \\
\sys{} (0.3) & 6.5\% & 3.1 & 13.5$\times$ & 6.5\% & 19.5 & 12.9$\times$ \\
\quad 2-pass BF16 & 6.5\% & 4.3 & 9.59$\times$ & 6.4\% & 26.2 & 9.61$\times$ \\
\bottomrule
\end{tabular}
\end{table}
\tabref{tab:eval-kernel} times the attention kernel of \sys alone, without planning.
At $\pi = 0.15$, the kernel alone runs 9.73\x and 8.57\x faster than dense on these calls.
Refining the same tiles in a separate BF16 pass, merged by log-sum-exp, instead gives almost the same error, within 0.1 points at both prices, but slows the kernel by 34--43\%, so the NVFP4 residual alone is sufficient for refinement on these calls.

\section{Conclusion}
\label{sec:conclusion}
\sys shows that sparse and low-bit attention need not be competing choices: deciding for each tile whether to skip, skim, or refine it lets attention spend precision according to the estimated error reduction.
A lightweight router makes this choice from block statistics, and one fused kernel runs all three tiers on the FP4 tensor cores.
Across three families of video DiTs, \sys achieves a better trade-off between speed and quality than existing methods, without retraining or modifying the model.

\bibliography{refs}
\bibliographystyle{mlsys2025}

\ifwithappendix
\clearpage
\appendix
\section{The Clustering Tree}
\label{sec:app-tree}

\algref{alg:tree} builds the tile-aligned PDDP tree~\cite{pddp} of \secref{sec:design-cluster} for the tokens of one head.
\sys runs it separately for the queries ($b = 128$) and the keys ($b = 64$), and processes all heads of a layer together.
Lines~1--3 project the tokens once onto their top $r$ principal components. We use $r = 32$ and $T = 8$ power iterations.
Each node of the tree is a half-open range $[s, e)$ of positions in the order $\sigma$, holding the tokens $\sigma[s], \dots, \sigma[e-1]$.
Each iteration of the loop then splits every node that spans more than one block, all nodes of a level at once: it finds the node's top principal direction in the $r$-dimensional subspace, sorts the node's tokens along it, and cuts after $\lfloor m/2 \rfloor$ blocks, so both children again span whole blocks.
The loop runs $\lceil \log_2 (N/b) \rceil$ times.
Each iteration reads $Z$ once to form the per-node $r \times r$ covariances in $O(N r^2)$, runs $T$ power iterations on them, and performs one segmented sort in $O(N \log N)$, so with $r$, $T$, and $b$ fixed the loop costs $O\big((N r^2 + N \log N) \log (N/b)\big)$ in total, in addition to the initial projection.
Ties in the sort are broken by token index, which makes the order deterministic.
If $N$ is not a multiple of $b$, the last block is partially filled.

\begin{algorithm}[t]
  \caption{Tile-aligned PDDP tree for one head.}
  \label{alg:tree}
  \begin{algorithmic}[1]
    \REQUIRE tokens $X \in \mathbb{R}^{N \times D}$, block size $b$, rank $r$, power iterations $T$
    \ENSURE order $\sigma$ (a permutation of $0..N-1$) whose consecutive groups of $b$ tokens are the blocks
    \STATE $\mu \gets \mathrm{mean}(X)$, \; $C \gets \mathrm{cov}(X)$
    \STATE $U \gets$ top-$r$ eigenvectors of $C$ \hfill $\triangleright$ $D \times r$
    \STATE $Z \gets (X - \mu)\, U$ \hfill $\triangleright$ $N \times r$
    \STATE $\sigma \gets (0, \dots, N-1)$; \; $\mathcal{S} \gets \{[0, N)\}$ \hfill $\triangleright$ nodes $[s, e)$
    \WHILE{some $[s, e) \in \mathcal{S}$ spans more than one block}
      \FORALL{$[s, e) \in \mathcal{S}$ with $m = \lceil (e - s)/b \rceil > 1$ \textbf{in parallel}}
        \STATE $\bar{z} \gets \mathrm{mean}(Z_{\sigma[s:e)})$, \; $\Sigma \gets \mathrm{cov}(Z_{\sigma[s:e)})$ \hfill $\triangleright$ $r \times r$
        \STATE $u \gets \textsc{PowerIteration}(\Sigma, T)$ \hfill $\triangleright$ top direction
        \STATE $p_i \gets (Z_{\sigma[i]} - \bar{z}) \cdot u$ \; for $i \in [s, e)$
        \STATE sort $\sigma[s:e)$ by $(p_i, \sigma[i])$
        \STATE $c \gets s + \lfloor m/2 \rfloor\, b$
        \STATE replace $[s, e)$ in $\mathcal{S}$ by $[s, c)$ and $[c, e)$
      \ENDFOR
    \ENDWHILE
    \STATE \textbf{return} $\sigma$
  \end{algorithmic}
\end{algorithm}

\section{Router Derivations}
\label{sec:app-routing}

This appendix derives the first-order decoupling of tile errors, explains the approximations behind \eqref{eq:skip-model}, and specifies the router score.

\subsection{Decoupling the Tiles}
\label{sec:app-decouple}
For one query $q$ with scores $s_j=q\cdot k_j/\sqrt{D}$ and values $v_j$, write the exact output as $O=U/Z$, where $Z=\sum_t Z_t$ is the softmax normalizer and $U=\sum_t U_t$ the weighted value sum, with $Z_t=\sum_{j\in t}e^{s_j}$ and $U_t=\sum_{j\in t}e^{s_j}v_j$ the contributions of tile $t$.
A tier changes tile $t$ by $\Delta Z_t$ and $\Delta U_t$, so the approximated output is $O'=(U+\sum_t\Delta U_t)/(Z+\sum_t\Delta Z_t)$ and
\begin{equation*}
  O'-O=\frac{\sum_t\Delta U_t-O\sum_t\Delta Z_t}{Z+\sum_t\Delta Z_t}.
\end{equation*}%
Dividing the numerator and the denominator by $Z$ and defining $\epsilon_t=\Delta Z_t/Z$, the relative change of the normalizer, and $\delta_t=(\Delta U_t-O\Delta Z_t)/Z$, the output change that tile $t$ would cause at the original normalizer, gives
\begin{equation}
  O'-O=\frac{\sum_t\delta_t}{1+\sum_t\epsilon_t}.
  \label{eq:app-decouple}
\end{equation}
Approximating tile $t$ alone changes the output by $\delta_t/(1+\epsilon_t)$.
When both the individual and aggregate changes of the normalizer are small, these single-tile changes and their sum are well approximated by $\delta_t$ and $\sum_t\delta_t$, respectively.
The triangle inequality gives the exact bound
\begin{equation*}
  \frac{\|O'-O\|}{\|O\|}
  \le \frac{\sum_t\|\delta_t\|/\|O\|}{|1+\sum_t\epsilon_t|}
  = \sum_t \gamma_t\,E_t,
\end{equation*}
where $E_t = \|\delta_t\|/(\|O\|\,|1+\epsilon_t|)$ is the relative error of approximating tile $t$ alone and $\gamma_t = |1+\epsilon_t|/|1+\sum_{t'}\epsilon_{t'}|$.
The tiles interact only through the shared normalizer in $\gamma_t$.
Under a joint small-perturbation expansion of $\Delta Z$ and $\Delta U$, both $\gamma_t-1$ and $E_t$ are first order, so $\sum_t\gamma_t E_t$ and $\sum_t E_t$ differ only at second order: to first order, the sum of single-tile relative errors bounds the error of the query's output.

For finite changes, $\gamma_t$ can be bounded explicitly for the tiers in use.
Skipping only lowers the normalizer, since by Jensen's inequality the mean key of a cluster receives at most the weight of the keys it replaces (\secref{sec:app-skip}), so $\epsilon_t\le0$ for every skipped tile; let $L=-\sum_{k_t=\text{skip}}\epsilon_t$ be the fraction of the normalizer that the skipped tiles lose in total.
A skimmed tile changes its contribution $Z_t$ only through the rounding of its scores: if every rounded score differs from the exact one by at most $\tau$, then $e^{-\tau}Z_t\le Z_t+\Delta Z_t\le e^{\tau}Z_t$.
Together, $|1+\epsilon_t|\le e^{\tau}$ for every tile and $1+\sum_{t'}\epsilon_{t'}\ge e^{-\tau}-L$, so if $e^{\tau}L<1$, then $\gamma_t\le e^{2\tau}/(1-e^{\tau}L)$ and
\begin{equation*}
  \frac{\|O'-O\|}{\|O\|}\le\frac{e^{2\tau}}{1-e^{\tau}L}\sum_t E_t.
\end{equation*}
For small $L$ and $\tau$ the factor is close to one, and the router keeps $L$ small because it skips the tiles that hold little mass.
This motivates the separable objective in \eqref{eq:objective}.

\subsection{The Error of Skipping a Tile}
\label{sec:app-skip}
We first derive the exact output change for the two mean representatives used by the kernel, then state the approximations behind \eqref{eq:skip-model}.
Consider one query $q$ with scores $s_j=q\cdot k_j/\sqrt{D}$, normalizer $Z=\sum_j e^{s_j}$, and exact output $O=\sum_j e^{s_j}v_j/Z$.
For each of the two key clusters $\beta$ of tile $t$, with $n_\beta\le64$ keys, mean key $\bar k_\beta$, and mean value $\bar v_\beta$, define
\begin{equation*}
  Z_\beta=\sum_{j\in\beta}e^{s_j},\qquad
  m_\beta=Z_\beta/Z,\qquad
  \mu_\beta=\frac{\sum_{j\in\beta}e^{s_j}v_j}{Z_\beta},
\end{equation*}
so that $m_t=\sum_{\beta\in t}m_\beta$ is the mass of tile $t$ for this query; averaged over the queries of a block, it is the mass of \eqref{eq:mass}.
Mean replacement gives cluster $\beta$ the weight $\hat Z_\beta=n_\beta e^{q\cdot\bar k_\beta/\sqrt{D}}$ and the weighted value $\hat Z_\beta\bar v_\beta$, so the cluster keeps a fraction $\rho_\beta=\hat Z_\beta/Z_\beta$ of its softmax weight and loses $J_\beta=1-\rho_\beta$.
Replacing only tile $t$ therefore changes the normalizer by $\Delta Z_t=-\sum_{\beta\in t}Z_\beta J_\beta$ and the weighted value sum by $\Delta U_t=\sum_{\beta\in t}Z_\beta(\rho_\beta\bar v_\beta-\mu_\beta)$.
In \eqref{eq:app-decouple} with only tile $t$ changed,
\begin{equation*}
  \epsilon_t=-\sum_{\beta\in t}m_\beta J_\beta,\qquad
  \delta_t=\sum_{\beta\in t}m_\beta(\rho_\beta\bar v_\beta-\mu_\beta+J_\beta O).
\end{equation*}
Writing $\mu_\beta=\rho_\beta\mu_\beta+J_\beta\mu_\beta$ in the $-\mu_\beta$ term, the change $\delta_t/(1+\epsilon_t)$ is exactly
\begin{equation}
  \begin{gathered}
  O'-O = \frac{\sum_{\beta\in t}m_\beta d_\beta}
                   {1-\sum_{\beta\in t}m_\beta J_\beta},\\
  d_\beta = \rho_\beta(\bar v_\beta-\mu_\beta)
                 +J_\beta(O-\mu_\beta).
  \end{gathered}
  \label{eq:app-skip-exact}
\end{equation}
The first term of $d_\beta$ is the value shift: $\bar v_\beta$ weights the values of the cluster equally, whereas $\mu_\beta$ weights them by attention.
The second is the effect of the lost weight: normalization spreads the lost fraction $J_\beta$ over all keys, which pulls the output away from $\mu_\beta$.
These are the two effects that $R_t$ and $J_t$ model in \secref{sec:design-router}.
This change is exact; the approximations start below.

\paragraph{Lost weight.}
The mean score of cluster $\beta$ is $q\cdot\bar k_\beta/\sqrt{D}$, so Jensen's inequality gives $\hat Z_\beta\le Z_\beta$, hence $0\le\rho_\beta\le1$.
If the scores within the cluster are approximately Gaussian with variance $\sigma_\beta^2=q^\top\Sigma_{K,\beta}q/D$, where $\Sigma_{K,\beta}$ is the key covariance of the cluster, then $\mathbb{E}[e^{s}]=e^{\mathbb{E}[s]+\sigma_\beta^2/2}$ gives
\begin{equation*}
  Z_\beta\approx n_\beta e^{q\cdot\bar k_\beta/\sqrt{D}+\sigma_\beta^2/2},
  \qquad J_\beta\approx1-e^{-\sigma_\beta^2/2}.
\end{equation*}
If we treat the direction of $q$ as uniformly random, then $q^\top\Sigma_{K,\beta}q\approx\|q\|^2\operatorname{tr}(\Sigma_{K,\beta})/D$, which makes this computable from block moments.
Here and below, a quadratic form in $q$ is replaced by its expectation over query directions of the same norm: $\mathbb{E}[q^\top A q]=\|q\|^2\operatorname{tr}(A)/D$ and $\mathbb{E}\|A^\top q\|^2=\|q\|^2\|A\|_F^2/D$ for any $D\times D$ matrix $A$.
Over the tile, the exact lost-weight fraction is $\sum_{\beta\in t}m_\beta J_\beta/m_t$, which requires the unknown cluster masses.
The router instead applies the exponential formula to the key-count-weighted moments of the two clusters to obtain $J_t$, a moment-based approximation to that fraction rather than an exact aggregation.

\paragraph{Value shift.}
Let $\tilde s_j=q\cdot(k_j-\bar k_\beta)/\sqrt{D}$ be the deviation of each score in cluster $\beta$ from the mean score; these deviations sum to zero.
Since $\mu_\beta=\sum_{j\in\beta}e^{\tilde s_j}v_j/\sum_{j\in\beta}e^{\tilde s_j}$, the first-order expansion $e^{\tilde s_j}\approx1+\tilde s_j$, accurate when these deviations are small, gives
\begin{equation*}
  \mu_\beta-\bar v_\beta\approx\frac{1}{n_\beta}\sum_{j\in\beta}\tilde s_j\,v_j=\Sigma_{KV,\beta}^\top q/\sqrt{D},
\end{equation*}
where $\Sigma_{KV,\beta}$ is the key--value covariance of the cluster; the last step uses $\sum_{j\in\beta}\tilde s_j=0$ to replace $v_j$ by $v_j-\bar v_\beta$, so that $\frac{1}{n_\beta}\sum_{j\in\beta}\tilde s_j v_j=\frac{1}{n_\beta}\sum_{j\in\beta}(v_j-\bar v_\beta)(k_j-\bar k_\beta)^\top q/\sqrt{D}$.
If we again treat the direction of $q$ as uniformly random, then $\|\Sigma_{KV,\beta}^\top q\|^2\approx\|q\|^2\|\Sigma_{KV,\beta}\|_F^2/D$, so the relative value shift $\|\mu_\beta-\bar v_\beta\|/\|\bar v_\beta\|$ is about $\|q\|\,\|\Sigma_{KV,\beta}\|_F/(D\|\bar v_\beta\|)$, its root mean square over query directions.
The router averages this estimate over the two clusters, weighted by key count, to obtain $R_t$; as for $J_t$, this replaces the exact mass-weighted aggregate $\sum_{\beta\in t}m_\beta r_\beta/m_t$, with $r_\beta$ defined below.

\paragraph{From output change to the routing model.}
For nonzero reference norms, define
$r_\beta=\|\bar v_\beta-\mu_\beta\|/\|\bar v_\beta\|$,
$\eta_\beta=\|\bar v_\beta\|/\|O\|$, and
$\ell_\beta=\|O-\mu_\beta\|/\|O\|$.
The triangle inequality applied to \eqref{eq:app-skip-exact} yields
\begin{equation*}
 E_t(\text{skip})\le
 \frac{\sum_{\beta\in t}m_\beta
       (\rho_\beta r_\beta\eta_\beta+J_\beta\ell_\beta)}
      {1-\sum_{\beta\in t}m_\beta J_\beta}.
\end{equation*}
To obtain the inexpensive model in \eqref{eq:skip-model}, we neglect the change of the denominator when the lost mass is small, approximate $\rho_\beta$ by one in the value-shift term, and use unit-scale proxies for $\eta_\beta$ and $\ell_\beta$, which gives $E_t(\text{skip})\approx\sum_{\beta\in t}m_\beta(r_\beta+J_\beta)$.
Replacing the per-cluster factors by the moment-based tile estimates $R_t$ and $J_t$ then gives $E_t(\text{skip})\approx m_t(J_t+R_t)$.
So far $m_t$, $\|q\|$, and the error are those of one query, whereas the objective \eqref{eq:objective} is for a row, the query block $I$; we take the row error to be the average of the per-query errors.
Because clustering makes the queries of a block similar, the router applies the model once per row: it replaces the mass of each query by its average over the block, the mass $m_t$ of \eqref{eq:mass}, which it estimates by $\tilde m_t$ (\secref{sec:app-mass}), and the norm of each query by the root-mean-square norm of the block, written $\|q\|$ from here on (\secref{sec:app-router}).
This ignores how the error varies across the queries of a block.
The unit-scale substitutions and these aggregations are modeling assumptions; the final expression is neither an exact error nor a certified upper bound.
It preserves the mass dependence and the two mechanisms of mean-replacement error while avoiding full attention scores.

\setcounter{dbltopnumber}{2}\renewcommand{\dbltopfraction}{0.95}\renewcommand{\textfraction}{0.05}

\subsection{Mass Estimates}
\label{sec:app-mass}
The router needs the mass $m_t$ of \eqref{eq:mass} for every tile of a row, which averages the softmax rows of the $|I|$ queries of the block and would require all $|I|\times N$ scores of the row, where $N$ is the number of keys.
Both estimates of \secref{sec:design-router} instead average one side before the exponential: $\tilde a_t$ replaces the queries by their mean and scores it against every key ($N$ scores per block), and $\tilde b_t$ replaces each key cluster by its mean and scores every query against the cluster means ($|I|\times N/n$ scores).
For simplicity, the formulas take every key cluster of a row to hold the same number of keys $n$; the implementation weights each cluster by its size $n_\beta$, which differs from $n$ only for the last cluster of a row.

Averaging before the exponential is biased: by Jensen's inequality, $e^{\mathbb{E}[s]}\le\mathbb{E}[e^s]$, so a group of scores loses weight when it is replaced by its mean score, the more the larger its variance.
Both estimates correct this bias with the tool of \secref{sec:app-skip}: if the scores of the group are Gaussian, $\mathbb{E}[e^s]=e^{\mathbb{E}[s]+\operatorname{Var}[s]/2}$, and the variance follows from block statistics.

\paragraph{Mean query against every key.}
Writing the mass as the tile's share of the row's weight and moving the average over the queries inside the ratio,
\begin{equation*}
  m_t=\frac{1}{|I|}\sum_{i\in I}\frac{\sum_{j\in t}e^{s_{ij}}}{\sum_j e^{s_{ij}}}
  \approx\frac{\sum_{j\in t}\mathbb{E}_i[e^{s_{ij}}]}{\sum_j\mathbb{E}_i[e^{s_{ij}}]},
\end{equation*}
with $s_{ij}=q_i\cdot k_j/\sqrt{D}$.
If the queries of the block are Gaussian around $\bar q$ with covariance $(\sigma_q^2/D)\,\mathbf{I}$, where $\sigma_q^2$ is their total variance, then $s_{ij}$ has mean $\bar q\cdot k_j/\sqrt{D}$ and variance $\sigma_q^2\|k_j\|^2/D^2$ over the queries, so $\mathbb{E}_i[e^{s_{ij}}]=e^{\bar q\cdot k_j/\sqrt{D}+\sigma_q^2\|k_j\|^2/(2D^2)}$ and
\begin{gather*}
  \tilde{\Lambda}_t = \log \sum_{j \in t} \exp\!\Big(\frac{\bar{q} \cdot k_j}{\sqrt{D}} + \frac{\sigma_q^2 \|k_j\|^2}{2D^2}\Big), \\
  \tilde{a}_t = e^{\tilde{\Lambda}_t} \Big/ \textstyle\sum_{t'} e^{\tilde{\Lambda}_{t'}}.
\end{gather*}
The term $\sigma_q^2\|k_j\|^2/(2D^2)$ restores the spread of the queries that the mean query hides; moving the average inside the ratio, which weights each query by its normalizer, is the one approximation beyond the isotropic Gaussian assumption.

\paragraph{Every query against the block means.}
For one query $q$, the weight of cluster $\beta$ is $Z_\beta=\sum_{j\in\beta}e^{s_j}$, and by the definition of $J_\beta$ in \secref{sec:app-skip}, replacing its keys by their mean keeps the fraction $1-J_\beta$ of it, $n\,e^{q\cdot\bar k_\beta/\sqrt{D}}=(1-J_\beta)Z_\beta$.
Dividing the mean-key weight by $1-J_\beta$ therefore restores the lost weight exactly, and with the moment estimate of $J_\beta$ in its place, the tile's share for this query is the softmax over the cluster means with the logit of cluster $\beta$ raised by $-\ln(1-J_\beta)$.
The router simplifies this in three ways so that it fuses with the preprocessing: it uses the tile-level $J_t$ of \secref{sec:design-router} for both clusters of a tile, evaluates it at the root-mean-square query norm of the block for all of its queries, and applies the correction after averaging the uncorrected shares over the queries of the block rather than to each query's shares:
\begin{gather*}
  B_t = \frac{1}{|I|} \sum_{i \in I} \sum_{\beta \in t} \frac{e^{q_i \cdot \bar{k}_\beta/\sqrt{D}}}{\sum_{\beta'} e^{q_i \cdot \bar{k}_{\beta'}/\sqrt{D}}}, \\
  \tilde{b}_t = \frac{B_t / (1 - J_t)}{\sum_{t'} B_{t'} / (1 - J_{t'})}.
\end{gather*}
With $B_{it}$ the share of tile $t$ for query $i$ (the inner sum of $B_t$, so $B_t=\frac{1}{|I|}\sum_{i\in I}B_{it}$) and $z_i=\sum_{t'}B_{it'}/(1-J_{t'})$, the last simplification makes $\tilde b_t$ the $z_i$-weighted average of the per-query corrected shares $B_{it}/\big((1-J_t)\,z_i\big)$ instead of their uniform average.

\paragraph{Combination.}
The two estimates keep complementary information: $\tilde a_t$ keeps the differences among the keys and $\tilde b_t$ those among the queries.
The router averages them, $\tilde{m}_t = \frac{1}{2}(\tilde{a}_t + \tilde{b}_t)$, which sums to one over the row; \tabref{tab:eval-router} measures the effect of the two corrections.

\begin{table*}[t]
\centering
\scriptsize
\setlength{\tabcolsep}{2.6pt}
\caption{End-to-end details over 50 prompts for the settings of \tabref{tab:e2e}: fidelity to the dense video, attention time per sparse step (Attn., including the first transformer block, which runs densely) and latency per request (E2E) with their speedups over dense, the share of a sparse step spent in attention (Share), share of tiles computed (Density) and refined (Refined), and the five VBench dimensions (imaging quality divided by 100) with their mean. Bold: best method.}
\label{tab:app-e2e}
\vskip 0.1in %
\begin{tabular}{lccc rrrrr rr cccccc}
\toprule
 & \multicolumn{3}{c}{Fidelity to dense} & \multicolumn{5}{c}{Time and speedup} & \multicolumn{2}{c}{Tiles} & \multicolumn{6}{c}{VBench} \\
\cmidrule(lr){2-4}\cmidrule(lr){5-9}\cmidrule(lr){10-11}\cmidrule(lr){12-17}
Method & PSNR & SSIM & LPIPS & Attn.\,(s) & Attn. & Share & E2E\,(s) & E2E & Density & Refined & Subj. & Backg. & Motion & Aesth. & Imag. & Mean \\
\midrule
\multicolumn{17}{l}{\emph{Wan2.2-TI2V-5B (RTX 5090)}} \\
Dense & -- & -- & -- & 2.54 & 1.00$\times$ & 47\% & 292 & 1.00$\times$ & 100.0\% & -- & 0.951 & 0.958 & 0.989 & 0.567 & 0.669 & 0.827 \\
Sol-Attn & 25.33 & 0.863 & 0.104 & 0.72 & 3.54$\times$ & 20\% & 220 & 1.33$\times$ & 23.1\% & -- & 0.950 & 0.958 & 0.989 & 0.559 & \textbf{0.672} & 0.826 \\
SVG-EAR & 28.51 & 0.907 & 0.062 & 1.48 & 1.71$\times$ & 35\% & 247 & 1.19$\times$ & 16.6\% & -- & 0.946 & 0.954 & 0.989 & 0.563 & 0.661 & 0.823 \\
SageAttn3 & 26.00 & 0.863 & 0.096 & 0.94 & 2.69$\times$ & 25\% & 228 & 1.28$\times$ & 100.0\% & -- & 0.947 & 0.957 & 0.989 & 0.554 & 0.651 & 0.820 \\
SpargeAttn & 27.19 & 0.895 & 0.074 & 0.71 & 3.56$\times$ & 20\% & 220 & 1.33$\times$ & 50.0\% & -- & 0.950 & 0.957 & 0.989 & 0.563 & 0.671 & 0.826 \\
\sys{} (0.15) & \textbf{29.35} & \textbf{0.919} & \textbf{0.048} & 0.57 & 4.42$\times$ & 17\% & 217 & 1.35$\times$ & 29.2\% & 9.6\% & 0.949 & 0.958 & 0.989 & \textbf{0.565} & 0.670 & 0.826 \\
\sys{} (0.3) & 27.58 & 0.892 & 0.066 & \textbf{0.48} & \textbf{5.28$\times$} & 14\% & \textbf{212} & \textbf{1.38$\times$} & 20.6\% & 6.0\% & 0.947 & 0.958 & 0.989 & 0.559 & 0.669 & 0.824 \\
\cmidrule(lr){1-17}
\multicolumn{17}{l}{\emph{HunyuanVideo 540p, FP8 (RTX 5090)}} \\
Dense & -- & -- & -- & 15.09 & 1.00$\times$ & 84\% & 916 & 1.00$\times$ & 100.0\% & -- & 0.955 & 0.966 & 0.994 & 0.592 & 0.624 & 0.826 \\
Sol-Attn & 27.57 & 0.875 & 0.100 & 4.05 & 3.73$\times$ & 59\% & 474 & 1.93$\times$ & 22.7\% & -- & 0.955 & 0.964 & 0.994 & 0.591 & 0.619 & 0.825 \\
SVG-EAR & 33.35 & 0.938 & 0.041 & 6.47 & 2.33$\times$ & 70\% & 569 & 1.61$\times$ & 21.8\% & -- & 0.953 & 0.965 & 0.994 & 0.582 & 0.617 & 0.822 \\
SageAttn3 & 32.58 & 0.933 & 0.048 & 5.03 & 3.00$\times$ & 64\% & 514 & 1.78$\times$ & 100.0\% & -- & 0.955 & 0.966 & 0.994 & 0.584 & 0.616 & 0.823 \\
SpargeAttn & 30.23 & 0.916 & 0.065 & 3.40 & 4.44$\times$ & 54\% & 450 & 2.04$\times$ & 50.0\% & -- & 0.956 & \textbf{0.967} & 0.994 & 0.585 & 0.620 & 0.824 \\
\sys{} (0.15) & \textbf{36.37} & \textbf{0.959} & \textbf{0.022} & 2.52 & 6.00$\times$ & 47\% & 415 & 2.21$\times$ & 28.2\% & 7.9\% & 0.956 & 0.965 & 0.994 & \textbf{0.594} & \textbf{0.621} & 0.826 \\
\sys{} (0.3) & 34.46 & 0.948 & 0.030 & \textbf{2.00} & \textbf{7.55$\times$} & 41\% & \textbf{394} & \textbf{2.32$\times$} & 18.9\% & 4.7\% & 0.955 & 0.965 & 0.994 & 0.593 & 0.619 & 0.826 \\
\cmidrule(lr){1-17}
\multicolumn{17}{l}{\emph{Wan2.2-T2V-A14B (RTX PRO 6000)}} \\
Dense & -- & -- & -- & 26.05 & 1.00$\times$ & 71\% & 1477 & 1.00$\times$ & 100.0\% & -- & 0.959 & 0.963 & 0.985 & 0.626 & 0.722 & 0.851 \\
Sol-Attn & 18.98 & 0.682 & 0.248 & 7.29 & 3.58$\times$ & 41\% & 876 & 1.69$\times$ & 22.2\% & -- & 0.957 & 0.962 & 0.986 & \textbf{0.629} & 0.719 & 0.850 \\
SVG-EAR & 23.26 & 0.812 & 0.128 & 14.39 & 1.81$\times$ & 58\% & 1094 & 1.35$\times$ & 25.9\% & -- & 0.957 & 0.962 & 0.985 & 0.625 & \textbf{0.723} & \textbf{0.851} \\
SageAttn3 & 22.53 & 0.783 & 0.147 & 13.96 & 1.87$\times$ & 58\% & 1076 & 1.37$\times$ & 100.0\% & -- & 0.953 & 0.958 & 0.985 & 0.616 & 0.721 & 0.847 \\
SpargeAttn & 20.24 & 0.732 & 0.196 & 8.88 & 2.93$\times$ & 46\% & 927 & 1.59$\times$ & 50.0\% & -- & 0.955 & 0.960 & 0.986 & 0.624 & 0.721 & 0.849 \\
\sys{} (0.15) & \textbf{24.40} & \textbf{0.839} & \textbf{0.104} & 6.88 & 3.79$\times$ & 40\% & 859 & 1.72$\times$ & 35.2\% & 8.5\% & \textbf{0.958} & 0.962 & 0.985 & 0.624 & 0.721 & 0.850 \\
\sys{} (0.3) & 22.98 & 0.802 & 0.131 & \textbf{5.25} & \textbf{4.97$\times$} & 34\% & \textbf{805} & \textbf{1.83$\times$} & 23.2\% & 4.6\% & 0.957 & 0.962 & 0.985 & 0.625 & 0.721 & 0.850 \\
\cmidrule(lr){1-17}
\multicolumn{17}{l}{\emph{HunyuanVideo 720p (RTX PRO 6000)}} \\
Dense & -- & -- & -- & 29.01 & 1.00$\times$ & 85\% & 1737 & 1.00$\times$ & 100.0\% & -- & 0.959 & 0.968 & 0.994 & 0.590 & 0.659 & 0.834 \\
Sol-Attn & 28.02 & 0.888 & 0.102 & 7.63 & 3.80$\times$ & 59\% & 882 & 1.97$\times$ & 21.4\% & -- & 0.958 & 0.968 & 0.994 & \textbf{0.597} & 0.656 & \textbf{0.835} \\
SVG-EAR & 35.73 & 0.957 & 0.028 & 12.98 & 2.24$\times$ & 72\% & 1088 & 1.60$\times$ & 21.3\% & -- & 0.958 & 0.968 & 0.994 & 0.586 & 0.657 & 0.832 \\
SageAttn3 & 33.13 & 0.938 & 0.049 & 15.09 & 1.92$\times$ & 75\% & 1169 & 1.49$\times$ & 100.0\% & -- & 0.958 & 0.967 & 0.994 & 0.585 & 0.655 & 0.832 \\
SpargeAttn & 32.23 & 0.936 & 0.051 & 9.38 & 3.09$\times$ & 64\% & 950 & 1.83$\times$ & 50.0\% & -- & 0.958 & \textbf{0.969} & 0.994 & 0.588 & \textbf{0.659} & 0.833 \\
\sys{} (0.3) & \textbf{36.03} & \textbf{0.958} & \textbf{0.026} & \textbf{4.72} & \textbf{6.16$\times$} & 48\% & \textbf{761} & \textbf{2.28$\times$} & 18.6\% & 4.5\% & \textbf{0.959} & 0.968 & 0.994 & 0.595 & 0.654 & 0.834 \\
\cmidrule(lr){1-17}
\multicolumn{17}{l}{\emph{LTX-2.5 I2V (RTX PRO 6000)}} \\
Dense & -- & -- & -- & 72.82 & 1.00$\times$ & 80\% & 389 & 1.00$\times$ & 100.0\% & -- & 0.935 & 0.945 & 0.993 & 0.632 & 0.715 & 0.844 \\
Sol-Attn & 27.91 & 0.880 & 0.081 & 17.65 & 4.13$\times$ & 50\% & 223 & 1.74$\times$ & 18.5\% & -- & 0.936 & \textbf{0.947} & 0.993 & \textbf{0.638} & 0.715 & \textbf{0.846} \\
SVG-EAR & 33.30 & 0.942 & 0.036 & 58.78 & 1.24$\times$ & 77\% & 346 & 1.12$\times$ & 44.1\% & -- & 0.935 & 0.944 & 0.993 & 0.633 & 0.715 & 0.844 \\
SageAttn3 & 33.12 & 0.940 & 0.037 & 38.68 & 1.88$\times$ & 69\% & 286 & 1.36$\times$ & 100.0\% & -- & 0.935 & 0.943 & 0.993 & 0.632 & 0.714 & 0.843 \\
SpargeAttn & 29.68 & 0.906 & 0.060 & 24.04 & 3.03$\times$ & 57\% & 242 & 1.60$\times$ & 50.0\% & -- & 0.936 & 0.944 & 0.993 & 0.635 & 0.714 & 0.844 \\
\sys{} (0.3) & \textbf{34.12} & \textbf{0.947} & \textbf{0.031} & \textbf{13.94} & \textbf{5.22$\times$} & 44\% & \textbf{211} & \textbf{1.84$\times$} & 27.4\% & 3.4\% & 0.935 & 0.945 & 0.993 & 0.633 & 0.714 & 0.844 \\
\bottomrule
\end{tabular}

\end{table*}

\subsection{The Router Score}
\label{sec:app-router}
This subsection lists every term of the router score for one head $h$ and one query block $I$ of $|I| = 128$ queries.
Tile $t$ consists of two key clusters $\beta \in t$ of at most 64 keys, with $n_\beta$ keys each, mean key $\bar{k}_\beta$, and mean value $\bar{v}_\beta$; $D$ is the head dimension, and $\Nt$ is the number of tiles in the row.

\paragraph{Query and tile statistics.}
The query block contributes its mean query $\bar{q}$, its root-mean-square norm $\|q\| = \big(\frac{1}{|I|} \sum_{i \in I} \|q_i\|^2\big)^{1/2}$, and its variance $\sigma_q^2 = \frac{1}{|I|} \sum_{i \in I} \|q_i - \bar{q}\|^2$.
For each cluster $\beta$, $\operatorname{tr}(\Sigma_{K,\beta}) = \frac{1}{n_\beta} \sum_{j \in \beta} \|k_j - \bar{k}_\beta\|^2$ and $\|\Sigma_{KV,\beta}\|_F = \frac{1}{n_\beta} \|\tilde{K}_\beta^\top \tilde{V}_\beta\|_F$, where $\tilde{K}_\beta$ and $\tilde{V}_\beta$ are the centered keys and values of the cluster.
With weights $\omega_\beta = n_\beta/\sum_{\beta'\in t}n_{\beta'}$, the tile's key trace is $\langle\operatorname{tr}\Sigma_K\rangle_t = \sum_{\beta\in t} \omega_\beta \operatorname{tr}(\Sigma_{K,\beta})$, as in \secref{sec:design-router}.
The covariances retain only variation within each cluster, since each cluster has its own mean representative; between-cluster variation is not discarded by skipping.

\paragraph{Error factor.}
\begin{align*}
  J_t &= 1 - \exp\!\big(-\|q\|^2 \langle\operatorname{tr}\Sigma_K\rangle_t / (2D^2)\big), \\
  R_t &= \frac{\|q\|}{D} \sum_{\beta\in t} \omega_\beta \frac{\|\Sigma_{KV,\beta}\|_F}{\|\bar{v}_\beta\|}.
\end{align*}

\paragraph{Head weight and score.}
With $\overline{\|O_h\|^2}$ the mean squared output norm of head $h$ at the previous call (at the first call, $\|\frac{1}{N} \sum_j v_j\|^2$ over the $N$ keys of the current call) and $N_{\mathrm{h}}$ the number of heads in the layer,
\begin{equation*}
  w_h = \frac{\overline{\|O_h\|^2}}{\frac{1}{N_{\mathrm{h}}} \sum_{h'} \overline{\|O_{h'}\|^2}}, \qquad
  \tilde{E}_t = \tilde{m}_t\, (J_t + R_t)\, \sqrt{w_h}.
\end{equation*}
Since $E_t$ is relative to the output of head $h$, the factor $\sqrt{w_h}$ converts it to a common scale across the heads of the layer: with $e_h = \|\Delta O_h\|/\|O_h\|$ and $w_h$ formed from the same call's $\|O_h\|^2$, $\sum_h \|\Delta O_h\|^2 / \sum_h \|O_h\|^2 = \frac{1}{N_{\mathrm{h}}}\sum_h w_h e_h^2$; the router uses the previous call's $w_h$ as a proxy.

\begin{figure}[t]
  \centering
  \includegraphics[width=\columnwidth]{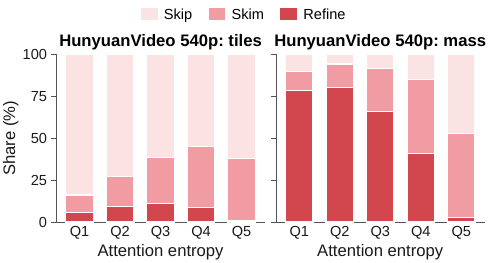}
  \caption{Tiers that the router assigns at $\pi = 0.15$ on HunyuanVideo 540p, as \figref{fig:eval-tiers} for Wan2.2-TI2V-5B.}
  \label{fig:app-tiers}
\end{figure}
\section{Additional Results}
\label{sec:appendix}

\subsection{End-to-End Details}
\label{sec:app-e2e}
\tabref{tab:app-e2e} lists every metric of the end-to-end runs: fidelity to the dense video, absolute time and speedup, the share of tiles each method computes, and every VBench dimension; the Share column gives the share of a step spent in attention, which for the dense run bounds the end-to-end speedup.

\subsection{Tiers Across Attention Entropy}
\label{sec:app-tiers}

\figref{fig:app-tiers} repeats \figref{fig:eval-tiers} for HunyuanVideo 540p.
Every quintile again contains all three tiers, and in the sharpest rows (Q1) the refined tiles hold 79\% of the mass.

\subsection{Qualitative Results on Wan2.2 and HunyuanVideo}
\label{sec:app-qual-t2v}
Figures~\ref{fig:app-qual-wan} and~\ref{fig:app-qual-hunyuan} show videos of Wan2.2-T2V-A14B ($\pi = 0.15$) and HunyuanVideo ($\pi = 0.3$) at 720p with dense attention and with \sys, from the same prompt and seed, as six evenly spaced frames; each video is labeled with its generation time, speedup, and PSNR against the dense video.

\makeatletter\global\@dblfptop=0pt\relax\global\@dblfpbot=0pt plus 1fil\relax\makeatother
\begin{figure*}[p]
\centering
\ifx\qualw\undefined\newlength{\qualw}\fi
\setlength{\qualw}{\dimexpr(\textwidth - 12.5pt)*84/600\relax}%
\makebox[\textwidth][c]{
\begin{tabular}{@{}c@{\hspace{1.5pt}}c@{\hspace{1pt}}c@{\hspace{1pt}}c@{\hspace{1pt}}c@{\hspace{1pt}}c@{\hspace{1pt}}c@{}}
\multicolumn{7}{@{}l@{}}{\makebox[\dimexpr6\qualw+5pt\relax][l]{\scriptsize\textbf{720p}, 5.1\,s\quad\textit{chessboard in sunlight}\hfill 1468\,s\,$\rightarrow$\,861\,s (1.70\x), 24.2\,dB}}\\[1pt]
\multicolumn{7}{@{}p{\dimexpr6\qualw+5pt\relax}@{}}{\tiny\raggedright\textit{Prompt (seed 42):} Top-down shot of a wooden chessboard in the middle of a game, the polished pieces casting long shadows in warm afternoon light as the camera slowly rotates.}\\[2pt]
 & \tiny $t{=}0.5$\,s & \tiny $t{=}1.3$\,s & \tiny $t{=}2.1$\,s & \tiny $t{=}2.9$\,s & \tiny $t{=}3.7$\,s & \tiny $t{=}4.5$\,s\\[1pt]
\raisebox{-.5\height}{\rotatebox{90}{\tiny Dense}} & \raisebox{-.5\height}{\includegraphics[width=\qualw]{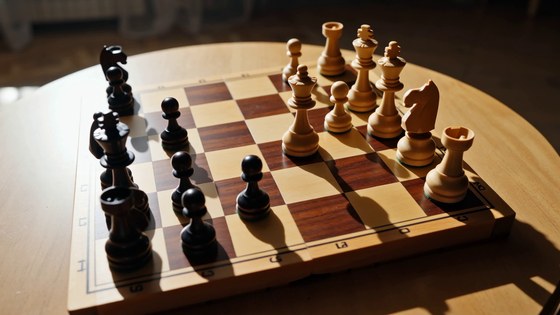}} & \raisebox{-.5\height}{\includegraphics[width=\qualw]{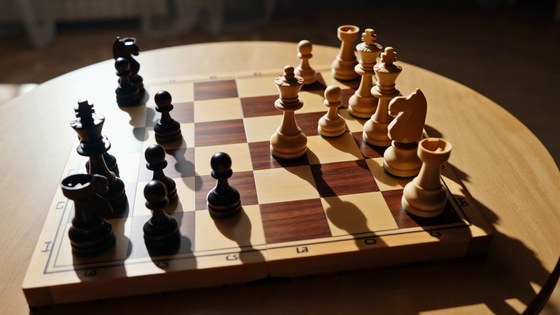}} & \raisebox{-.5\height}{\includegraphics[width=\qualw]{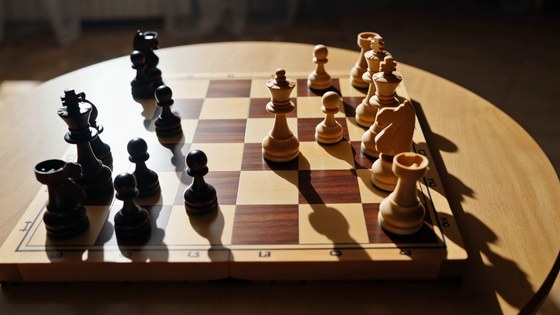}} & \raisebox{-.5\height}{\includegraphics[width=\qualw]{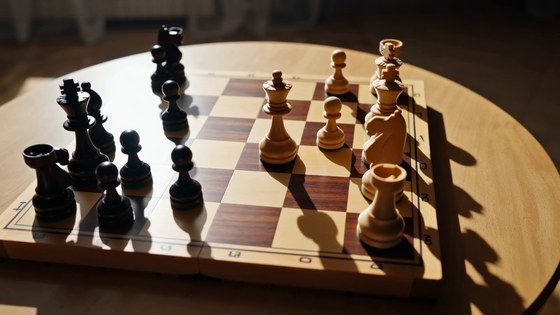}} & \raisebox{-.5\height}{\includegraphics[width=\qualw]{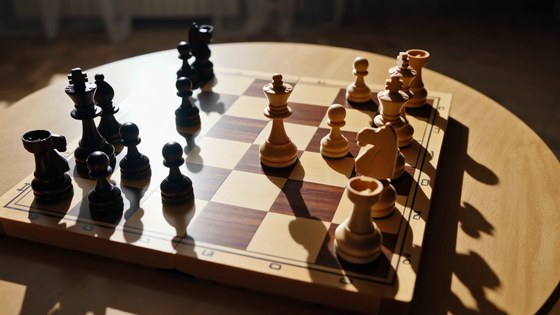}} & \raisebox{-.5\height}{\includegraphics[width=\qualw]{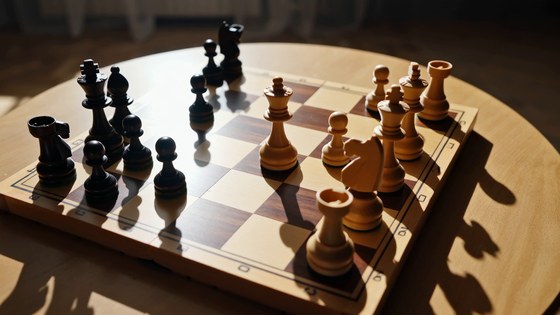}}\\[1pt]
\raisebox{-.5\height}{\rotatebox{90}{\tiny \sys}} & \raisebox{-.5\height}{\includegraphics[width=\qualw]{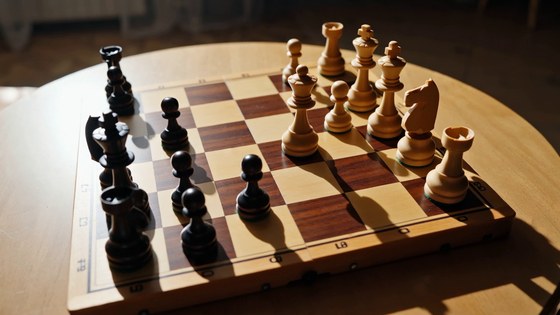}} & \raisebox{-.5\height}{\includegraphics[width=\qualw]{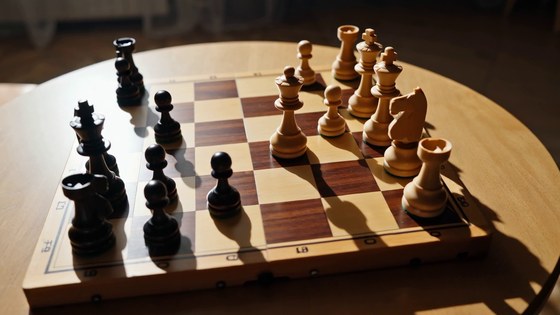}} & \raisebox{-.5\height}{\includegraphics[width=\qualw]{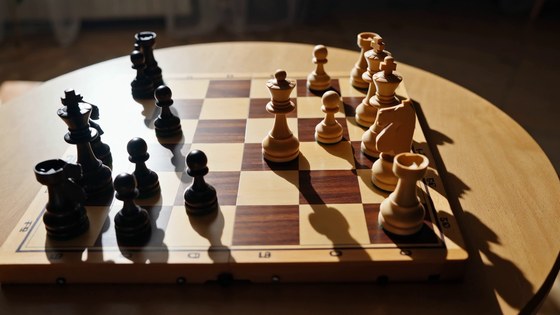}} & \raisebox{-.5\height}{\includegraphics[width=\qualw]{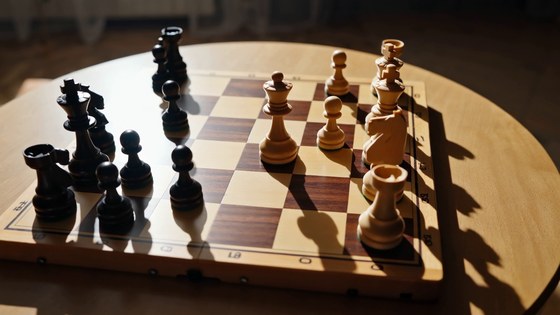}} & \raisebox{-.5\height}{\includegraphics[width=\qualw]{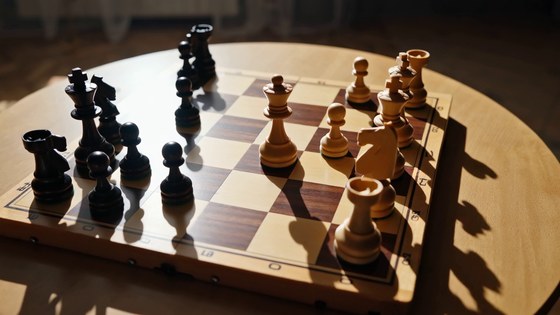}} & \raisebox{-.5\height}{\includegraphics[width=\qualw]{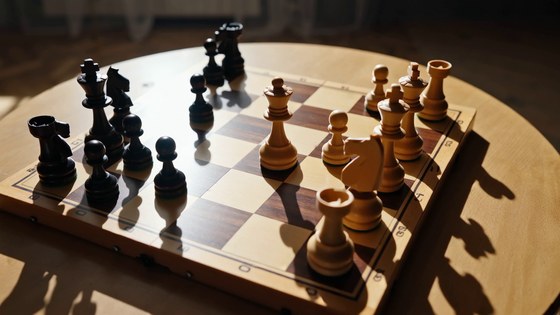}}\\
\end{tabular}}
\par\vspace{3pt}
\makebox[\textwidth][c]{\begin{tabular}{@{}c@{\hspace{1.5pt}}c@{\hspace{1pt}}c@{\hspace{1pt}}c@{\hspace{1pt}}c@{\hspace{1pt}}c@{\hspace{1pt}}c@{}}
\multicolumn{7}{@{}l@{}}{\makebox[\dimexpr6\qualw+5pt\relax][l]{\scriptsize\textbf{720p}, 5.1\,s\quad\textit{cat at a rainy window}\hfill 1469\,s\,$\rightarrow$\,858\,s (1.71\x), 32.3\,dB}}\\[1pt]
\multicolumn{7}{@{}p{\dimexpr6\qualw+5pt\relax}@{}}{\tiny\raggedright\textit{Prompt (seed 42):} A calico cat sits on a sunny windowsill, its tail swaying slowly, watching raindrops run down the glass.}\\[2pt]
 & \tiny $t{=}0.5$\,s & \tiny $t{=}1.3$\,s & \tiny $t{=}2.1$\,s & \tiny $t{=}2.9$\,s & \tiny $t{=}3.7$\,s & \tiny $t{=}4.5$\,s\\[1pt]
\raisebox{-.5\height}{\rotatebox{90}{\tiny Dense}} & \raisebox{-.5\height}{\includegraphics[width=\qualw]{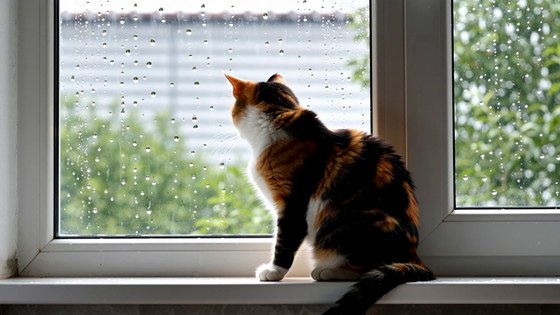}} & \raisebox{-.5\height}{\includegraphics[width=\qualw]{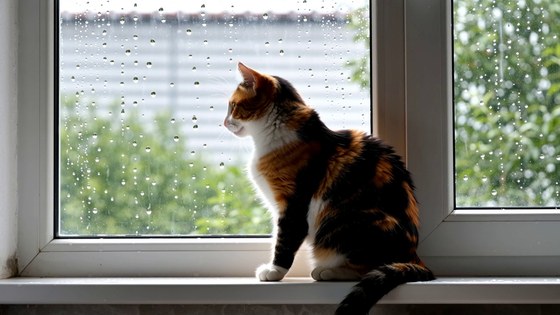}} & \raisebox{-.5\height}{\includegraphics[width=\qualw]{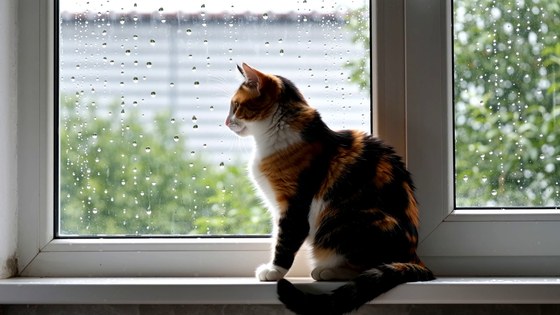}} & \raisebox{-.5\height}{\includegraphics[width=\qualw]{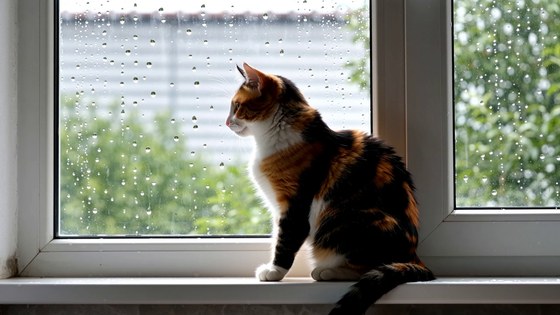}} & \raisebox{-.5\height}{\includegraphics[width=\qualw]{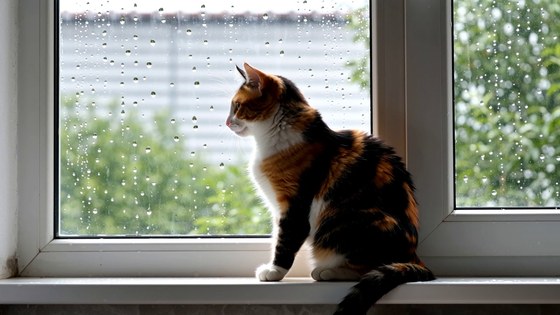}} & \raisebox{-.5\height}{\includegraphics[width=\qualw]{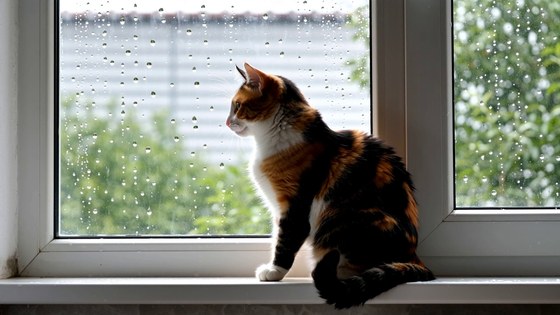}}\\[1pt]
\raisebox{-.5\height}{\rotatebox{90}{\tiny \sys}} & \raisebox{-.5\height}{\includegraphics[width=\qualw]{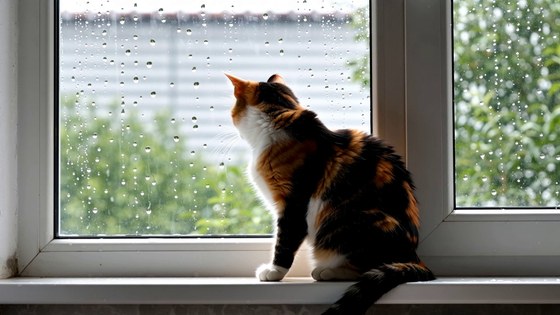}} & \raisebox{-.5\height}{\includegraphics[width=\qualw]{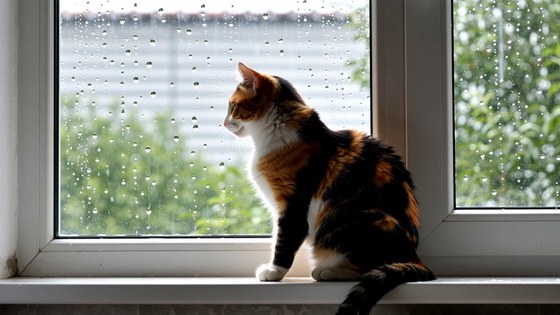}} & \raisebox{-.5\height}{\includegraphics[width=\qualw]{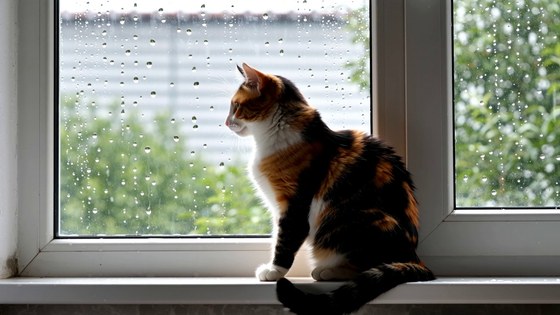}} & \raisebox{-.5\height}{\includegraphics[width=\qualw]{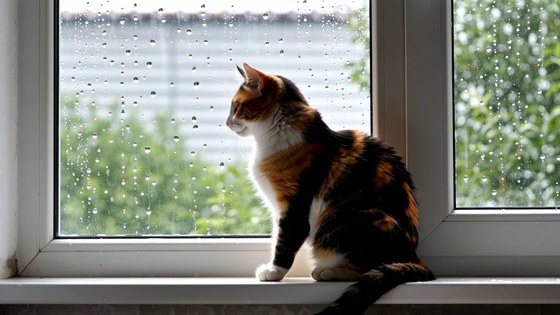}} & \raisebox{-.5\height}{\includegraphics[width=\qualw]{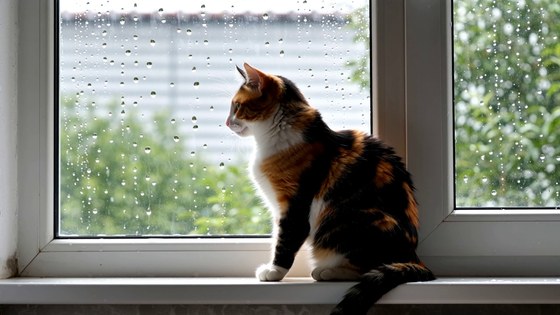}} & \raisebox{-.5\height}{\includegraphics[width=\qualw]{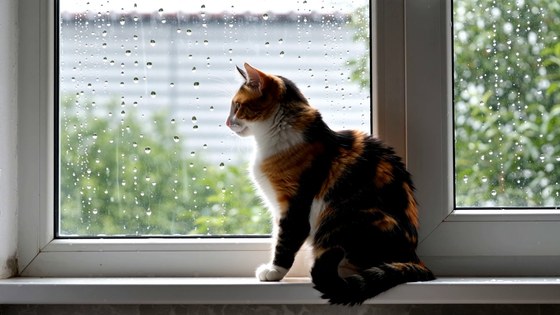}}\\
\end{tabular}}
\par\vspace{3pt}
\makebox[\textwidth][c]{\begin{tabular}{@{}c@{\hspace{1.5pt}}c@{\hspace{1pt}}c@{\hspace{1pt}}c@{\hspace{1pt}}c@{\hspace{1pt}}c@{\hspace{1pt}}c@{}}
\multicolumn{7}{@{}l@{}}{\makebox[\dimexpr6\qualw+5pt\relax][l]{\scriptsize\textbf{720p}, 5.1\,s\quad\textit{dog catching a frisbee}\hfill 1467\,s\,$\rightarrow$\,850\,s (1.72\x), 23.8\,dB}}\\[1pt]
\multicolumn{7}{@{}p{\dimexpr6\qualw+5pt\relax}@{}}{\tiny\raggedright\textit{Prompt (seed 42):} A golden retriever leaps to catch a red frisbee in a sunny park, grass and fallen leaves flying around it.}\\[2pt]
 & \tiny $t{=}0.5$\,s & \tiny $t{=}1.3$\,s & \tiny $t{=}2.1$\,s & \tiny $t{=}2.9$\,s & \tiny $t{=}3.7$\,s & \tiny $t{=}4.5$\,s\\[1pt]
\raisebox{-.5\height}{\rotatebox{90}{\tiny Dense}} & \raisebox{-.5\height}{\includegraphics[width=\qualw]{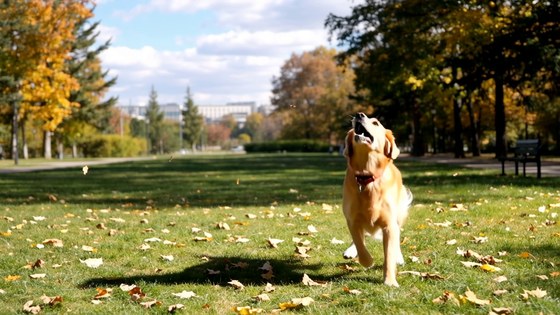}} & \raisebox{-.5\height}{\includegraphics[width=\qualw]{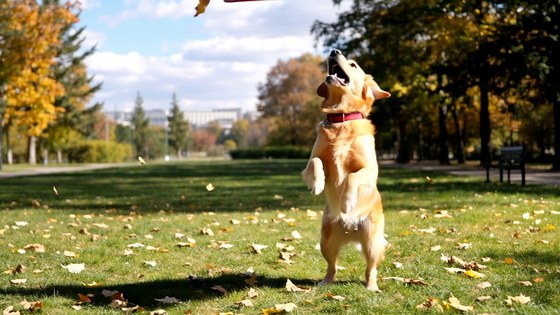}} & \raisebox{-.5\height}{\includegraphics[width=\qualw]{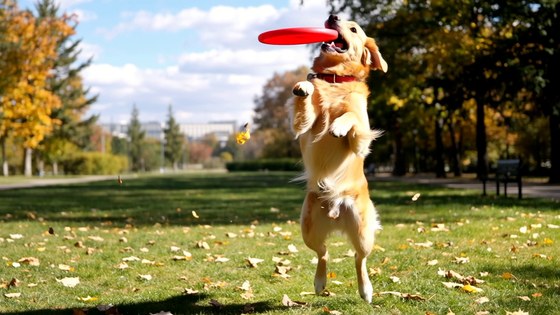}} & \raisebox{-.5\height}{\includegraphics[width=\qualw]{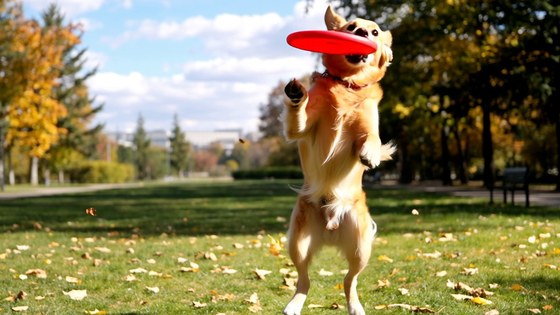}} & \raisebox{-.5\height}{\includegraphics[width=\qualw]{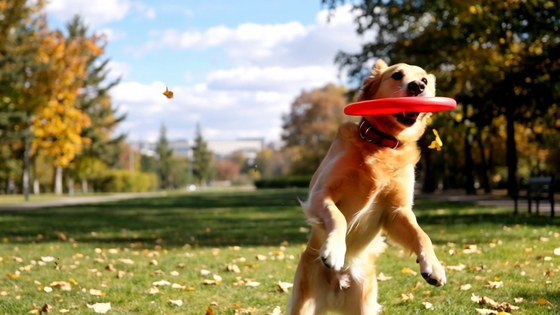}} & \raisebox{-.5\height}{\includegraphics[width=\qualw]{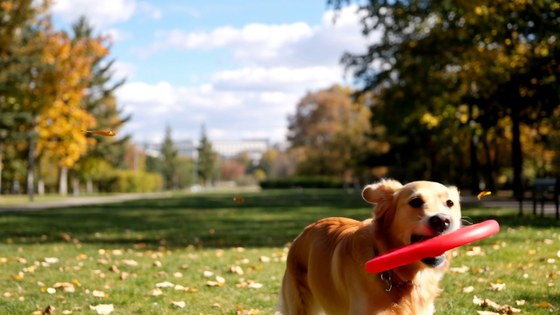}}\\[1pt]
\raisebox{-.5\height}{\rotatebox{90}{\tiny \sys}} & \raisebox{-.5\height}{\includegraphics[width=\qualw]{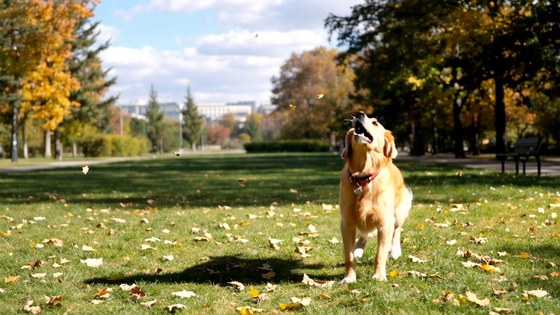}} & \raisebox{-.5\height}{\includegraphics[width=\qualw]{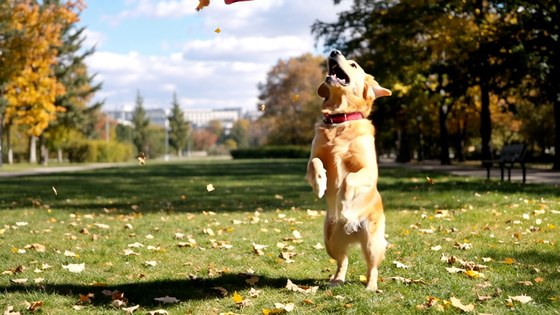}} & \raisebox{-.5\height}{\includegraphics[width=\qualw]{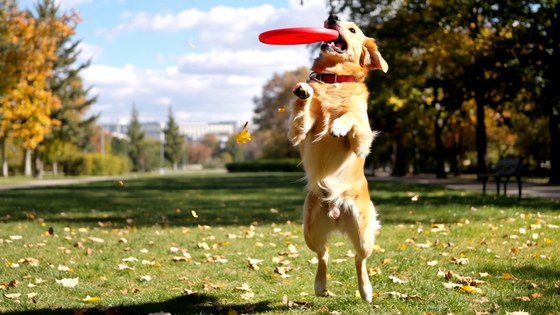}} & \raisebox{-.5\height}{\includegraphics[width=\qualw]{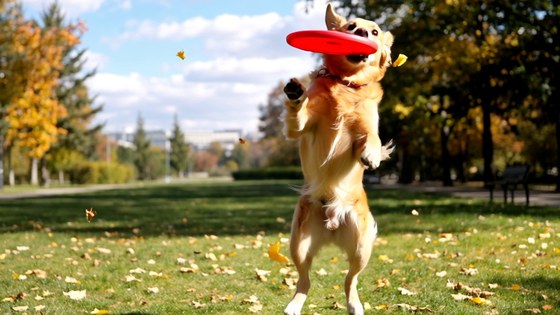}} & \raisebox{-.5\height}{\includegraphics[width=\qualw]{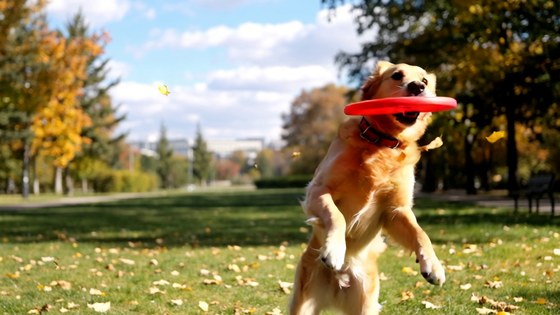}} & \raisebox{-.5\height}{\includegraphics[width=\qualw]{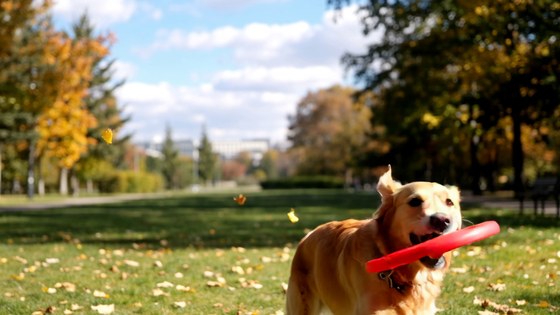}}\\
\end{tabular}}
\par\vspace{3pt}
\makebox[\textwidth][c]{\begin{tabular}{@{}c@{\hspace{1.5pt}}c@{\hspace{1pt}}c@{\hspace{1pt}}c@{\hspace{1pt}}c@{\hspace{1pt}}c@{\hspace{1pt}}c@{}}
\multicolumn{7}{@{}l@{}}{\makebox[\dimexpr6\qualw+5pt\relax][l]{\scriptsize\textbf{720p}, 5.1\,s\quad\textit{corgi on a sunset beach}\hfill 1469\,s\,$\rightarrow$\,850\,s (1.73\x), 27.3\,dB}}\\[1pt]
\multicolumn{7}{@{}p{\dimexpr6\qualw+5pt\relax}@{}}{\tiny\raggedright\textit{Prompt (seed 42):} A corgi runs along the beach at sunset, splashing through the shallow waves.}\\[2pt]
 & \tiny $t{=}0.5$\,s & \tiny $t{=}1.3$\,s & \tiny $t{=}2.1$\,s & \tiny $t{=}2.9$\,s & \tiny $t{=}3.7$\,s & \tiny $t{=}4.5$\,s\\[1pt]
\raisebox{-.5\height}{\rotatebox{90}{\tiny Dense}} & \raisebox{-.5\height}{\includegraphics[width=\qualw]{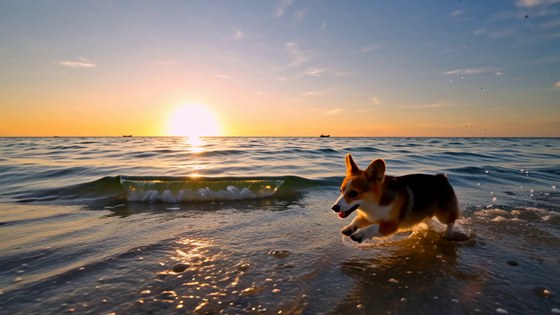}} & \raisebox{-.5\height}{\includegraphics[width=\qualw]{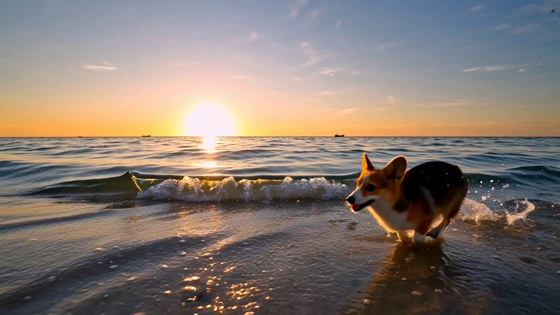}} & \raisebox{-.5\height}{\includegraphics[width=\qualw]{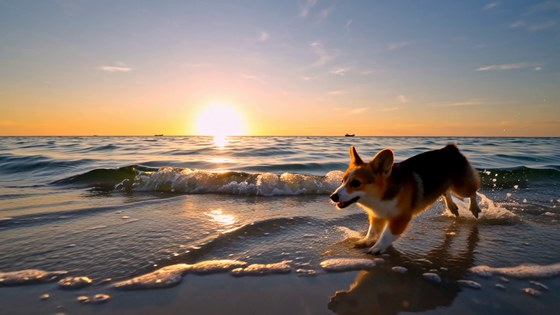}} & \raisebox{-.5\height}{\includegraphics[width=\qualw]{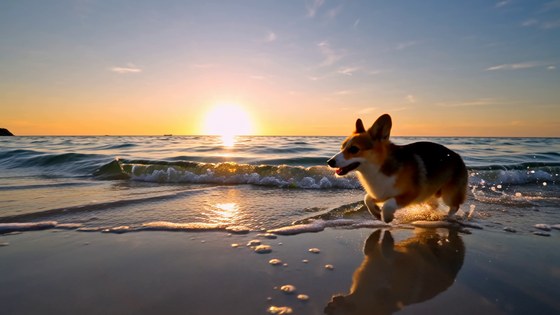}} & \raisebox{-.5\height}{\includegraphics[width=\qualw]{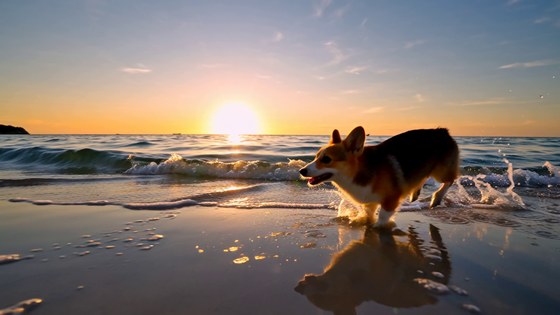}} & \raisebox{-.5\height}{\includegraphics[width=\qualw]{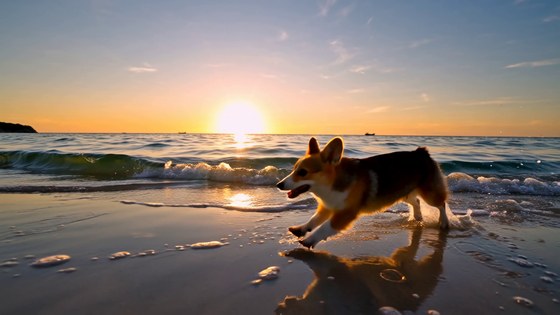}}\\[1pt]
\raisebox{-.5\height}{\rotatebox{90}{\tiny \sys}} & \raisebox{-.5\height}{\includegraphics[width=\qualw]{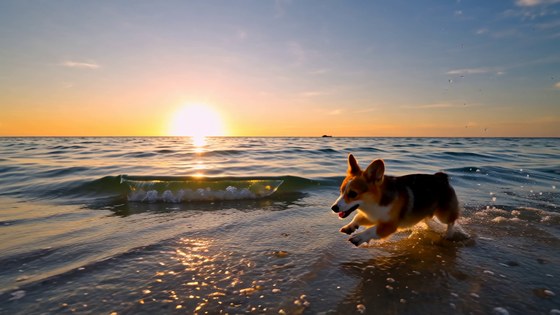}} & \raisebox{-.5\height}{\includegraphics[width=\qualw]{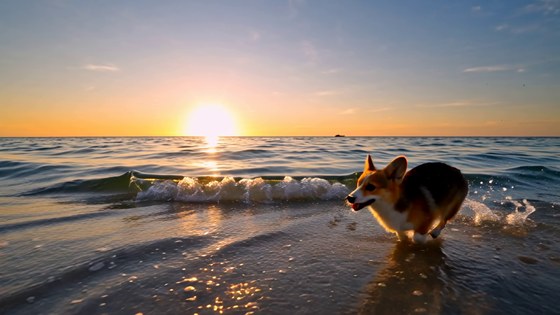}} & \raisebox{-.5\height}{\includegraphics[width=\qualw]{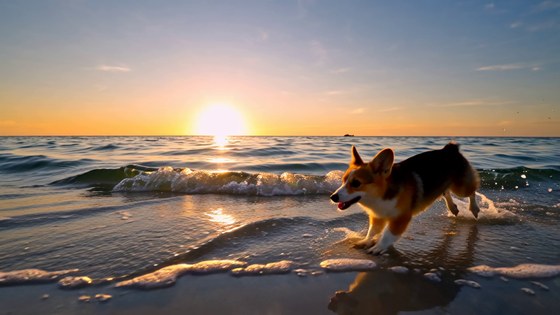}} & \raisebox{-.5\height}{\includegraphics[width=\qualw]{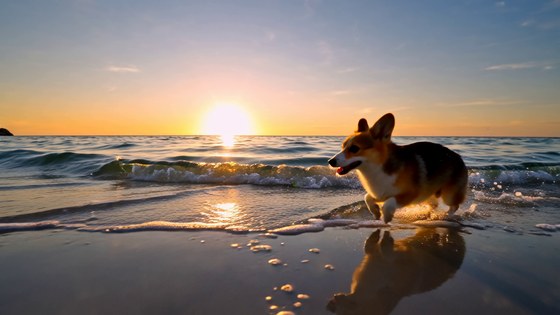}} & \raisebox{-.5\height}{\includegraphics[width=\qualw]{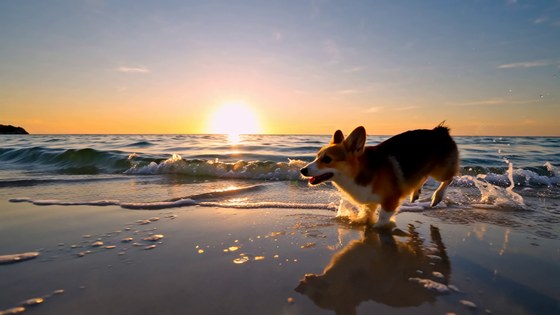}} & \raisebox{-.5\height}{\includegraphics[width=\qualw]{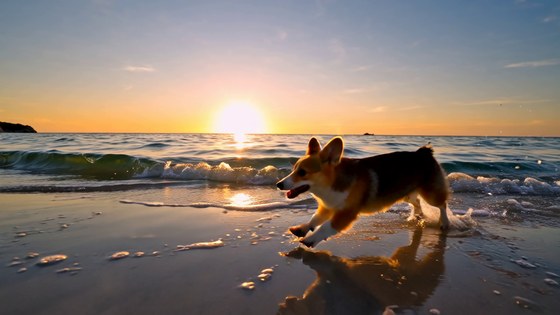}}\\
\end{tabular}}
\par\vspace{3pt}
\makebox[\textwidth][c]{\begin{tabular}{@{}c@{\hspace{1.5pt}}c@{\hspace{1pt}}c@{\hspace{1pt}}c@{\hspace{1pt}}c@{\hspace{1pt}}c@{\hspace{1pt}}c@{}}
\multicolumn{7}{@{}l@{}}{\makebox[\dimexpr6\qualw+5pt\relax][l]{\scriptsize\textbf{720p}, 5.1\,s\quad\textit{palace rising from the sea}\hfill 1479\,s\,$\rightarrow$\,853\,s (1.73\x), 27.2\,dB}}\\[1pt]
\multicolumn{7}{@{}p{\dimexpr6\qualw+5pt\relax}@{}}{\tiny\raggedright\textit{Prompt (seed 1040):} A magnificent palace emerges gradually from the shimmering sea, as if a mysterious force from the ocean depths gently lifts it. The palace's spires gleam with golden light in the setting sun, resembling a mirage, and is breathtakingly beautiful.}\\[2pt]
 & \tiny $t{=}0.5$\,s & \tiny $t{=}1.3$\,s & \tiny $t{=}2.1$\,s & \tiny $t{=}2.9$\,s & \tiny $t{=}3.7$\,s & \tiny $t{=}4.5$\,s\\[1pt]
\raisebox{-.5\height}{\rotatebox{90}{\tiny Dense}} & \raisebox{-.5\height}{\includegraphics[width=\qualw]{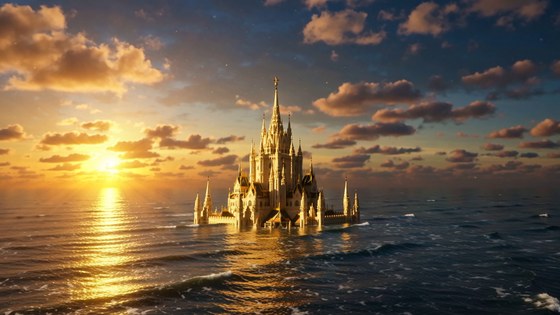}} & \raisebox{-.5\height}{\includegraphics[width=\qualw]{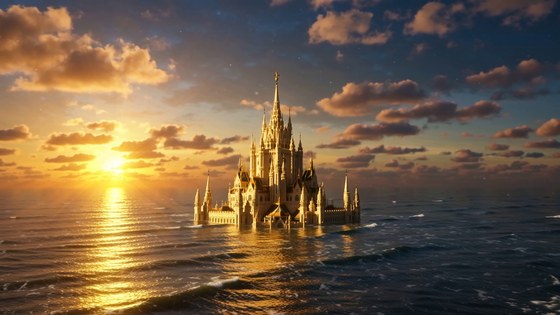}} & \raisebox{-.5\height}{\includegraphics[width=\qualw]{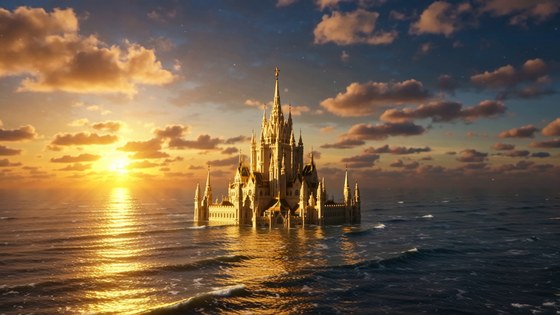}} & \raisebox{-.5\height}{\includegraphics[width=\qualw]{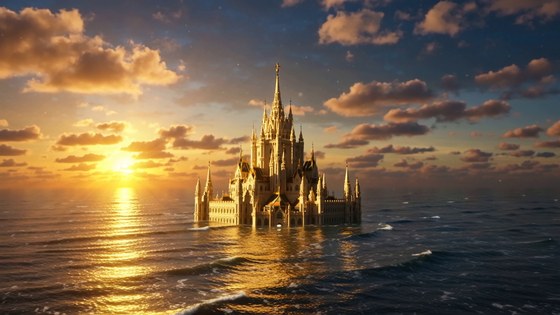}} & \raisebox{-.5\height}{\includegraphics[width=\qualw]{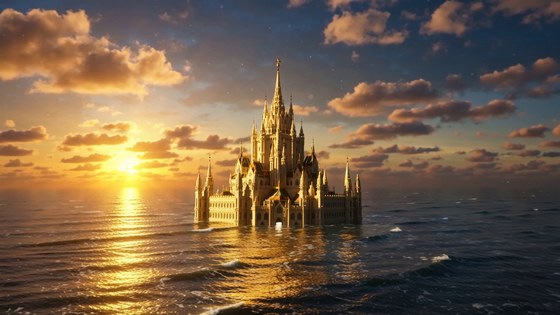}} & \raisebox{-.5\height}{\includegraphics[width=\qualw]{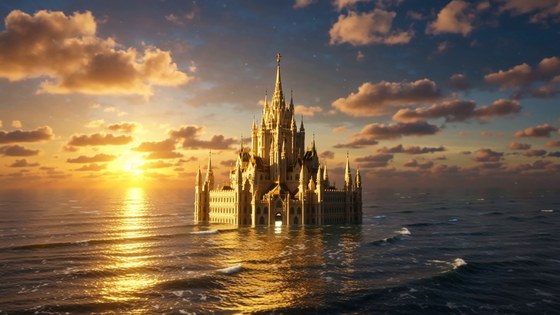}}\\
\end{tabular}}
\caption{Wan2.2-T2V-A14B at 720p (81 frames) on the RTX PRO 6000 with dense attention and with \sys ($\pi = 0.15$), from the same prompt and seed: six evenly spaced frames per video, with the generation time (dense $\rightarrow$ \sys), speedup, and PSNR against the dense video.}
\label{fig:app-qual-wan}
\end{figure*}

\begin{figure*}[p]
\centering
\ifx\qualw\undefined\newlength{\qualw}\fi
\setlength{\qualw}{\dimexpr(\textwidth - 12.5pt)*84/600\relax}%
\makebox[\textwidth][c]{\begin{tabular}{@{}c@{\hspace{1.5pt}}c@{\hspace{1pt}}c@{\hspace{1pt}}c@{\hspace{1pt}}c@{\hspace{1pt}}c@{\hspace{1pt}}c@{}}
\multicolumn{7}{@{}l@{}}{\makebox[\dimexpr6\qualw+5pt\relax][l]{\scriptsize\textbf{720p}, 5.4\,s\quad\textit{fisherman at dusk}\hfill 1734\,s\,$\rightarrow$\,760\,s (2.28\x), 35.8\,dB}}\\[1pt]
\multicolumn{7}{@{}p{\dimexpr6\qualw+5pt\relax}@{}}{\tiny\raggedright\textit{Prompt (seed 1100):} A close-up of an elderly fisherman with a weathered face and a white beard, mending a fishing net on a wooden pier at sunset. Warm golden light, cinematic, shallow depth of field.}\\[2pt]
 & \tiny $t{=}0.5$\,s & \tiny $t{=}1.4$\,s & \tiny $t{=}2.2$\,s & \tiny $t{=}3.1$\,s & \tiny $t{=}4.0$\,s & \tiny $t{=}4.8$\,s\\[1pt]
\raisebox{-.5\height}{\rotatebox{90}{\tiny Dense}} & \raisebox{-.5\height}{\includegraphics[width=\qualw]{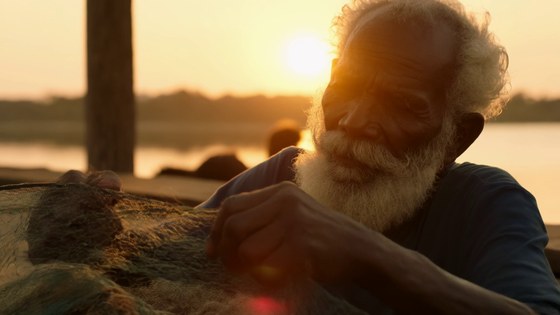}} & \raisebox{-.5\height}{\includegraphics[width=\qualw]{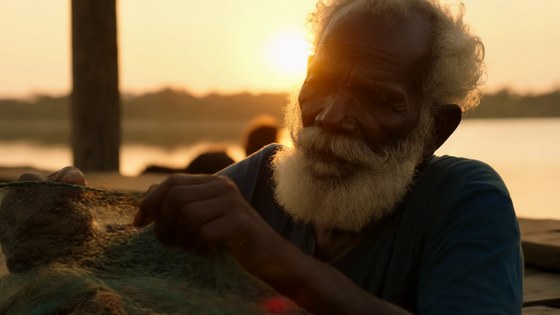}} & \raisebox{-.5\height}{\includegraphics[width=\qualw]{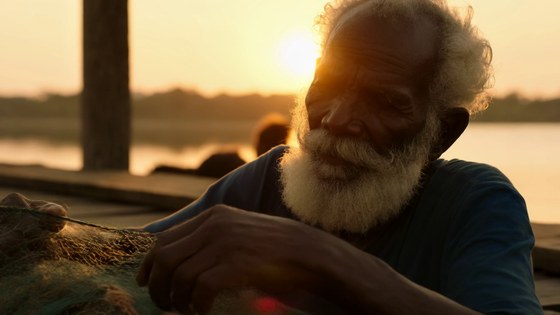}} & \raisebox{-.5\height}{\includegraphics[width=\qualw]{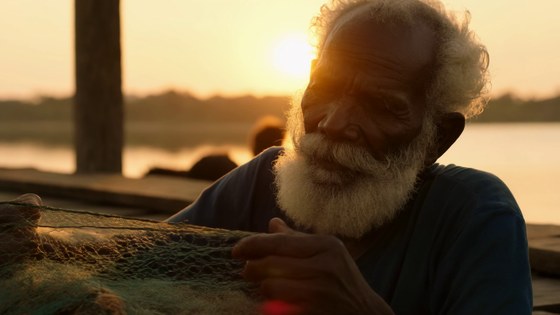}} & \raisebox{-.5\height}{\includegraphics[width=\qualw]{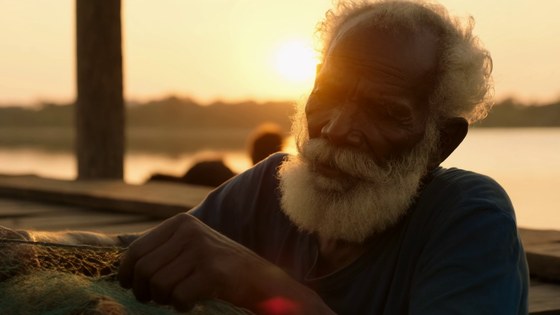}} & \raisebox{-.5\height}{\includegraphics[width=\qualw]{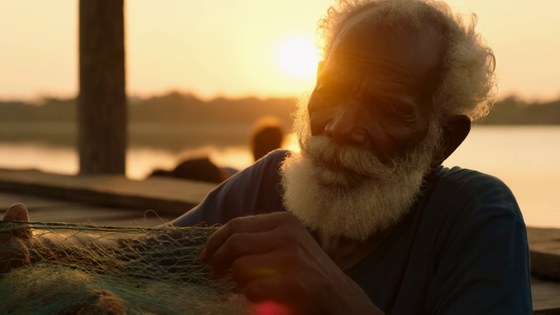}}\\
\end{tabular}}
\par\vspace{3pt}
\makebox[\textwidth][c]{\begin{tabular}{@{}c@{\hspace{1.5pt}}c@{\hspace{1pt}}c@{\hspace{1pt}}c@{\hspace{1pt}}c@{\hspace{1pt}}c@{\hspace{1pt}}c@{}}
\multicolumn{7}{@{}l@{}}{\makebox[\dimexpr6\qualw+5pt\relax][l]{\scriptsize\textbf{720p}, 5.4\,s\quad\textit{watch movement}\hfill 1739\,s\,$\rightarrow$\,775\,s (2.25\x), 34.3\,dB}}\\[1pt]
\multicolumn{7}{@{}p{\dimexpr6\qualw+5pt\relax}@{}}{\tiny\raggedright\textit{Prompt (seed 1102):} Macro shot of the inside of a mechanical watch. Polished brass gears and a balance wheel turn steadily under warm studio light. Extremely detailed.}\\[2pt]
 & \tiny $t{=}0.5$\,s & \tiny $t{=}1.4$\,s & \tiny $t{=}2.2$\,s & \tiny $t{=}3.1$\,s & \tiny $t{=}4.0$\,s & \tiny $t{=}4.8$\,s\\[1pt]
\raisebox{-.5\height}{\rotatebox{90}{\tiny Dense}} & \raisebox{-.5\height}{\includegraphics[width=\qualw]{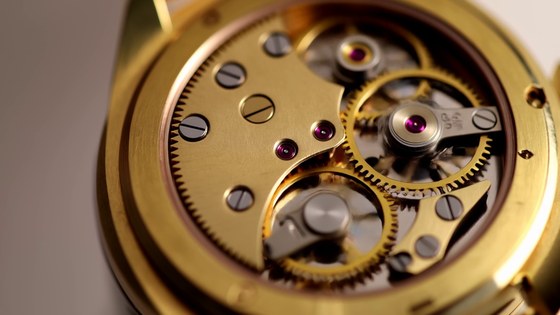}} & \raisebox{-.5\height}{\includegraphics[width=\qualw]{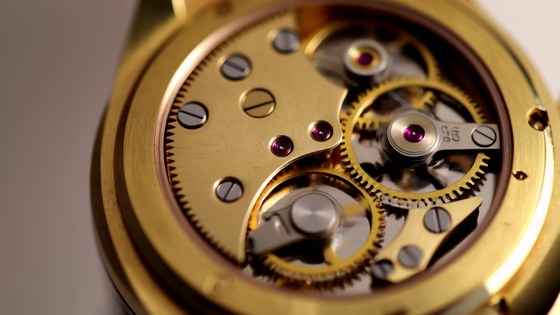}} & \raisebox{-.5\height}{\includegraphics[width=\qualw]{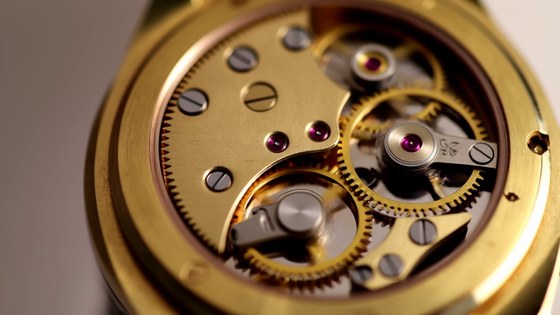}} & \raisebox{-.5\height}{\includegraphics[width=\qualw]{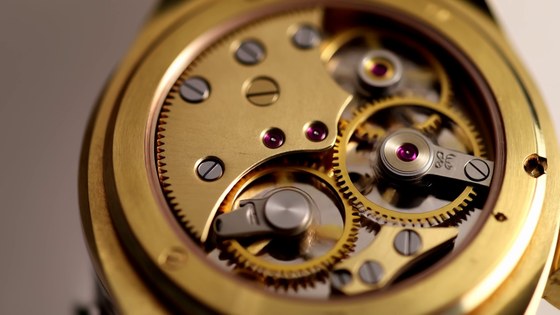}} & \raisebox{-.5\height}{\includegraphics[width=\qualw]{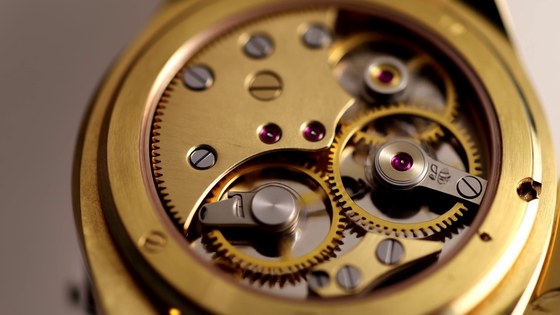}} & \raisebox{-.5\height}{\includegraphics[width=\qualw]{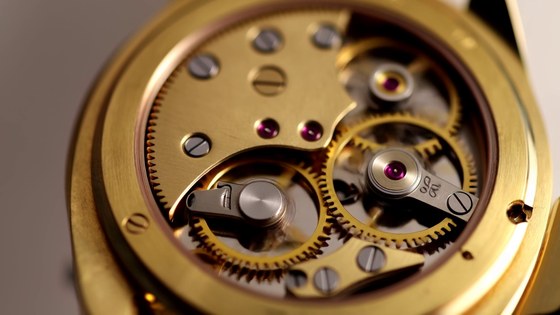}}\\
\end{tabular}}
\par\vspace{3pt}
\makebox[\textwidth][c]{\begin{tabular}{@{}c@{\hspace{1.5pt}}c@{\hspace{1pt}}c@{\hspace{1pt}}c@{\hspace{1pt}}c@{\hspace{1pt}}c@{\hspace{1pt}}c@{}}
\multicolumn{7}{@{}l@{}}{\makebox[\dimexpr6\qualw+5pt\relax][l]{\scriptsize\textbf{720p}, 5.4\,s\quad\textit{temple in the snow}\hfill 1745\,s\,$\rightarrow$\,772\,s (2.26\x), 37.0\,dB}}\\[1pt]
\multicolumn{7}{@{}p{\dimexpr6\qualw+5pt\relax}@{}}{\tiny\raggedright\textit{Prompt (seed 1104):} A woman in a dark kimono holding a red umbrella walks past a snow-covered temple with a red torii gate while snow falls gently. Cinematic.}\\[2pt]
 & \tiny $t{=}0.5$\,s & \tiny $t{=}1.4$\,s & \tiny $t{=}2.2$\,s & \tiny $t{=}3.1$\,s & \tiny $t{=}4.0$\,s & \tiny $t{=}4.8$\,s\\[1pt]
\raisebox{-.5\height}{\rotatebox{90}{\tiny Dense}} & \raisebox{-.5\height}{\includegraphics[width=\qualw]{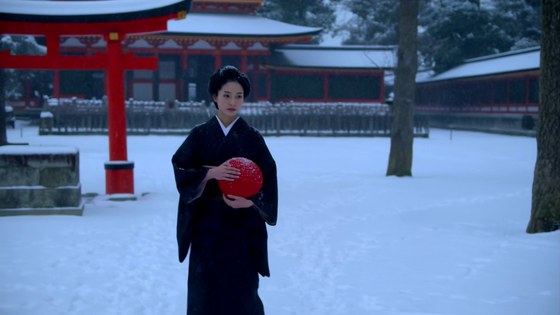}} & \raisebox{-.5\height}{\includegraphics[width=\qualw]{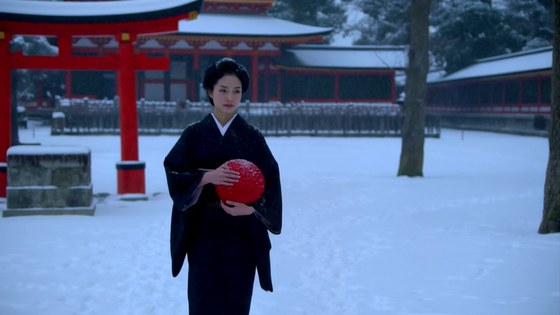}} & \raisebox{-.5\height}{\includegraphics[width=\qualw]{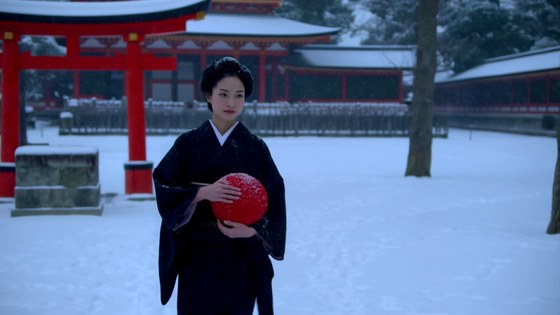}} & \raisebox{-.5\height}{\includegraphics[width=\qualw]{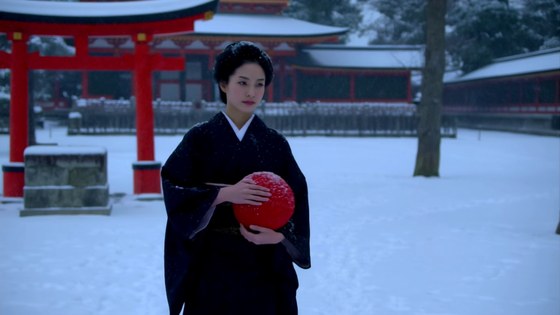}} & \raisebox{-.5\height}{\includegraphics[width=\qualw]{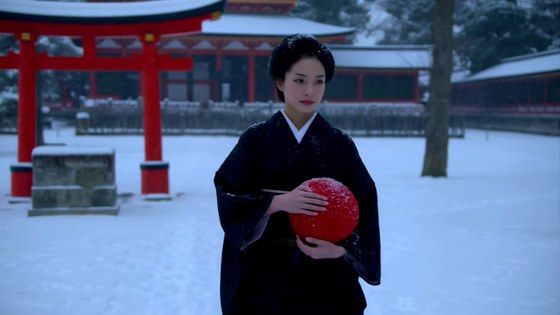}} & \raisebox{-.5\height}{\includegraphics[width=\qualw]{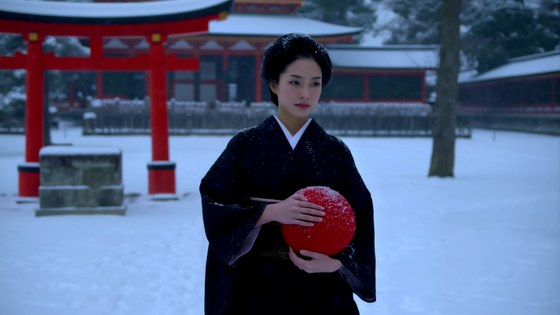}}\\
\end{tabular}}
\par\vspace{3pt}
\makebox[\textwidth][c]{\begin{tabular}{@{}c@{\hspace{1.5pt}}c@{\hspace{1pt}}c@{\hspace{1pt}}c@{\hspace{1pt}}c@{\hspace{1pt}}c@{\hspace{1pt}}c@{}}
\multicolumn{7}{@{}l@{}}{\makebox[\dimexpr6\qualw+5pt\relax][l]{\scriptsize\textbf{720p}, 5.4\,s\quad\textit{violinist on stage}\hfill 1744\,s\,$\rightarrow$\,770\,s (2.26\x), 35.2\,dB}}\\[1pt]
\multicolumn{7}{@{}p{\dimexpr6\qualw+5pt\relax}@{}}{\tiny\raggedright\textit{Prompt (seed 1105):} A violinist in a black dress performs on a dark stage under a single spotlight, the polished wood of the violin glowing. Slow dolly-in shot.}\\[2pt]
 & \tiny $t{=}0.5$\,s & \tiny $t{=}1.4$\,s & \tiny $t{=}2.2$\,s & \tiny $t{=}3.1$\,s & \tiny $t{=}4.0$\,s & \tiny $t{=}4.8$\,s\\[1pt]
\raisebox{-.5\height}{\rotatebox{90}{\tiny Dense}} & \raisebox{-.5\height}{\includegraphics[width=\qualw]{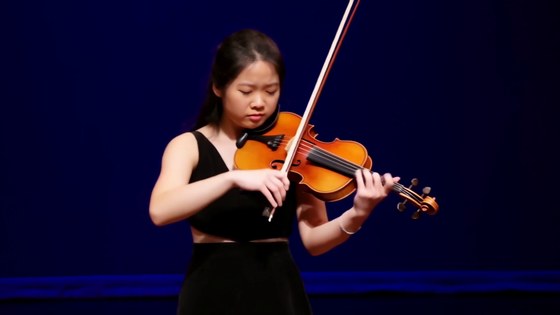}} & \raisebox{-.5\height}{\includegraphics[width=\qualw]{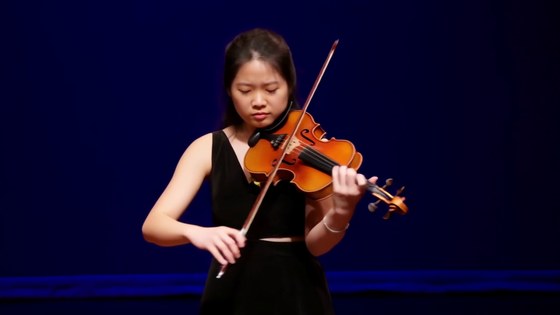}} & \raisebox{-.5\height}{\includegraphics[width=\qualw]{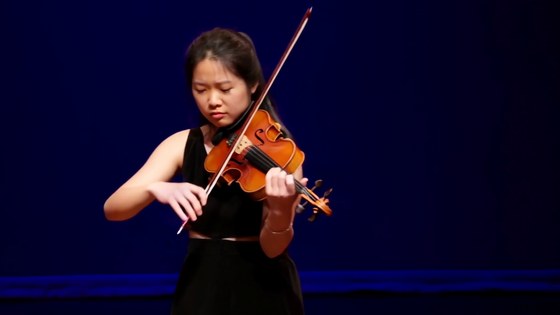}} & \raisebox{-.5\height}{\includegraphics[width=\qualw]{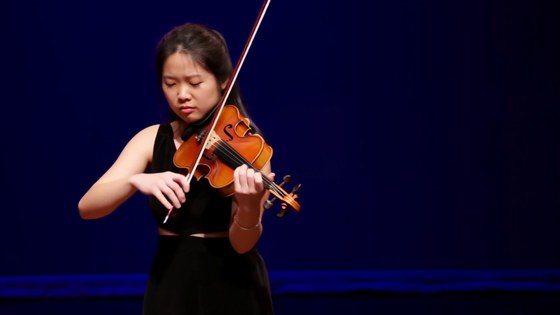}} & \raisebox{-.5\height}{\includegraphics[width=\qualw]{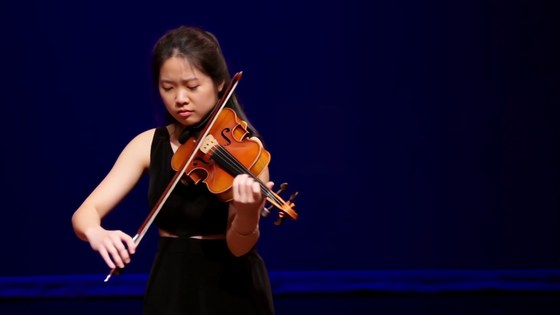}} & \raisebox{-.5\height}{\includegraphics[width=\qualw]{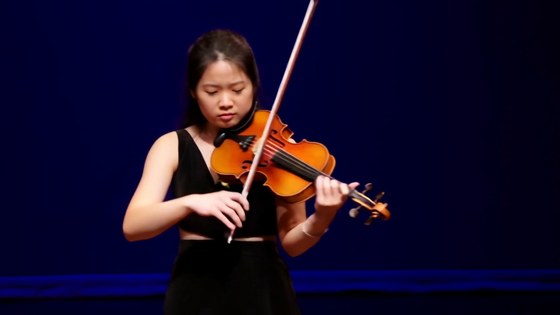}}\\
\end{tabular}}
\par\vspace{3pt}
\makebox[\textwidth][c]{\begin{tabular}{@{}c@{\hspace{1.5pt}}c@{\hspace{1pt}}c@{\hspace{1pt}}c@{\hspace{1pt}}c@{\hspace{1pt}}c@{\hspace{1pt}}c@{}}
\multicolumn{7}{@{}l@{}}{\makebox[\dimexpr6\qualw+5pt\relax][l]{\scriptsize\textbf{720p}, 5.4\,s\quad\textit{old town in morning mist}\hfill 1749\,s\,$\rightarrow$\,769\,s (2.28\x), 35.2\,dB}}\\[1pt]
\multicolumn{7}{@{}p{\dimexpr6\qualw+5pt\relax}@{}}{\tiny\raggedright\textit{Prompt (seed 1106):} Aerial shot of a medieval town with red-tiled roofs and a gothic cathedral at sunrise, mist drifting through the streets.}\\[2pt]
 & \tiny $t{=}0.5$\,s & \tiny $t{=}1.4$\,s & \tiny $t{=}2.2$\,s & \tiny $t{=}3.1$\,s & \tiny $t{=}4.0$\,s & \tiny $t{=}4.8$\,s\\[1pt]
\raisebox{-.5\height}{\rotatebox{90}{\tiny Dense}} & \raisebox{-.5\height}{\includegraphics[width=\qualw]{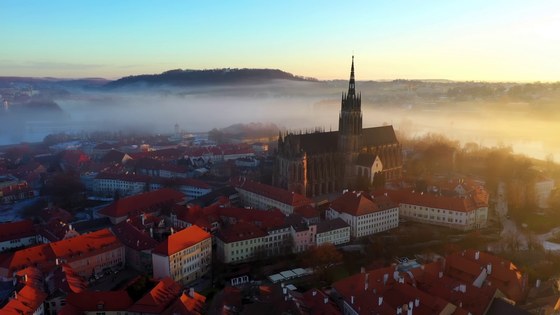}} & \raisebox{-.5\height}{\includegraphics[width=\qualw]{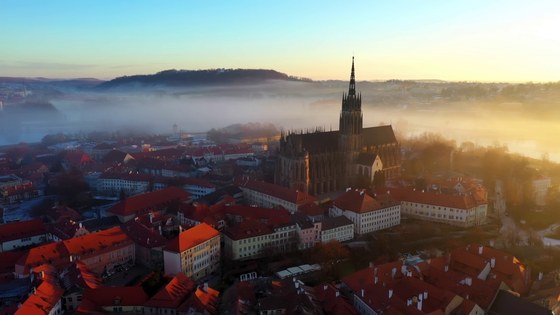}} & \raisebox{-.5\height}{\includegraphics[width=\qualw]{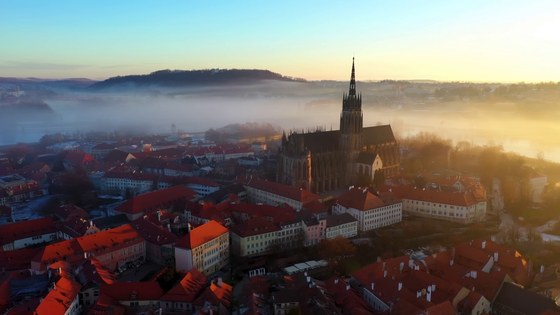}} & \raisebox{-.5\height}{\includegraphics[width=\qualw]{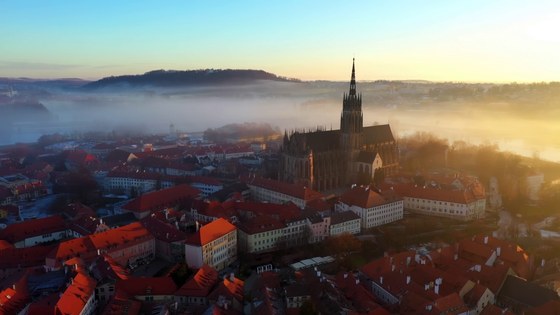}} & \raisebox{-.5\height}{\includegraphics[width=\qualw]{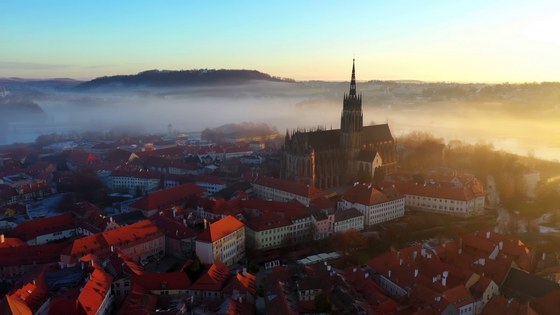}} & \raisebox{-.5\height}{\includegraphics[width=\qualw]{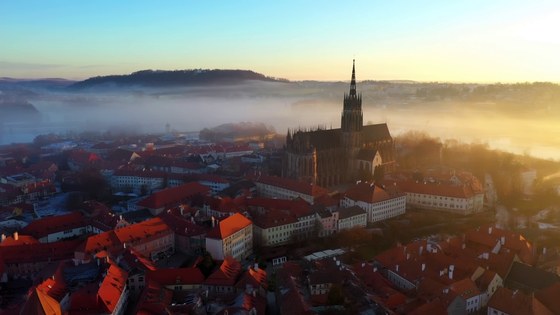}}\\
\end{tabular}}
\caption{HunyuanVideo at 720p (129 frames) on the RTX PRO 6000 with dense attention and with \sys ($\pi = 0.3$), from the same prompt and seed: six evenly spaced frames per video, with the generation time (dense $\rightarrow$ \sys), speedup, and PSNR against the dense video.}
\label{fig:app-qual-hunyuan}
\end{figure*}

\subsection{Qualitative Results on LTX-2.5}
\label{sec:app-qual-ltx}
Figures~\ref{fig:app-qual-ltx}--\ref{fig:app-qual-ltx-3} show videos of LTX-2.5 ($\pi = 0.3$) at 1088p with 721 frames, 1440p with 361 frames, and 4K with 169 frames, with dense attention and with \sys from the same image, prompt, and seed, as six evenly spaced frames; each video is labeled with its generation time, speedup, PSNR against the dense video, and prompt.

\begin{figure*}[p]
\centering
\ifx\qualw\undefined\newlength{\qualw}\fi
\setlength{\qualw}{\dimexpr(\textwidth - 12.5pt)*84/600\relax}%
\makebox[\textwidth][c]{\begin{tabular}{@{}c@{\hspace{1.5pt}}c@{\hspace{1pt}}c@{\hspace{1pt}}c@{\hspace{1pt}}c@{\hspace{1pt}}c@{\hspace{1pt}}c@{}}
\multicolumn{7}{@{}l@{}}{\makebox[\dimexpr6\qualw+5pt\relax][l]{\scriptsize\textbf{1088p}, 30\,s\quad\textit{perfume ad}\hfill 372\,s\,$\rightarrow$\,204\,s (1.82\x), 32.7\,dB}}\\[1pt]
\multicolumn{7}{@{}p{\dimexpr6\qualw+5pt\relax}@{}}{\tiny\raggedright\textit{Prompt (seed 1000):} A luxury perfume commercial. A faceted crystal perfume bottle filled with amber liquid rotates slowly on a black glossy pedestal while golden light rays sweep across it. Tiny droplets of mist drift through the air and catch the light, and silk ribbons in deep red float gracefully behind the bottle. The camera glides in a slow arc around the bottle, ending on a close-up of the engraved glass. Cinematic studio lighting, shallow depth of field, rich contrast, 8K commercial quality. Soft orchestral strings and a gentle shimmer sound.}\\[2pt]
 & \tiny $t{=}3.0$\,s & \tiny $t{=}7.8$\,s & \tiny $t{=}12.6$\,s & \tiny $t{=}17.4$\,s & \tiny $t{=}22.2$\,s & \tiny $t{=}27.0$\,s\\[1pt]
\raisebox{-.5\height}{\rotatebox{90}{\tiny Dense}} & \raisebox{-.5\height}{\includegraphics[width=\qualw]{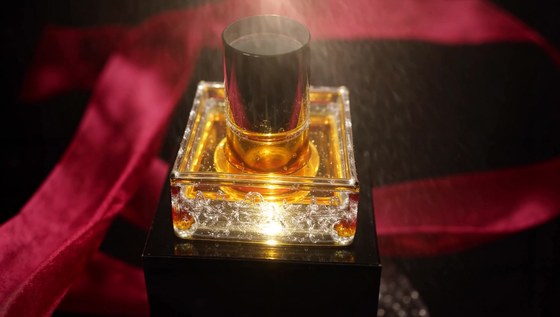}} & \raisebox{-.5\height}{\includegraphics[width=\qualw]{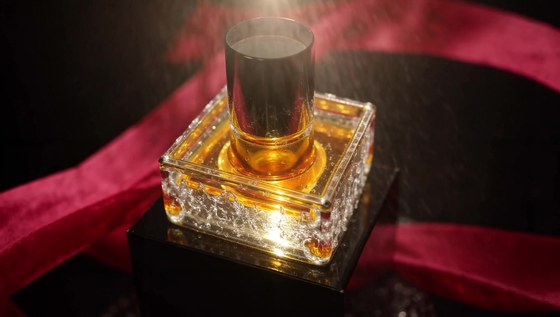}} & \raisebox{-.5\height}{\includegraphics[width=\qualw]{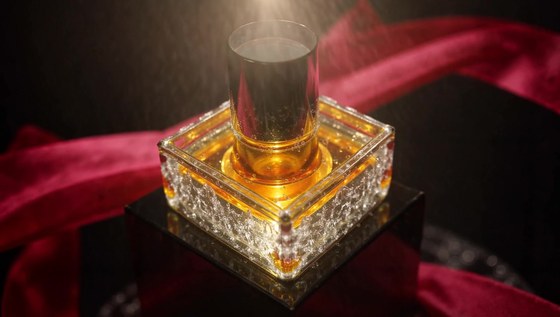}} & \raisebox{-.5\height}{\includegraphics[width=\qualw]{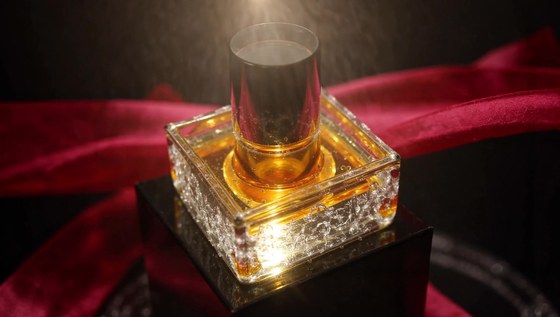}} & \raisebox{-.5\height}{\includegraphics[width=\qualw]{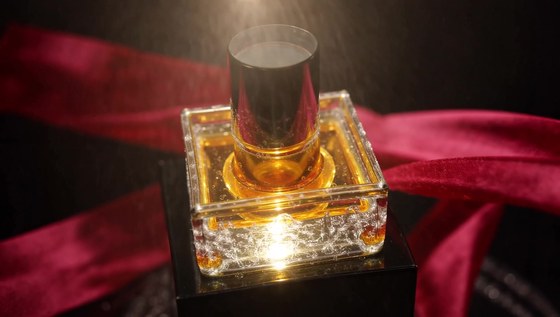}} & \raisebox{-.5\height}{\includegraphics[width=\qualw]{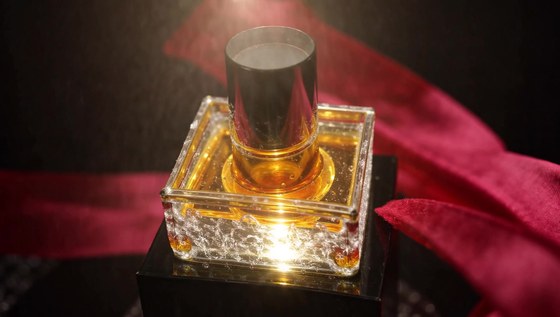}}\\[1pt]
\raisebox{-.5\height}{\rotatebox{90}{\tiny \sys}} & \raisebox{-.5\height}{\includegraphics[width=\qualw]{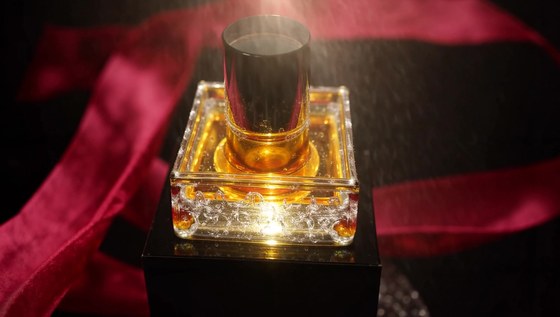}} & \raisebox{-.5\height}{\includegraphics[width=\qualw]{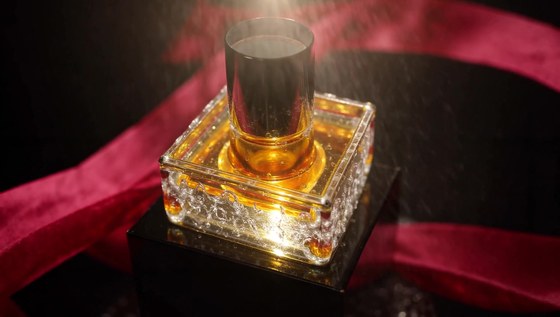}} & \raisebox{-.5\height}{\includegraphics[width=\qualw]{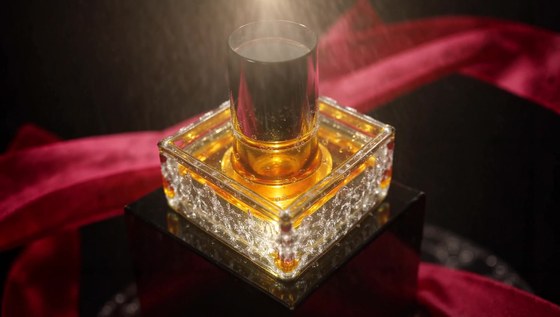}} & \raisebox{-.5\height}{\includegraphics[width=\qualw]{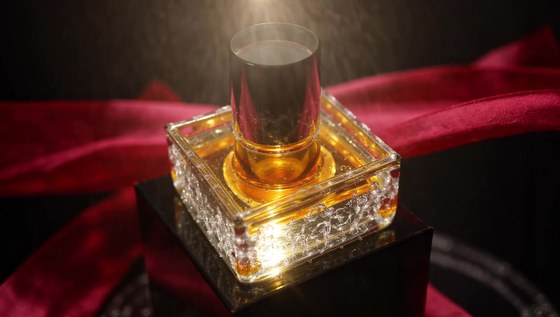}} & \raisebox{-.5\height}{\includegraphics[width=\qualw]{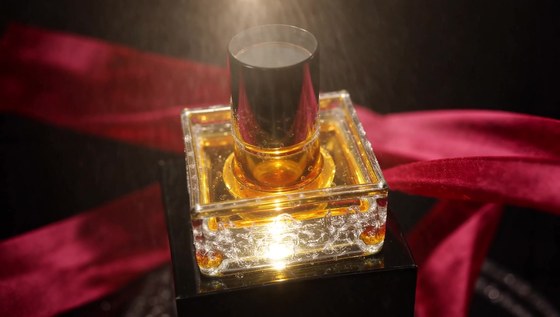}} & \raisebox{-.5\height}{\includegraphics[width=\qualw]{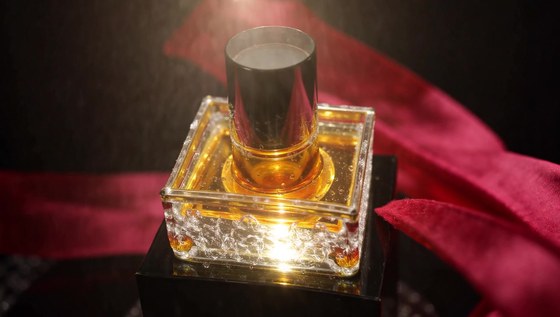}}\\
\end{tabular}}
\par\vspace{1pt}
\makebox[\textwidth][c]{\begin{tabular}{@{}c@{\hspace{1.5pt}}c@{\hspace{1pt}}c@{\hspace{1pt}}c@{\hspace{1pt}}c@{\hspace{1pt}}c@{\hspace{1pt}}c@{}}
\multicolumn{7}{@{}l@{}}{\makebox[\dimexpr6\qualw+5pt\relax][l]{\scriptsize\textbf{1088p}, 30\,s\quad\textit{fox 3d}\hfill 372\,s\,$\rightarrow$\,204\,s (1.82\x), 36.0\,dB}}\\[1pt]
\multicolumn{7}{@{}p{\dimexpr6\qualw+5pt\relax}@{}}{\tiny\raggedright\textit{Prompt (seed 1003):} A charming 3D animated scene in the style of a Pixar film. A small round orange fox with big curious eyes tiptoes through a magical forest at night, where glowing mushrooms and fireflies light up the path. The fox discovers a tiny floating lantern, gently nudges it with its nose, and its face lights up with a delighted smile as the lantern rises into the sky. Soft volumetric lighting, rich colors, expressive character animation. Whimsical music box melody, soft crickets and the gentle chime of the lantern.}\\[2pt]
 & \tiny $t{=}3.0$\,s & \tiny $t{=}7.8$\,s & \tiny $t{=}12.6$\,s & \tiny $t{=}17.4$\,s & \tiny $t{=}22.2$\,s & \tiny $t{=}27.0$\,s\\[1pt]
\raisebox{-.5\height}{\rotatebox{90}{\tiny Dense}} & \raisebox{-.5\height}{\includegraphics[width=\qualw]{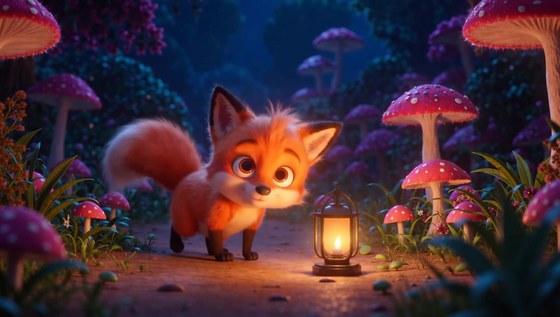}} & \raisebox{-.5\height}{\includegraphics[width=\qualw]{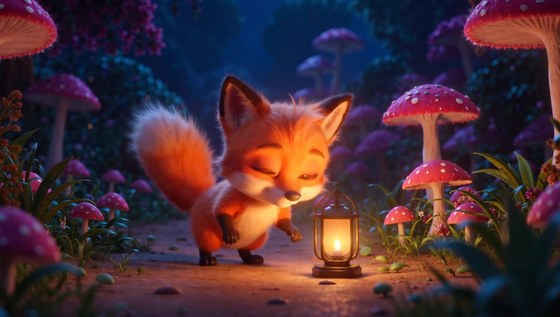}} & \raisebox{-.5\height}{\includegraphics[width=\qualw]{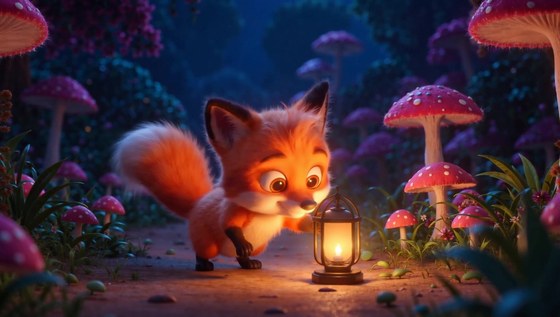}} & \raisebox{-.5\height}{\includegraphics[width=\qualw]{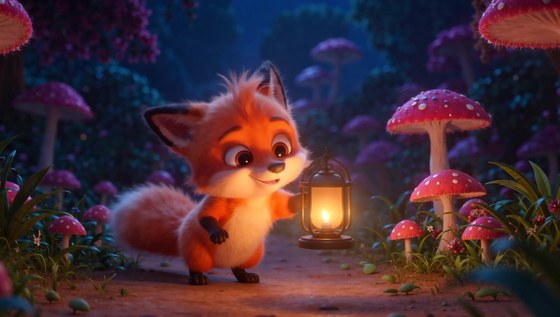}} & \raisebox{-.5\height}{\includegraphics[width=\qualw]{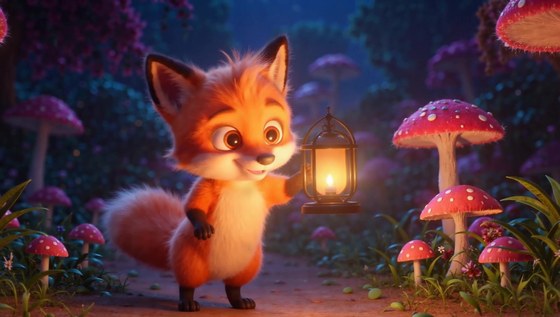}} & \raisebox{-.5\height}{\includegraphics[width=\qualw]{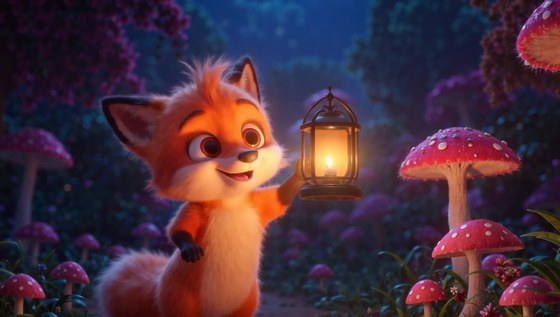}}\\[1pt]
\raisebox{-.5\height}{\rotatebox{90}{\tiny \sys}} & \raisebox{-.5\height}{\includegraphics[width=\qualw]{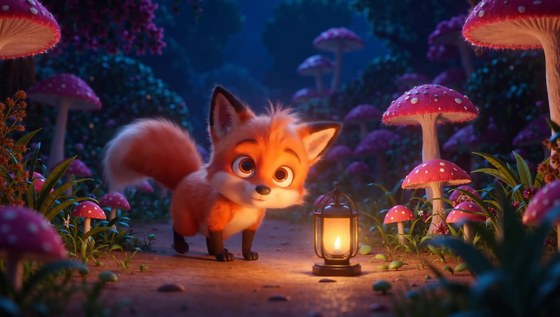}} & \raisebox{-.5\height}{\includegraphics[width=\qualw]{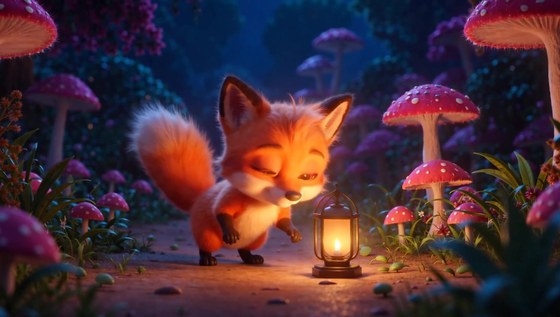}} & \raisebox{-.5\height}{\includegraphics[width=\qualw]{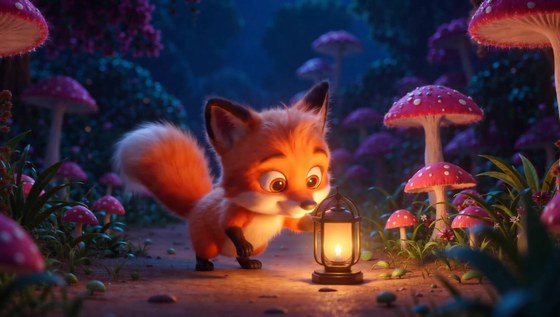}} & \raisebox{-.5\height}{\includegraphics[width=\qualw]{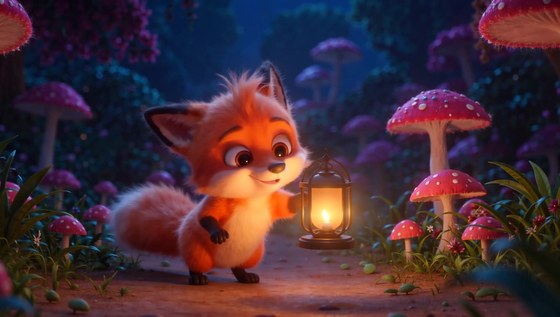}} & \raisebox{-.5\height}{\includegraphics[width=\qualw]{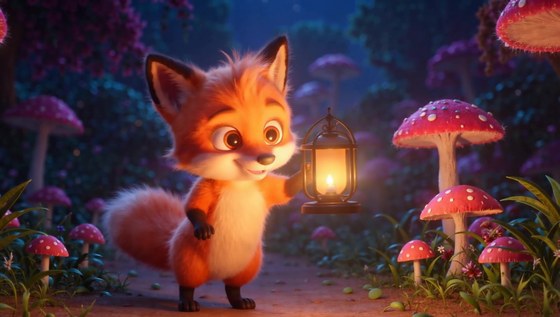}} & \raisebox{-.5\height}{\includegraphics[width=\qualw]{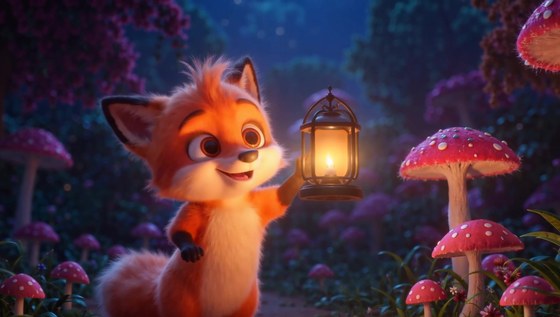}}\\
\end{tabular}}
\par\vspace{1pt}
\makebox[\textwidth][c]{\begin{tabular}{@{}c@{\hspace{1.5pt}}c@{\hspace{1pt}}c@{\hspace{1pt}}c@{\hspace{1pt}}c@{\hspace{1pt}}c@{\hspace{1pt}}c@{}}
\multicolumn{7}{@{}l@{}}{\makebox[\dimexpr6\qualw+5pt\relax][l]{\scriptsize\textbf{1088p}, 30\,s\quad\textit{ghibli bicycle}\hfill 372\,s\,$\rightarrow$\,203\,s (1.83\x), 30.8\,dB}}\\[1pt]
\multicolumn{7}{@{}p{\dimexpr6\qualw+5pt\relax}@{}}{\tiny\raggedright\textit{Prompt (seed 1004):} A hand-drawn anime scene in the style of Studio Ghibli. A girl with a straw hat rides a bicycle down a hill through vast green rice fields under a bright blue sky with towering white clouds, her hair and scarf fluttering in the wind. The camera follows alongside her, then rises to reveal a small seaside town and the sparkling ocean in the distance. Painterly backgrounds, soft watercolor textures, warm nostalgic summer atmosphere. Light piano and strings, wind rustling through the grass and the ring of a bicycle bell.}\\[2pt]
 & \tiny $t{=}3.0$\,s & \tiny $t{=}7.8$\,s & \tiny $t{=}12.6$\,s & \tiny $t{=}17.4$\,s & \tiny $t{=}22.2$\,s & \tiny $t{=}27.0$\,s\\[1pt]
\raisebox{-.5\height}{\rotatebox{90}{\tiny Dense}} & \raisebox{-.5\height}{\includegraphics[width=\qualw]{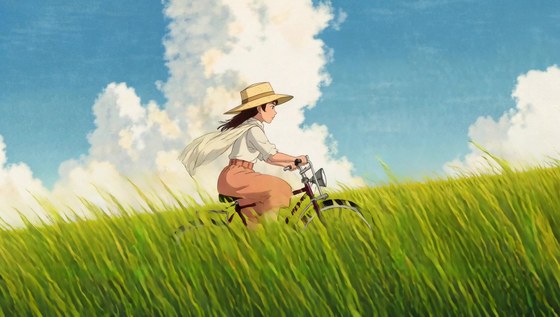}} & \raisebox{-.5\height}{\includegraphics[width=\qualw]{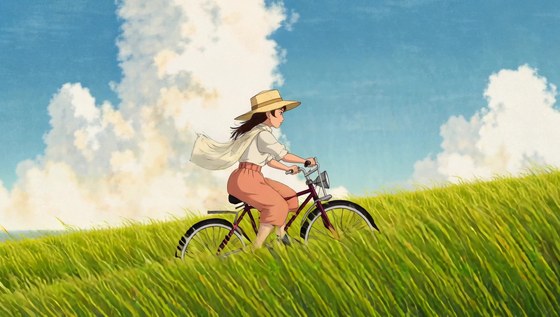}} & \raisebox{-.5\height}{\includegraphics[width=\qualw]{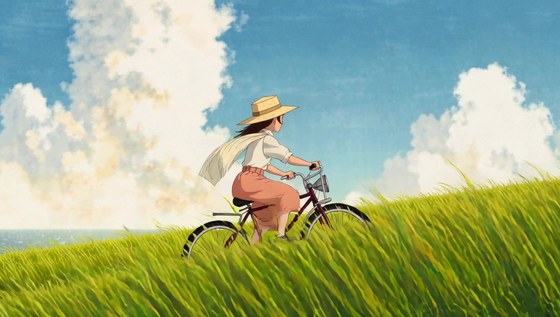}} & \raisebox{-.5\height}{\includegraphics[width=\qualw]{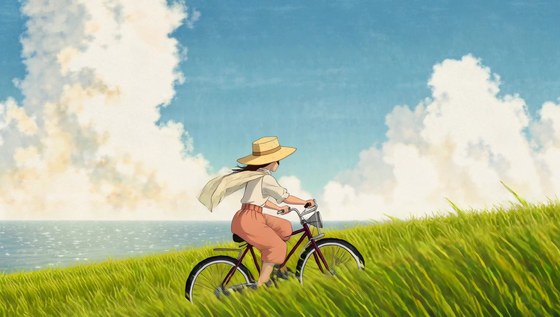}} & \raisebox{-.5\height}{\includegraphics[width=\qualw]{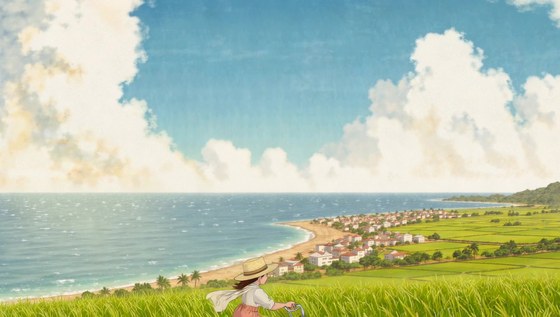}} & \raisebox{-.5\height}{\includegraphics[width=\qualw]{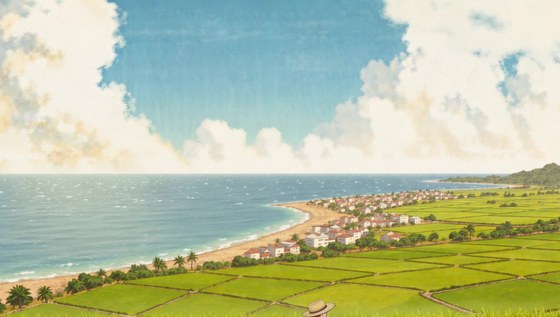}}\\[1pt]
\raisebox{-.5\height}{\rotatebox{90}{\tiny \sys}} & \raisebox{-.5\height}{\includegraphics[width=\qualw]{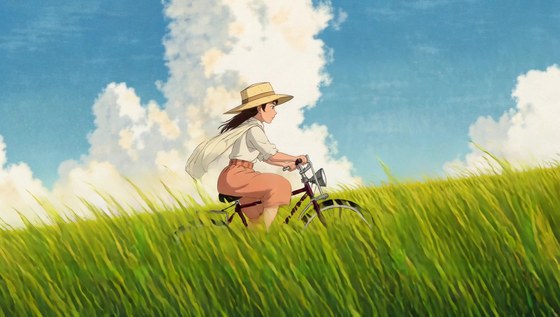}} & \raisebox{-.5\height}{\includegraphics[width=\qualw]{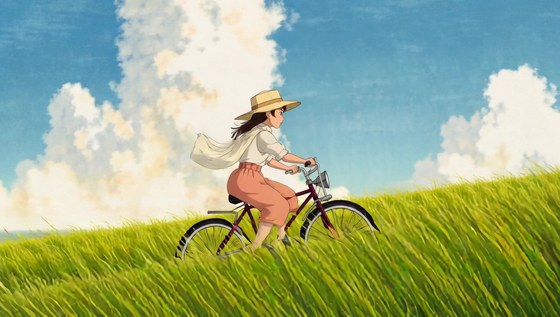}} & \raisebox{-.5\height}{\includegraphics[width=\qualw]{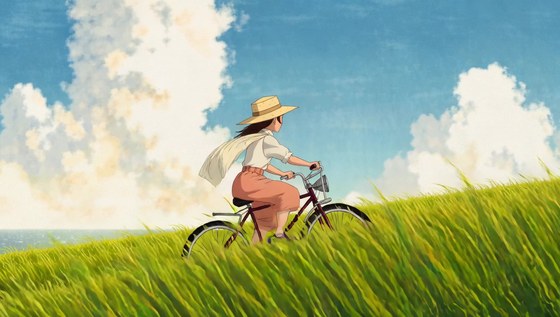}} & \raisebox{-.5\height}{\includegraphics[width=\qualw]{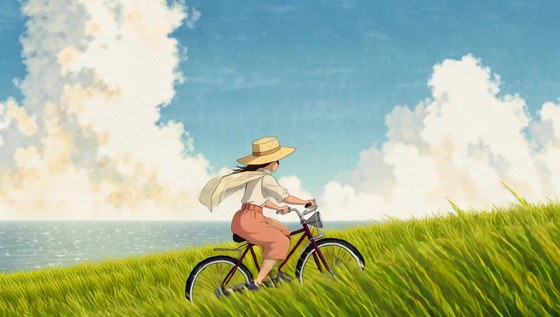}} & \raisebox{-.5\height}{\includegraphics[width=\qualw]{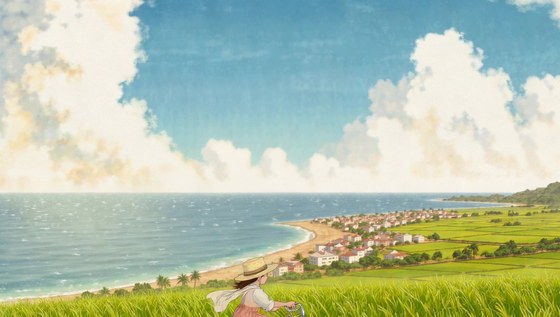}} & \raisebox{-.5\height}{\includegraphics[width=\qualw]{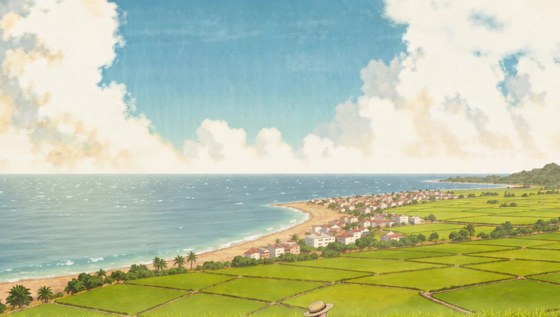}}\\
\end{tabular}}
\par\vspace{1pt}
\makebox[\textwidth][c]{\begin{tabular}{@{}c@{\hspace{1.5pt}}c@{\hspace{1pt}}c@{\hspace{1pt}}c@{\hspace{1pt}}c@{\hspace{1pt}}c@{\hspace{1pt}}c@{}}
\multicolumn{7}{@{}l@{}}{\makebox[\dimexpr6\qualw+5pt\relax][l]{\scriptsize\textbf{1088p}, 30\,s\quad\textit{humpback whale}\hfill 376\,s\,$\rightarrow$\,205\,s (1.84\x), 33.2\,dB}}\\[1pt]
\multicolumn{7}{@{}p{\dimexpr6\qualw+5pt\relax}@{}}{\tiny\raggedright\textit{Prompt (seed 1009):} A breathtaking nature documentary shot. A majestic humpback whale breaches out of a calm turquoise ocean at golden hour, water cascading off its massive body in slow motion before it crashes back with an enormous splash. Aerial drone view pulls back to reveal a pod of whales and a glowing orange horizon. Hyper-realistic, crisp detail, rich natural colors, epic scale. Deep whale song, crashing water and a swelling orchestral score.}\\[2pt]
 & \tiny $t{=}3.0$\,s & \tiny $t{=}7.8$\,s & \tiny $t{=}12.6$\,s & \tiny $t{=}17.4$\,s & \tiny $t{=}22.2$\,s & \tiny $t{=}27.0$\,s\\[1pt]
\raisebox{-.5\height}{\rotatebox{90}{\tiny Dense}} & \raisebox{-.5\height}{\includegraphics[width=\qualw]{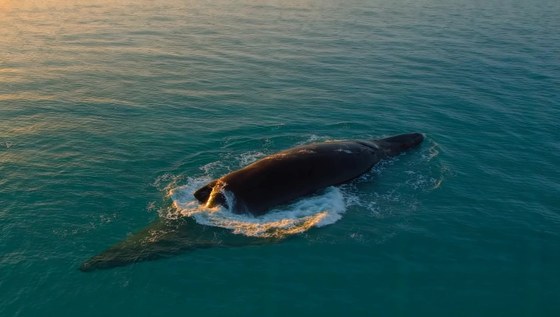}} & \raisebox{-.5\height}{\includegraphics[width=\qualw]{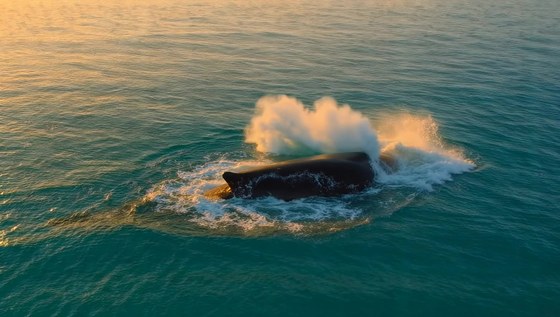}} & \raisebox{-.5\height}{\includegraphics[width=\qualw]{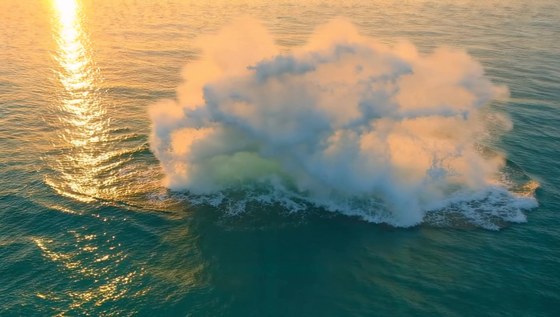}} & \raisebox{-.5\height}{\includegraphics[width=\qualw]{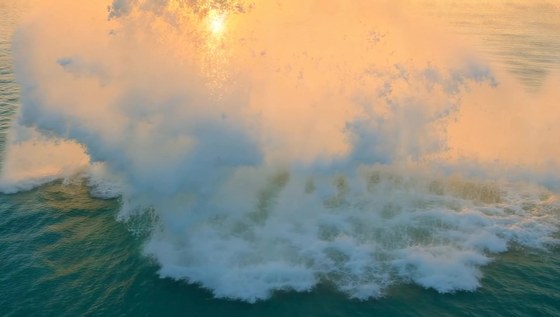}} & \raisebox{-.5\height}{\includegraphics[width=\qualw]{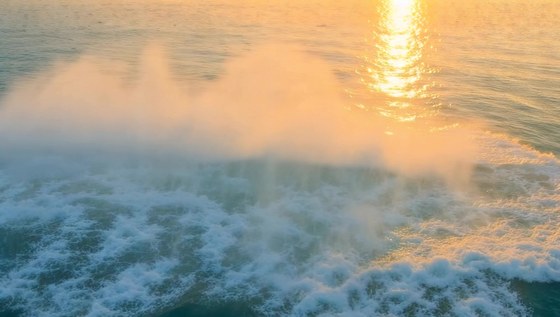}} & \raisebox{-.5\height}{\includegraphics[width=\qualw]{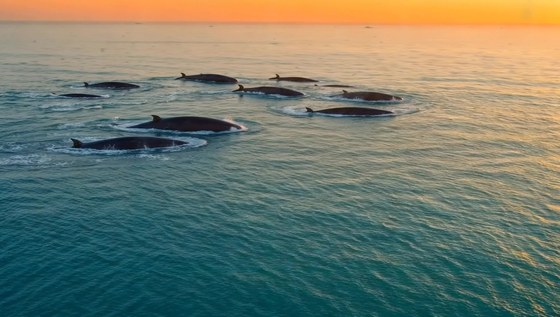}}\\[1pt]
\raisebox{-.5\height}{\rotatebox{90}{\tiny \sys}} & \raisebox{-.5\height}{\includegraphics[width=\qualw]{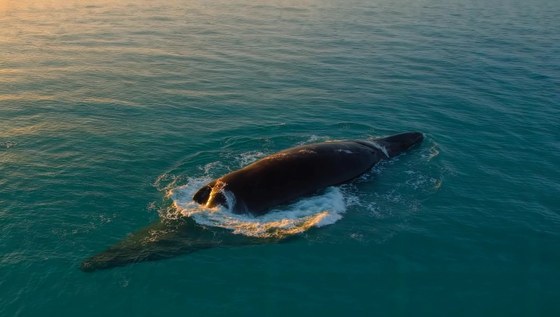}} & \raisebox{-.5\height}{\includegraphics[width=\qualw]{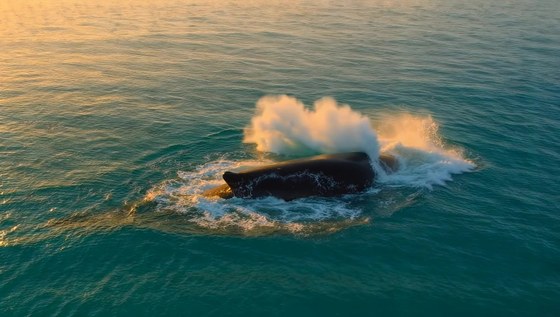}} & \raisebox{-.5\height}{\includegraphics[width=\qualw]{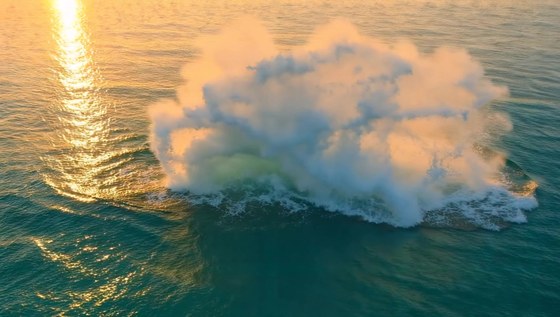}} & \raisebox{-.5\height}{\includegraphics[width=\qualw]{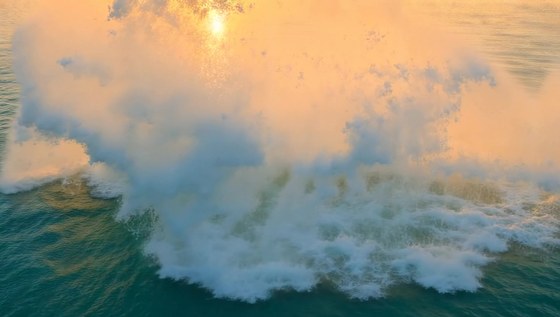}} & \raisebox{-.5\height}{\includegraphics[width=\qualw]{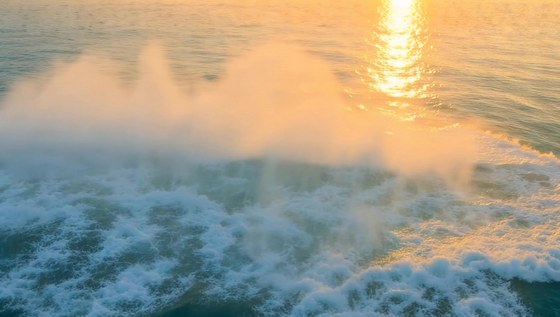}} & \raisebox{-.5\height}{\includegraphics[width=\qualw]{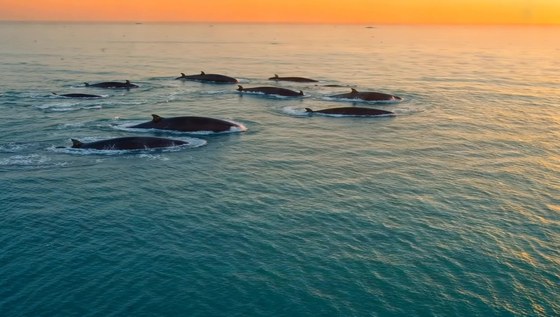}}\\
\end{tabular}}
\caption{LTX-2.5 image-to-video at 1088p (721 frames), 1440p (361 frames), and 4K (169 frames) on the RTX PRO 6000 with dense attention and with \sys ($\pi = 0.3$), from the same image, prompt, and seed: six evenly spaced frames per video, with the setting, generation time (dense $\rightarrow$ \sys), speedup, and PSNR against the dense video (1/3).}
\label{fig:app-qual-ltx}
\end{figure*}

\begin{figure*}[p]
\centering
\ifx\qualw\undefined\newlength{\qualw}\fi
\setlength{\qualw}{\dimexpr(\textwidth - 12.5pt)*84/600\relax}%
\makebox[\textwidth][c]{\begin{tabular}{@{}c@{\hspace{1.5pt}}c@{\hspace{1pt}}c@{\hspace{1pt}}c@{\hspace{1pt}}c@{\hspace{1pt}}c@{\hspace{1pt}}c@{}}
\multicolumn{7}{@{}l@{}}{\makebox[\dimexpr6\qualw+5pt\relax][l]{\scriptsize\textbf{1440p}, 15\,s\quad\textit{wasteland mech dog}\hfill 315\,s\,$\rightarrow$\,176\,s (1.79\x), 34.6\,dB}}\\[1pt]
\multicolumn{7}{@{}p{\dimexpr6\qualw+5pt\relax}@{}}{\tiny\raggedright\textit{Prompt (seed 1009):} A post-apocalyptic adventure. A lone wanderer in a dusty cloak walks across a vast desert of sand dunes with a robotic mechanical dog at her side, passing the ruins of colossal skyscrapers half-buried in sand, as a massive sandstorm wall rolls in behind them. Epic wide shots, cinematic sci-fi, warm orange haze. Howling wind, mechanical footsteps and a haunting score.}\\[2pt]
 & \tiny $t{=}1.5$\,s & \tiny $t{=}3.9$\,s & \tiny $t{=}6.3$\,s & \tiny $t{=}8.7$\,s & \tiny $t{=}11.1$\,s & \tiny $t{=}13.5$\,s\\[1pt]
\raisebox{-.5\height}{\rotatebox{90}{\tiny Dense}} & \raisebox{-.5\height}{\includegraphics[width=\qualw]{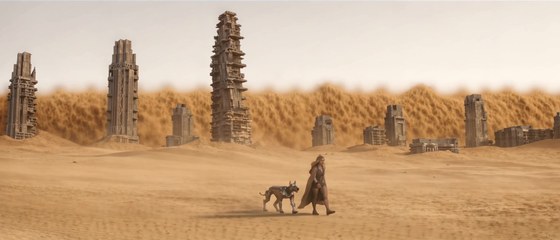}} & \raisebox{-.5\height}{\includegraphics[width=\qualw]{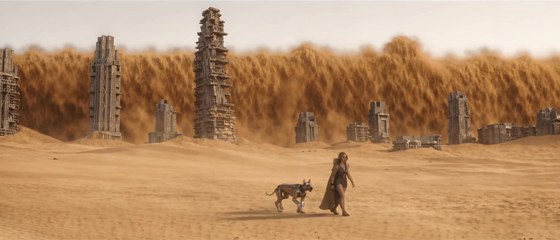}} & \raisebox{-.5\height}{\includegraphics[width=\qualw]{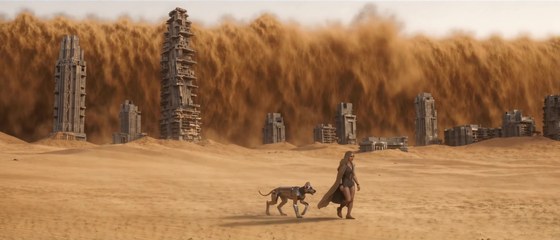}} & \raisebox{-.5\height}{\includegraphics[width=\qualw]{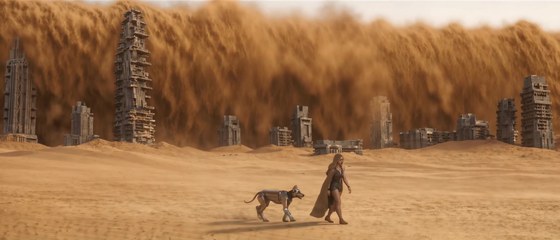}} & \raisebox{-.5\height}{\includegraphics[width=\qualw]{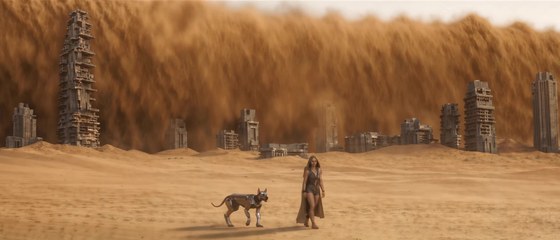}} & \raisebox{-.5\height}{\includegraphics[width=\qualw]{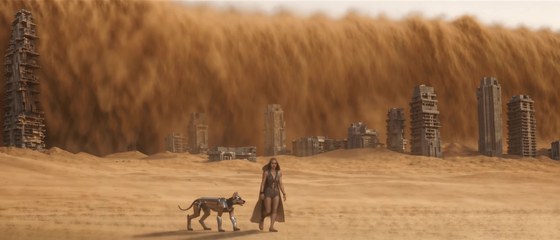}}\\[1pt]
\raisebox{-.5\height}{\rotatebox{90}{\tiny \sys}} & \raisebox{-.5\height}{\includegraphics[width=\qualw]{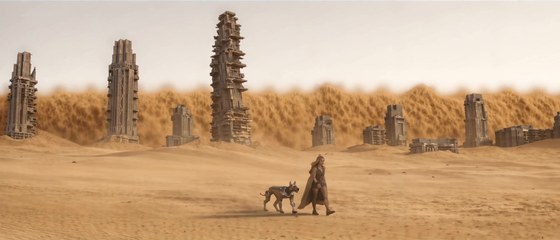}} & \raisebox{-.5\height}{\includegraphics[width=\qualw]{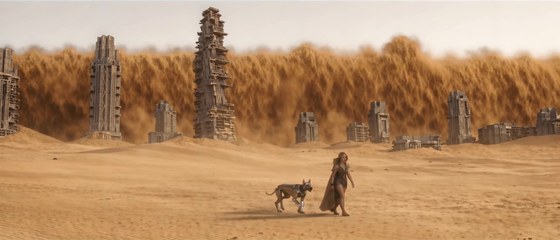}} & \raisebox{-.5\height}{\includegraphics[width=\qualw]{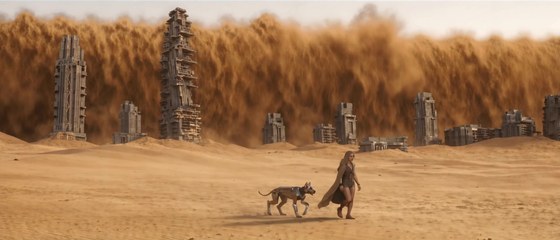}} & \raisebox{-.5\height}{\includegraphics[width=\qualw]{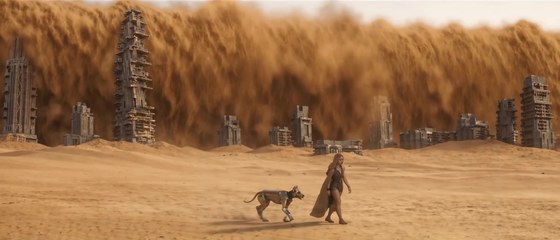}} & \raisebox{-.5\height}{\includegraphics[width=\qualw]{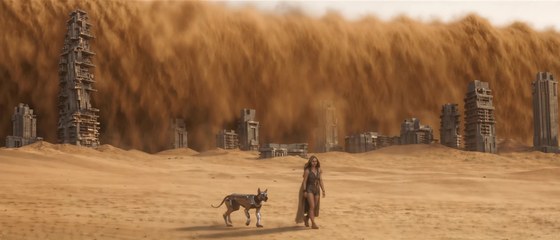}} & \raisebox{-.5\height}{\includegraphics[width=\qualw]{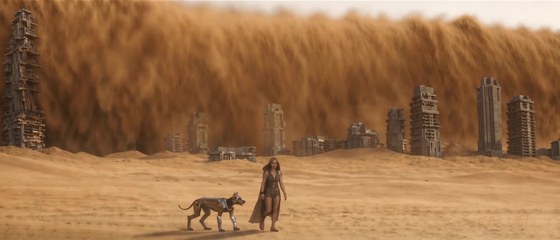}}\\
\end{tabular}}
\par\vspace{1pt}
\makebox[\textwidth][c]{\begin{tabular}{@{}c@{\hspace{1.5pt}}c@{\hspace{1pt}}c@{\hspace{1pt}}c@{\hspace{1pt}}c@{\hspace{1pt}}c@{\hspace{1pt}}c@{}}
\multicolumn{7}{@{}l@{}}{\makebox[\dimexpr6\qualw+5pt\relax][l]{\scriptsize\textbf{1440p}, 15\,s\quad\textit{wizard duel}\hfill 319\,s\,$\rightarrow$\,180\,s (1.77\x), 26.8\,dB}}\\[1pt]
\multicolumn{7}{@{}p{\dimexpr6\qualw+5pt\relax}@{}}{\tiny\raggedright\textit{Prompt (seed 1011):} An anime fantasy duel between two wizards in a ruined cathedral. One summons giant glowing magic circles that fire beams of light, the other conjures a swirling dragon of black fire; their spells collide in a massive explosion of light and stained-glass shards. Vibrant anime effects, dramatic lighting, high-energy camera. Booming spells, shattering glass and an epic choir.}\\[2pt]
 & \tiny $t{=}1.5$\,s & \tiny $t{=}3.9$\,s & \tiny $t{=}6.3$\,s & \tiny $t{=}8.7$\,s & \tiny $t{=}11.1$\,s & \tiny $t{=}13.5$\,s\\[1pt]
\raisebox{-.5\height}{\rotatebox{90}{\tiny Dense}} & \raisebox{-.5\height}{\includegraphics[width=\qualw]{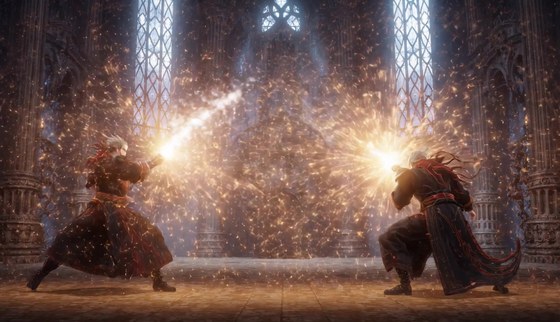}} & \raisebox{-.5\height}{\includegraphics[width=\qualw]{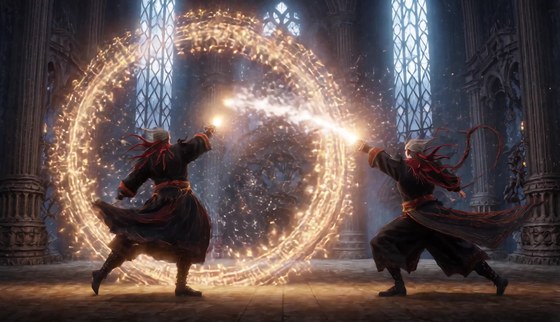}} & \raisebox{-.5\height}{\includegraphics[width=\qualw]{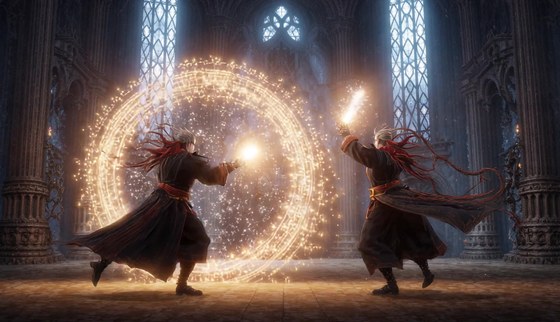}} & \raisebox{-.5\height}{\includegraphics[width=\qualw]{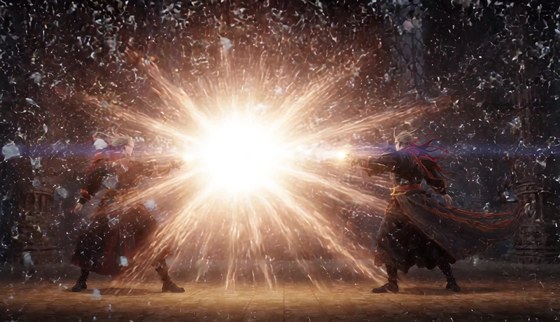}} & \raisebox{-.5\height}{\includegraphics[width=\qualw]{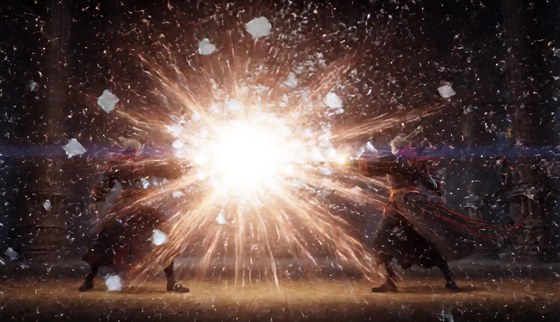}} & \raisebox{-.5\height}{\includegraphics[width=\qualw]{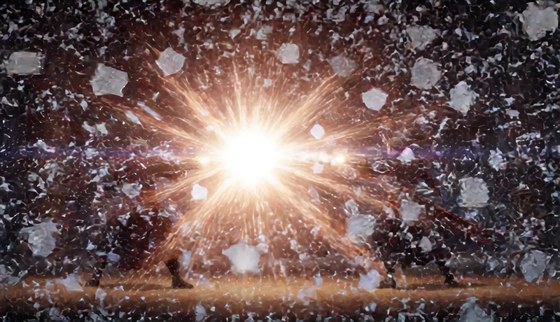}}\\[1pt]
\raisebox{-.5\height}{\rotatebox{90}{\tiny \sys}} & \raisebox{-.5\height}{\includegraphics[width=\qualw]{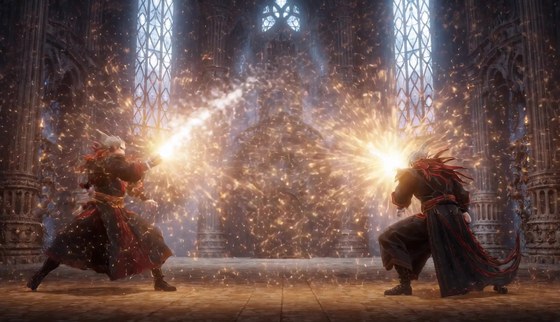}} & \raisebox{-.5\height}{\includegraphics[width=\qualw]{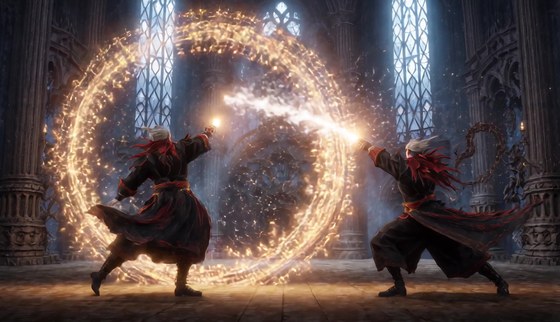}} & \raisebox{-.5\height}{\includegraphics[width=\qualw]{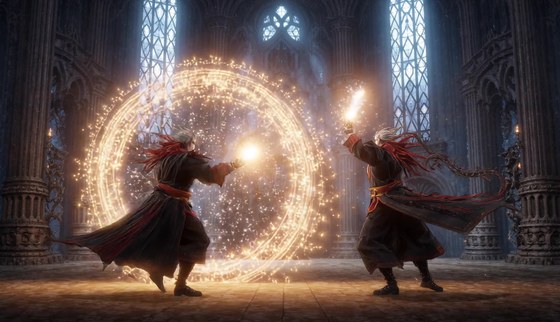}} & \raisebox{-.5\height}{\includegraphics[width=\qualw]{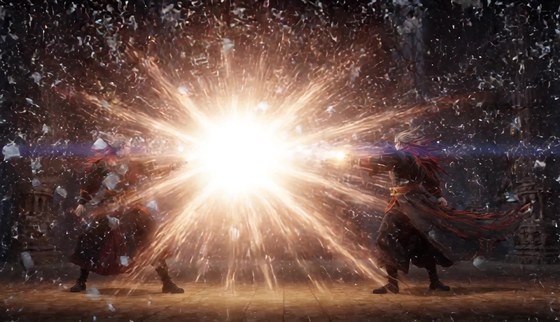}} & \raisebox{-.5\height}{\includegraphics[width=\qualw]{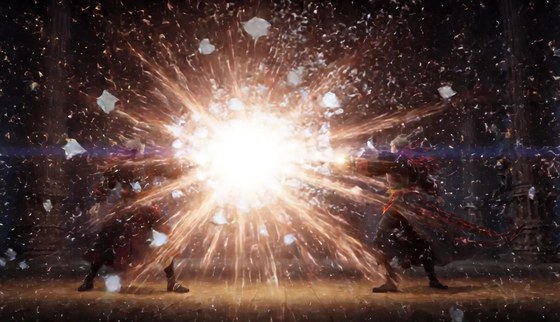}} & \raisebox{-.5\height}{\includegraphics[width=\qualw]{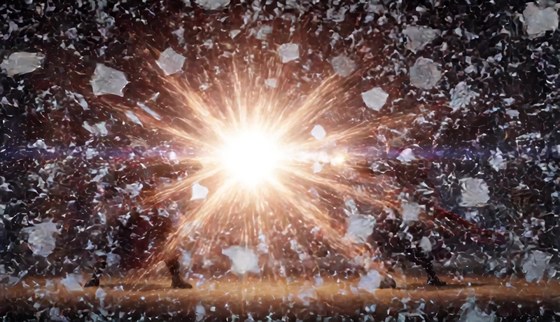}}\\
\end{tabular}}
\par\vspace{1pt}
\makebox[\textwidth][c]{\begin{tabular}{@{}c@{\hspace{1.5pt}}c@{\hspace{1pt}}c@{\hspace{1pt}}c@{\hspace{1pt}}c@{\hspace{1pt}}c@{\hspace{1pt}}c@{}}
\multicolumn{7}{@{}l@{}}{\makebox[\dimexpr6\qualw+5pt\relax][l]{\scriptsize\textbf{1440p}, 15\,s\quad\textit{rocket launch night}\hfill 319\,s\,$\rightarrow$\,179\,s (1.78\x), 39.6\,dB}}\\[1pt]
\multicolumn{7}{@{}p{\dimexpr6\qualw+5pt\relax}@{}}{\tiny\raggedright\textit{Prompt (seed 1014):} A realistic rocket launch at night. A giant rocket stands on the launch pad lit by floodlights; the engines ignite in a blinding burst of fire and billowing smoke, and the rocket rises slowly, then streaks into the dark sky leaving a glowing trail, its flame reflecting in the water. Cinematic slow motion, ultra-detailed, awe-inspiring. Deep thunderous rumble, a crackling roar and a soaring score.}\\[2pt]
 & \tiny $t{=}1.5$\,s & \tiny $t{=}3.9$\,s & \tiny $t{=}6.3$\,s & \tiny $t{=}8.7$\,s & \tiny $t{=}11.1$\,s & \tiny $t{=}13.5$\,s\\[1pt]
\raisebox{-.5\height}{\rotatebox{90}{\tiny Dense}} & \raisebox{-.5\height}{\includegraphics[width=\qualw]{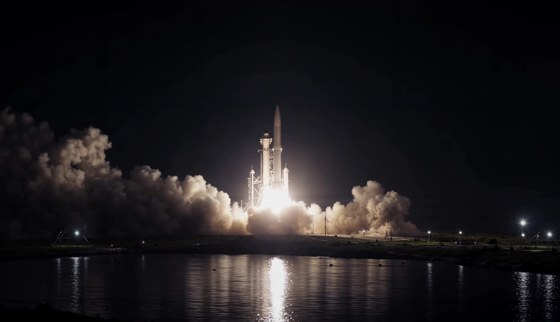}} & \raisebox{-.5\height}{\includegraphics[width=\qualw]{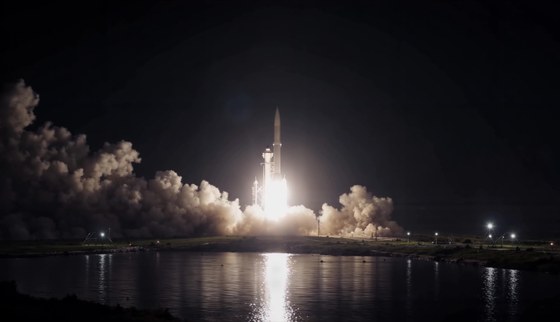}} & \raisebox{-.5\height}{\includegraphics[width=\qualw]{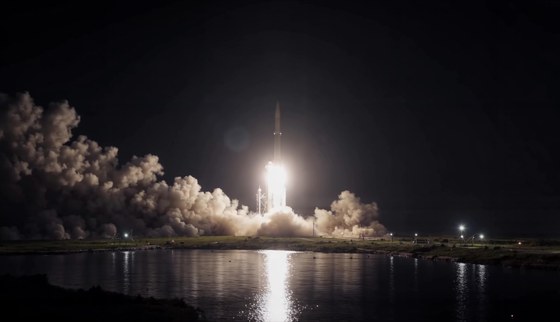}} & \raisebox{-.5\height}{\includegraphics[width=\qualw]{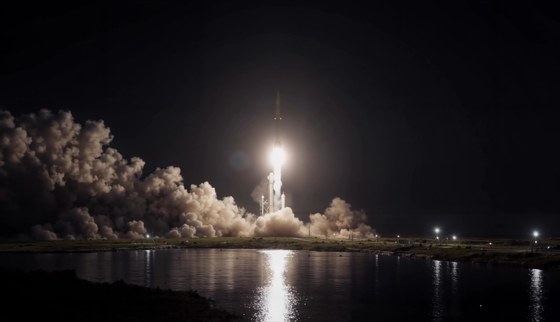}} & \raisebox{-.5\height}{\includegraphics[width=\qualw]{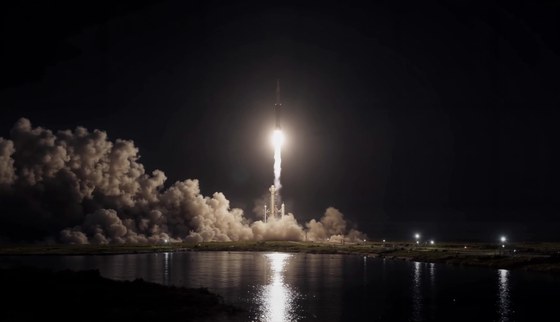}} & \raisebox{-.5\height}{\includegraphics[width=\qualw]{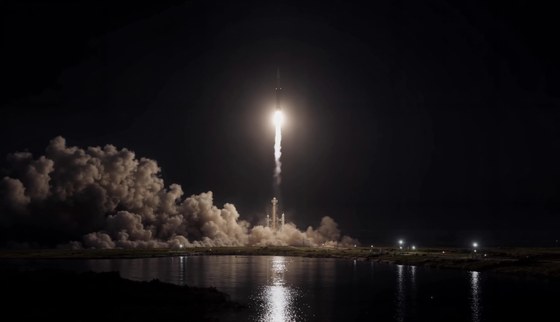}}\\[1pt]
\raisebox{-.5\height}{\rotatebox{90}{\tiny \sys}} & \raisebox{-.5\height}{\includegraphics[width=\qualw]{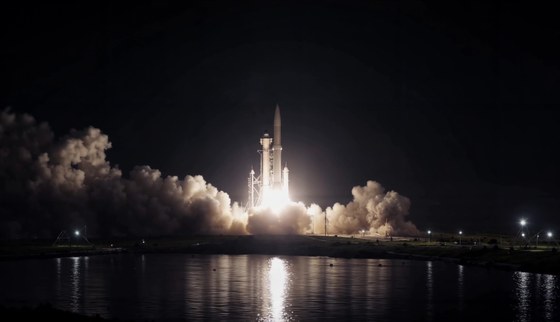}} & \raisebox{-.5\height}{\includegraphics[width=\qualw]{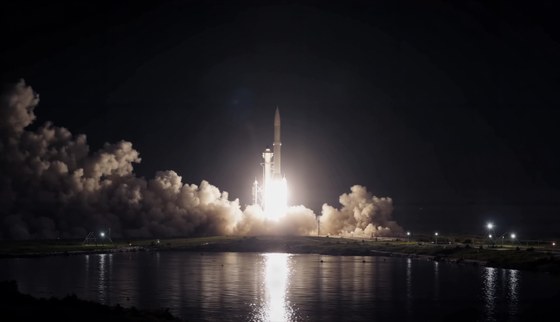}} & \raisebox{-.5\height}{\includegraphics[width=\qualw]{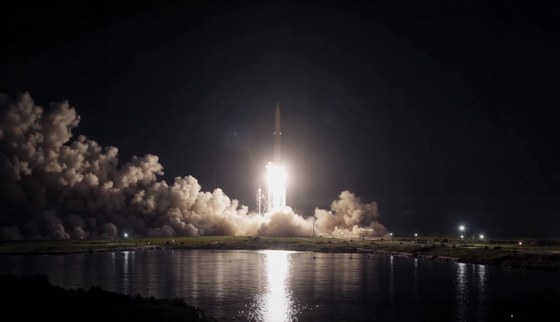}} & \raisebox{-.5\height}{\includegraphics[width=\qualw]{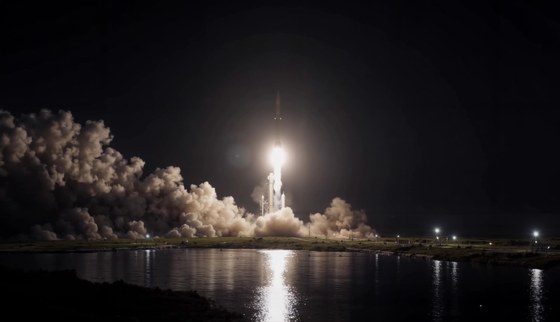}} & \raisebox{-.5\height}{\includegraphics[width=\qualw]{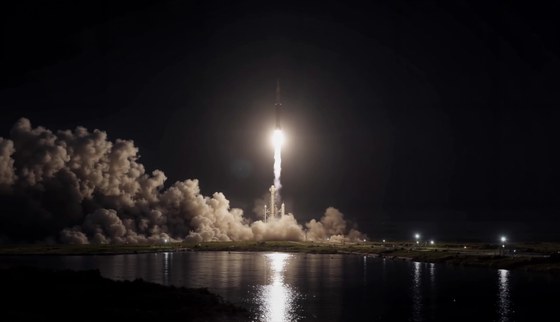}} & \raisebox{-.5\height}{\includegraphics[width=\qualw]{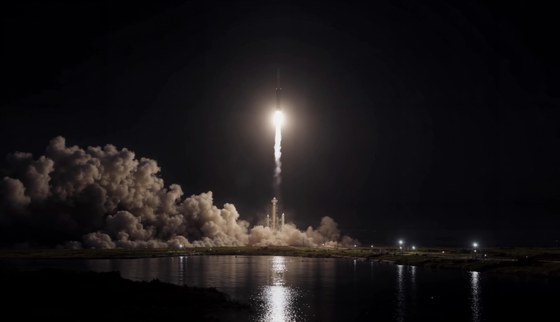}}\\
\end{tabular}}
\par\vspace{1pt}
\makebox[\textwidth][c]{\begin{tabular}{@{}c@{\hspace{1.5pt}}c@{\hspace{1pt}}c@{\hspace{1pt}}c@{\hspace{1pt}}c@{\hspace{1pt}}c@{\hspace{1pt}}c@{}}
\multicolumn{7}{@{}l@{}}{\makebox[\dimexpr6\qualw+5pt\relax][l]{\scriptsize\textbf{1440p}, 15\,s\quad\textit{lantern river}\hfill 318\,s\,$\rightarrow$\,180\,s (1.77\x), 30.4\,dB}}\\[1pt]
\multicolumn{7}{@{}p{\dimexpr6\qualw+5pt\relax}@{}}{\tiny\raggedright\textit{Prompt (seed 1016):} A peaceful night festival in an old Asian river town. Thousands of glowing paper lanterns drift slowly down a calm river and float up into the starry sky, their warm light reflecting on the water as a wooden boat glides past under a stone bridge. Dreamy, cinematic, warm golden tones. Gentle flowing water, distant music and soft wind.}\\[2pt]
 & \tiny $t{=}1.5$\,s & \tiny $t{=}3.9$\,s & \tiny $t{=}6.3$\,s & \tiny $t{=}8.7$\,s & \tiny $t{=}11.1$\,s & \tiny $t{=}13.5$\,s\\[1pt]
\raisebox{-.5\height}{\rotatebox{90}{\tiny Dense}} & \raisebox{-.5\height}{\includegraphics[width=\qualw]{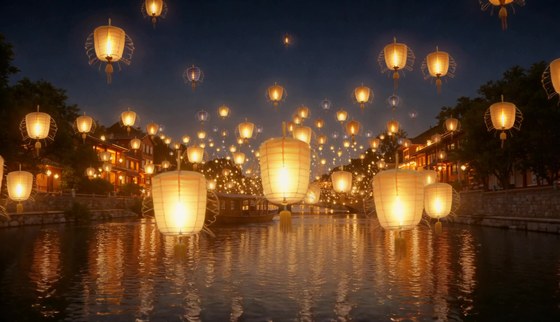}} & \raisebox{-.5\height}{\includegraphics[width=\qualw]{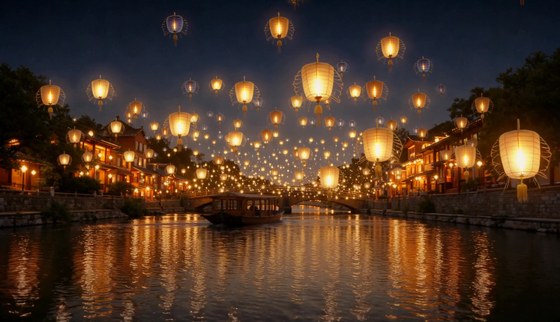}} & \raisebox{-.5\height}{\includegraphics[width=\qualw]{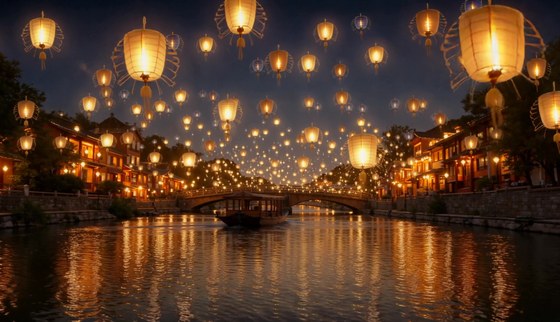}} & \raisebox{-.5\height}{\includegraphics[width=\qualw]{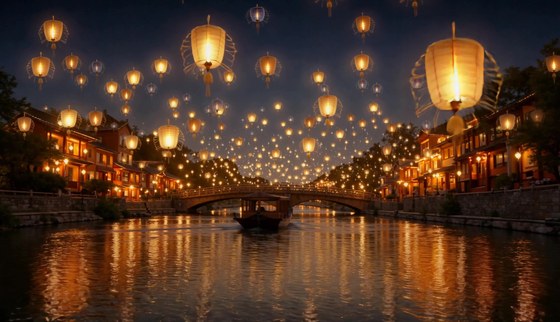}} & \raisebox{-.5\height}{\includegraphics[width=\qualw]{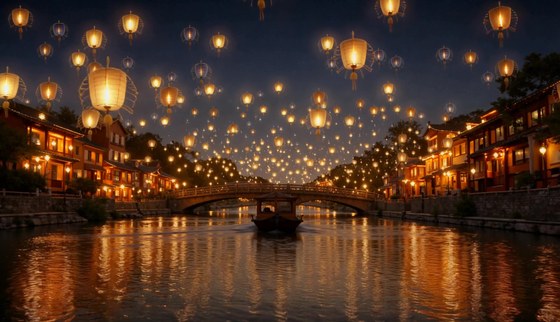}} & \raisebox{-.5\height}{\includegraphics[width=\qualw]{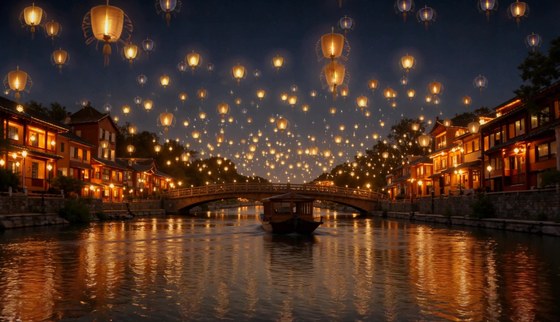}}\\[1pt]
\raisebox{-.5\height}{\rotatebox{90}{\tiny \sys}} & \raisebox{-.5\height}{\includegraphics[width=\qualw]{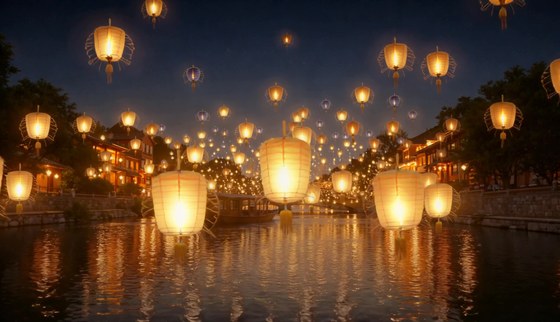}} & \raisebox{-.5\height}{\includegraphics[width=\qualw]{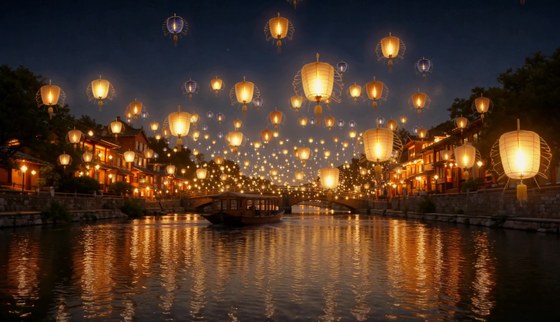}} & \raisebox{-.5\height}{\includegraphics[width=\qualw]{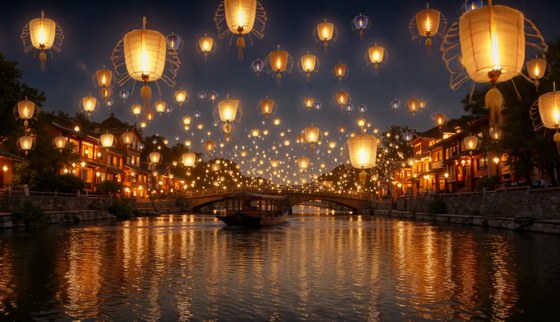}} & \raisebox{-.5\height}{\includegraphics[width=\qualw]{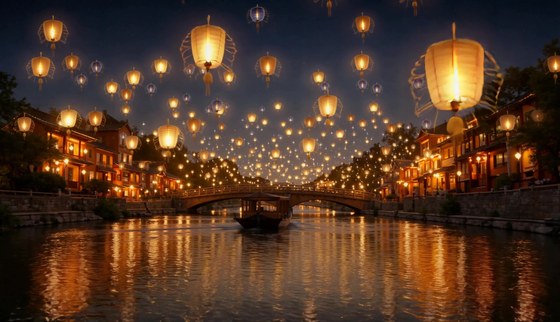}} & \raisebox{-.5\height}{\includegraphics[width=\qualw]{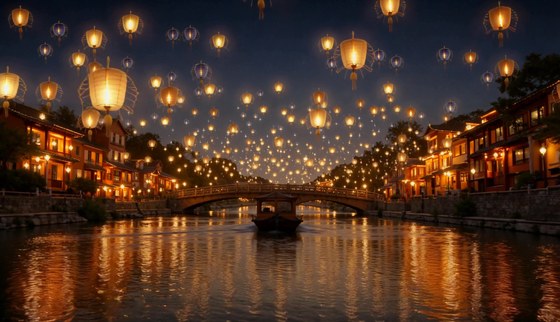}} & \raisebox{-.5\height}{\includegraphics[width=\qualw]{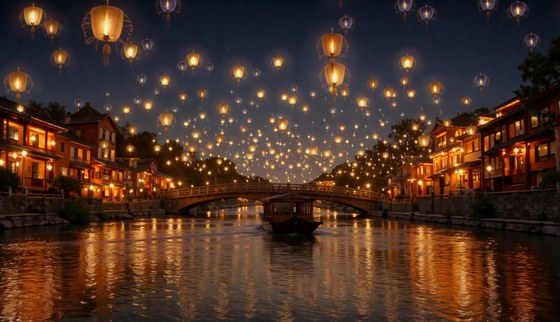}}\\
\end{tabular}}
\par\vspace{1pt}
\makebox[\textwidth][c]{\begin{tabular}{@{}c@{\hspace{1.5pt}}c@{\hspace{1pt}}c@{\hspace{1pt}}c@{\hspace{1pt}}c@{\hspace{1pt}}c@{\hspace{1pt}}c@{}}
\multicolumn{7}{@{}l@{}}{\makebox[\dimexpr6\qualw+5pt\relax][l]{\scriptsize\textbf{1440p}, 15\,s\quad\textit{swiss lake dawn}\hfill 318\,s\,$\rightarrow$\,179\,s (1.78\x), 38.0\,dB}}\\[1pt]
\multicolumn{7}{@{}p{\dimexpr6\qualw+5pt\relax}@{}}{\tiny\raggedright\textit{Prompt (seed 1018):} A tranquil alpine lake at dawn in the Swiss mountains. Mist drifts across the mirror-still water, a lone wooden rowboat floats near the shore, and the first sunlight turns the snowy peaks pink as the camera rises slowly over the lake. Ultra-detailed, calm, majestic. Gentle water lapping, distant cowbells and soft strings.}\\[2pt]
 & \tiny $t{=}1.5$\,s & \tiny $t{=}3.9$\,s & \tiny $t{=}6.3$\,s & \tiny $t{=}8.7$\,s & \tiny $t{=}11.1$\,s & \tiny $t{=}13.5$\,s\\[1pt]
\raisebox{-.5\height}{\rotatebox{90}{\tiny Dense}} & \raisebox{-.5\height}{\includegraphics[width=\qualw]{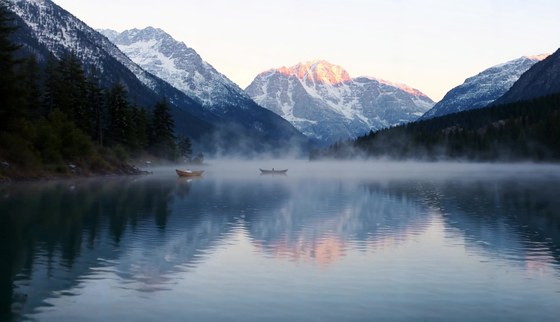}} & \raisebox{-.5\height}{\includegraphics[width=\qualw]{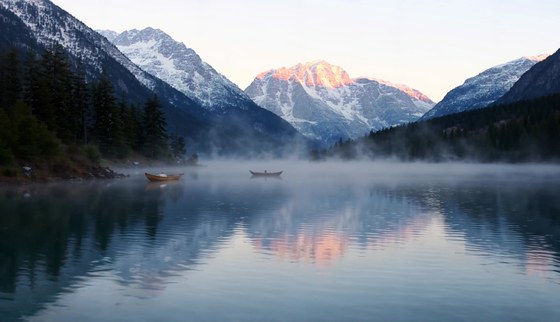}} & \raisebox{-.5\height}{\includegraphics[width=\qualw]{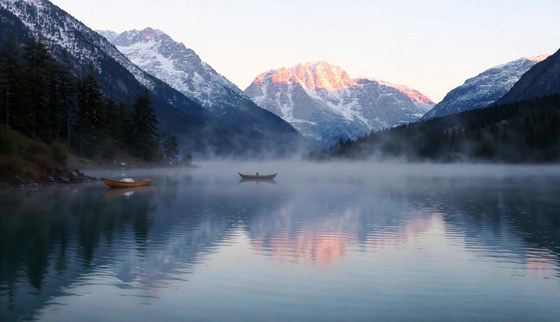}} & \raisebox{-.5\height}{\includegraphics[width=\qualw]{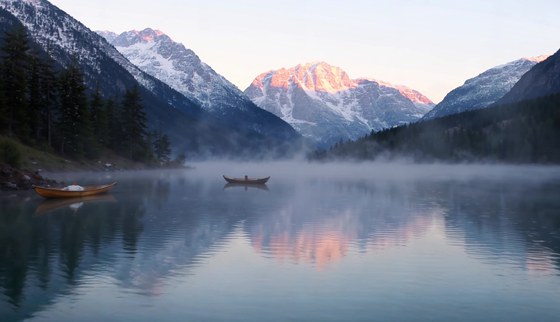}} & \raisebox{-.5\height}{\includegraphics[width=\qualw]{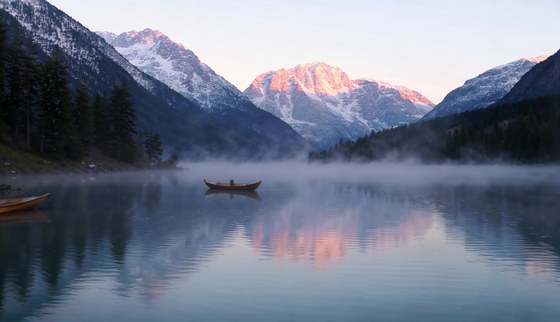}} & \raisebox{-.5\height}{\includegraphics[width=\qualw]{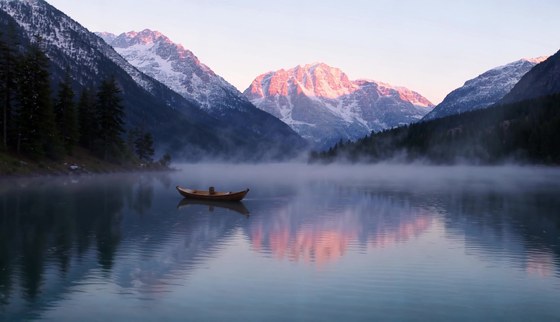}}\\[1pt]
\raisebox{-.5\height}{\rotatebox{90}{\tiny \sys}} & \raisebox{-.5\height}{\includegraphics[width=\qualw]{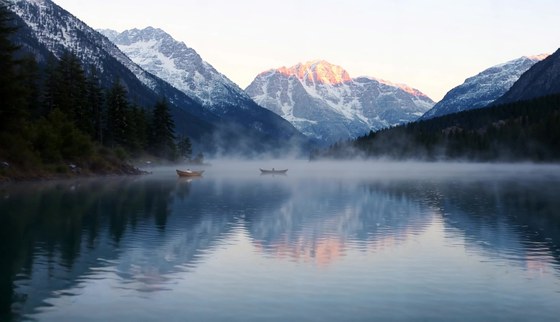}} & \raisebox{-.5\height}{\includegraphics[width=\qualw]{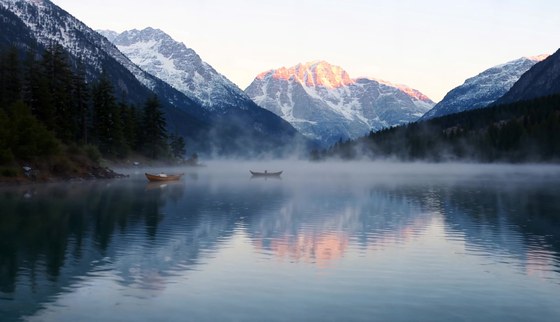}} & \raisebox{-.5\height}{\includegraphics[width=\qualw]{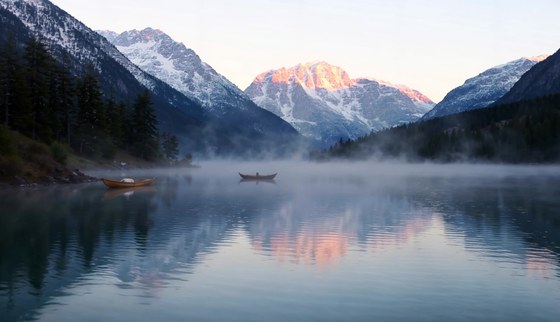}} & \raisebox{-.5\height}{\includegraphics[width=\qualw]{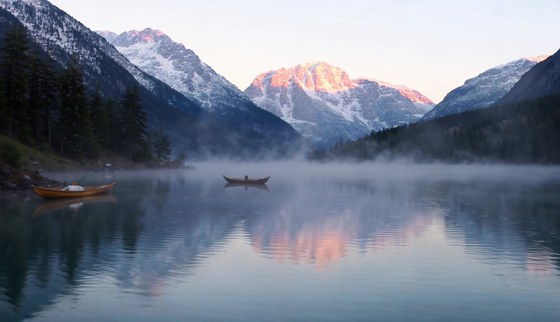}} & \raisebox{-.5\height}{\includegraphics[width=\qualw]{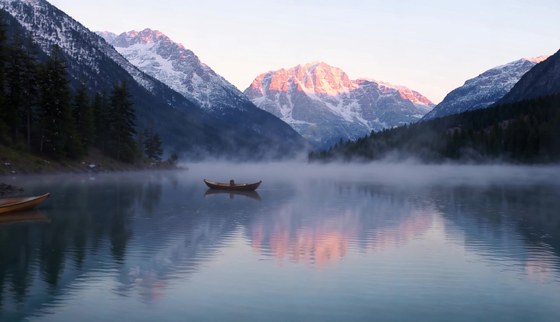}} & \raisebox{-.5\height}{\includegraphics[width=\qualw]{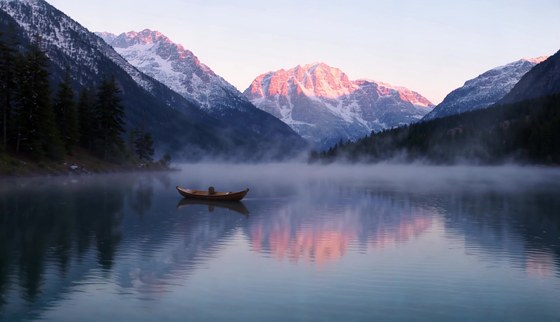}}\\
\end{tabular}}
\caption{LTX-2.5 with dense attention and with \sys, continued from \figref{fig:app-qual-ltx} (2/3).}
\label{fig:app-qual-ltx-2}
\end{figure*}

\begin{figure*}[p]
\centering
\ifx\qualw\undefined\newlength{\qualw}\fi
\setlength{\qualw}{\dimexpr(\textwidth - 12.5pt)*84/600\relax}%
\makebox[\textwidth][c]{\begin{tabular}{@{}c@{\hspace{1.5pt}}c@{\hspace{1pt}}c@{\hspace{1pt}}c@{\hspace{1pt}}c@{\hspace{1pt}}c@{\hspace{1pt}}c@{}}
\multicolumn{7}{@{}l@{}}{\makebox[\dimexpr6\qualw+5pt\relax][l]{\scriptsize\textbf{4K}, 7\,s\quad\textit{carrier jet takeoff}\hfill 344\,s\,$\rightarrow$\,188\,s (1.83\x), 40.4\,dB}}\\[1pt]
\multicolumn{7}{@{}p{\dimexpr6\qualw+5pt\relax}@{}}{\tiny\raggedright\textit{Prompt (seed 1003):} A fighter jet launches from an aircraft carrier at sunset; the afterburners blast blue-orange flames, steam billows from the catapult and the jet roars off the deck and climbs into the sky. Realistic, ultra-detailed, cinematic low angle. A deafening jet roar and wind.}\\[2pt]
 & \tiny $t{=}0.7$\,s & \tiny $t{=}1.8$\,s & \tiny $t{=}3.0$\,s & \tiny $t{=}4.0$\,s & \tiny $t{=}5.2$\,s & \tiny $t{=}6.3$\,s\\[1pt]
\raisebox{-.5\height}{\rotatebox{90}{\tiny Dense}} & \raisebox{-.5\height}{\includegraphics[width=\qualw]{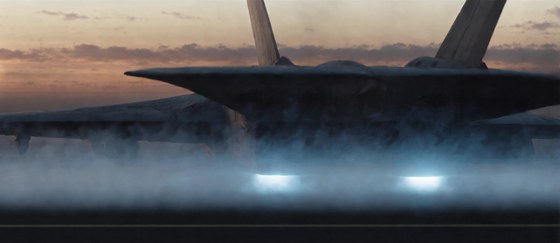}} & \raisebox{-.5\height}{\includegraphics[width=\qualw]{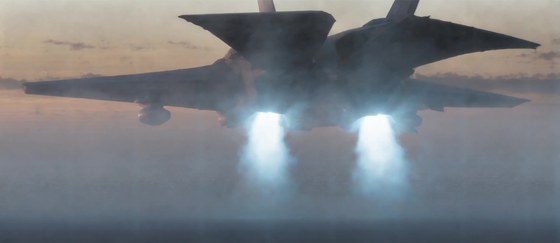}} & \raisebox{-.5\height}{\includegraphics[width=\qualw]{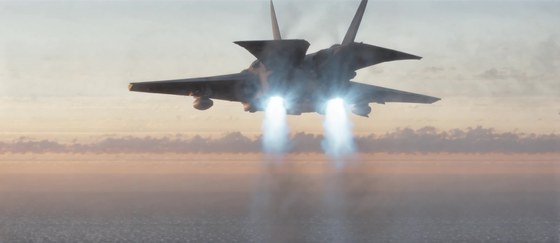}} & \raisebox{-.5\height}{\includegraphics[width=\qualw]{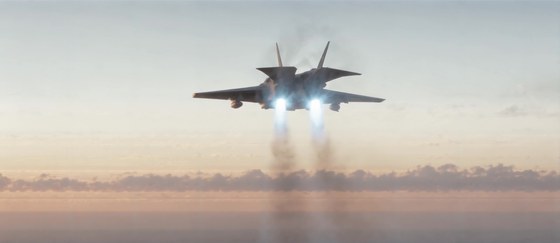}} & \raisebox{-.5\height}{\includegraphics[width=\qualw]{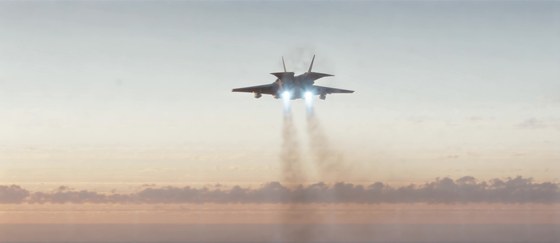}} & \raisebox{-.5\height}{\includegraphics[width=\qualw]{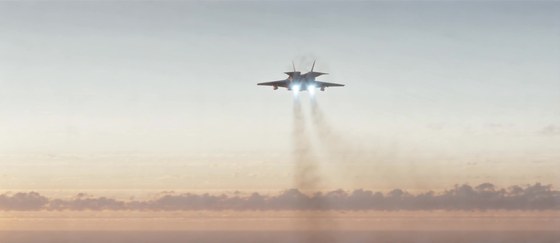}}\\[1pt]
\raisebox{-.5\height}{\rotatebox{90}{\tiny \sys}} & \raisebox{-.5\height}{\includegraphics[width=\qualw]{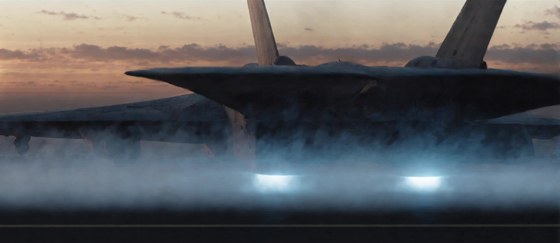}} & \raisebox{-.5\height}{\includegraphics[width=\qualw]{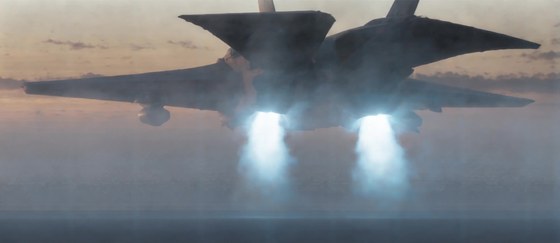}} & \raisebox{-.5\height}{\includegraphics[width=\qualw]{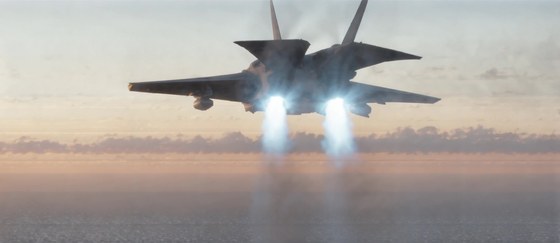}} & \raisebox{-.5\height}{\includegraphics[width=\qualw]{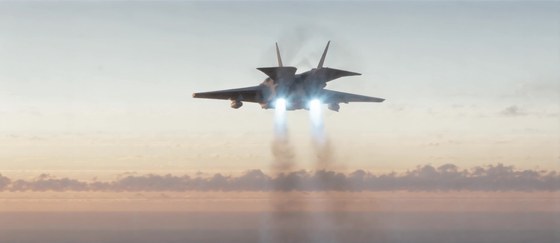}} & \raisebox{-.5\height}{\includegraphics[width=\qualw]{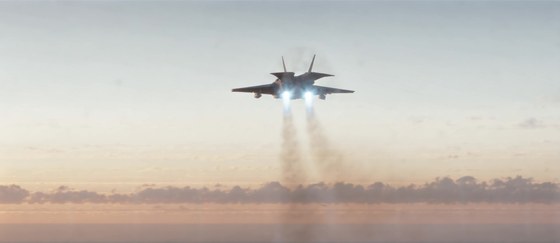}} & \raisebox{-.5\height}{\includegraphics[width=\qualw]{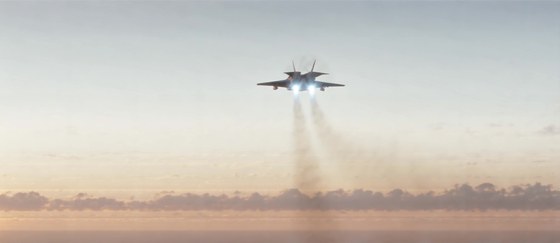}}\\
\end{tabular}}
\par\vspace{1pt}
\makebox[\textwidth][c]{\begin{tabular}{@{}c@{\hspace{1.5pt}}c@{\hspace{1pt}}c@{\hspace{1pt}}c@{\hspace{1pt}}c@{\hspace{1pt}}c@{\hspace{1pt}}c@{}}
\multicolumn{7}{@{}l@{}}{\makebox[\dimexpr6\qualw+5pt\relax][l]{\scriptsize\textbf{4K}, 7\,s\quad\textit{dragon eye open}\hfill 345\,s\,$\rightarrow$\,190\,s (1.82\x), 34.6\,dB}}\\[1pt]
\multicolumn{7}{@{}p{\dimexpr6\qualw+5pt\relax}@{}}{\tiny\raggedright\textit{Prompt (seed 1005):} An extreme close-up of an ancient dragon's eye slowly opening; the reptilian pupil narrows, and flames reflect in its molten gold iris as smoke curls across its obsidian scales. Hyper-detailed fantasy, dramatic lighting. A deep rumbling growl and crackling fire.}\\[2pt]
 & \tiny $t{=}0.7$\,s & \tiny $t{=}1.8$\,s & \tiny $t{=}3.0$\,s & \tiny $t{=}4.0$\,s & \tiny $t{=}5.2$\,s & \tiny $t{=}6.3$\,s\\[1pt]
\raisebox{-.5\height}{\rotatebox{90}{\tiny Dense}} & \raisebox{-.5\height}{\includegraphics[width=\qualw]{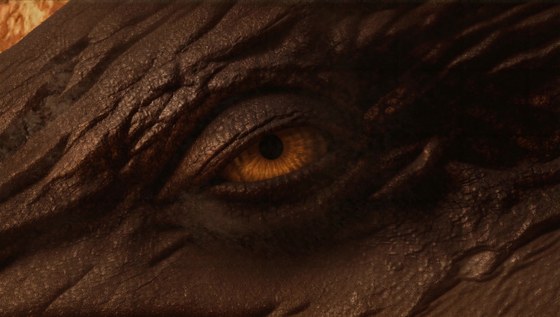}} & \raisebox{-.5\height}{\includegraphics[width=\qualw]{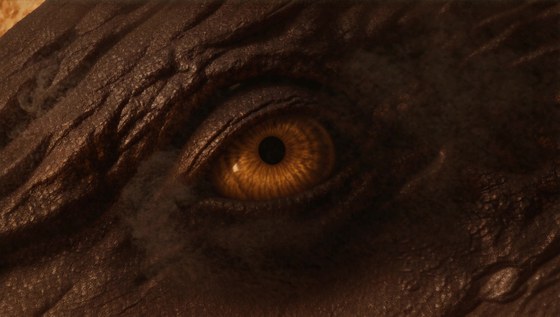}} & \raisebox{-.5\height}{\includegraphics[width=\qualw]{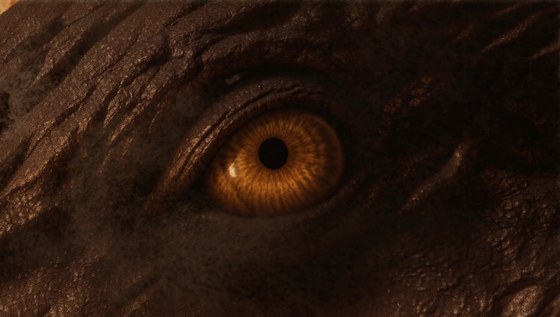}} & \raisebox{-.5\height}{\includegraphics[width=\qualw]{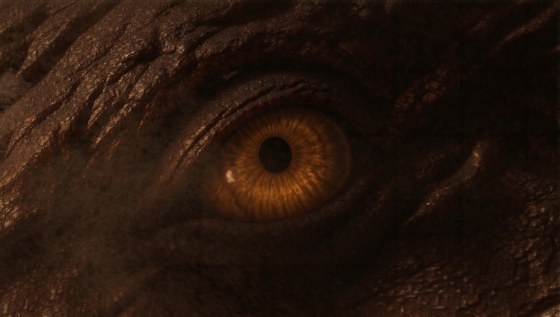}} & \raisebox{-.5\height}{\includegraphics[width=\qualw]{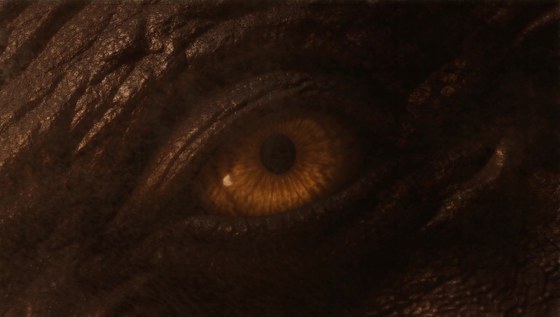}} & \raisebox{-.5\height}{\includegraphics[width=\qualw]{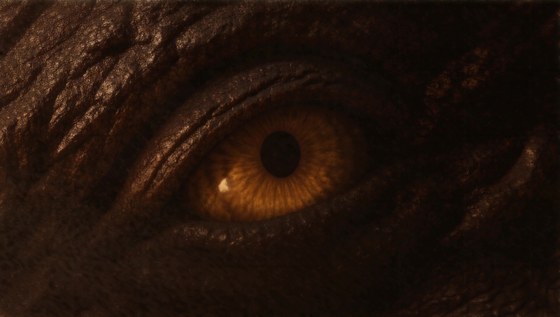}}\\[1pt]
\raisebox{-.5\height}{\rotatebox{90}{\tiny \sys}} & \raisebox{-.5\height}{\includegraphics[width=\qualw]{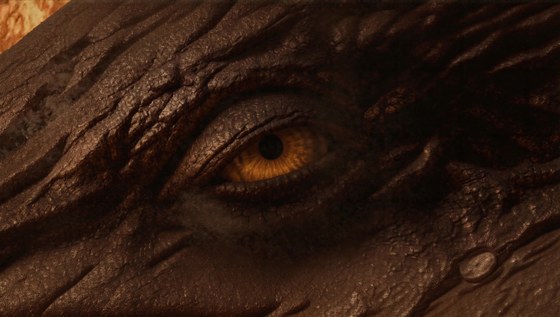}} & \raisebox{-.5\height}{\includegraphics[width=\qualw]{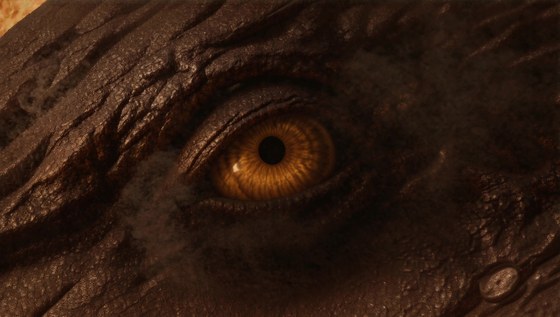}} & \raisebox{-.5\height}{\includegraphics[width=\qualw]{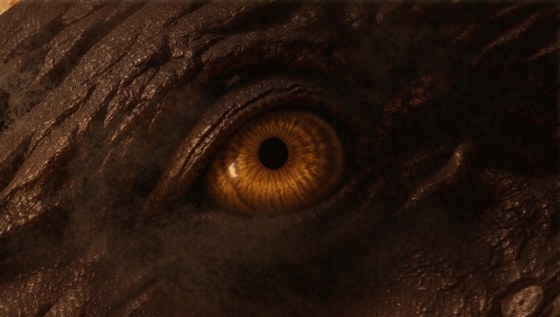}} & \raisebox{-.5\height}{\includegraphics[width=\qualw]{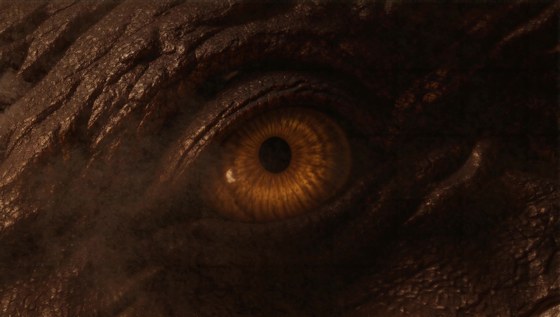}} & \raisebox{-.5\height}{\includegraphics[width=\qualw]{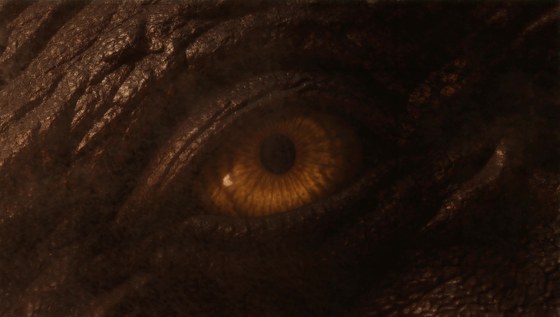}} & \raisebox{-.5\height}{\includegraphics[width=\qualw]{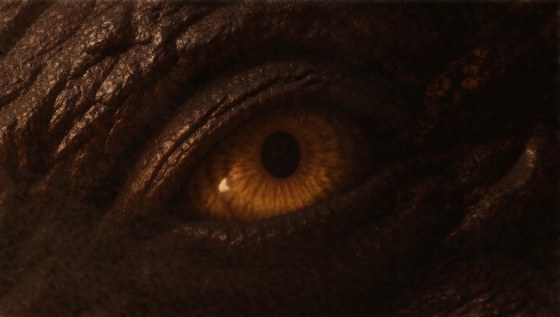}}\\
\end{tabular}}
\par\vspace{1pt}
\makebox[\textwidth][c]{\begin{tabular}{@{}c@{\hspace{1.5pt}}c@{\hspace{1pt}}c@{\hspace{1pt}}c@{\hspace{1pt}}c@{\hspace{1pt}}c@{\hspace{1pt}}c@{}}
\multicolumn{7}{@{}l@{}}{\makebox[\dimexpr6\qualw+5pt\relax][l]{\scriptsize\textbf{4K}, 7\,s\quad\textit{holo war room}\hfill 351\,s\,$\rightarrow$\,192\,s (1.83\x), 31.9\,dB}}\\[1pt]
\multicolumn{7}{@{}p{\dimexpr6\qualw+5pt\relax}@{}}{\tiny\raggedright\textit{Prompt (seed 1010):} A futuristic command center: a giant holographic 3D map of a city floats above a table, red threat markers pulse and spread, and an officer swipes through glowing data panels as alarms flash. Sleek sci-fi, blue and red glow, ultra-detailed. Beeping alerts and a tense low drone.}\\[2pt]
 & \tiny $t{=}0.7$\,s & \tiny $t{=}1.8$\,s & \tiny $t{=}3.0$\,s & \tiny $t{=}4.0$\,s & \tiny $t{=}5.2$\,s & \tiny $t{=}6.3$\,s\\[1pt]
\raisebox{-.5\height}{\rotatebox{90}{\tiny Dense}} & \raisebox{-.5\height}{\includegraphics[width=\qualw]{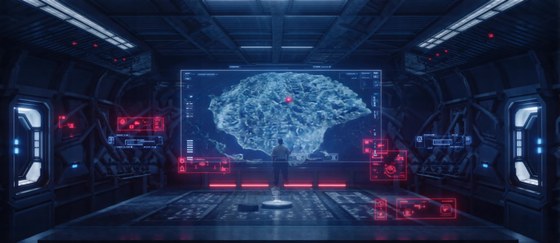}} & \raisebox{-.5\height}{\includegraphics[width=\qualw]{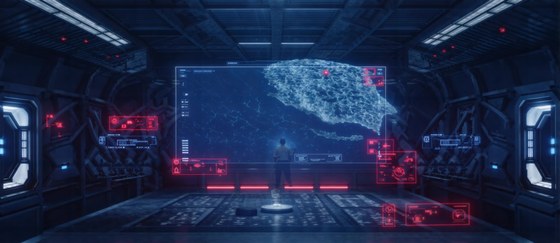}} & \raisebox{-.5\height}{\includegraphics[width=\qualw]{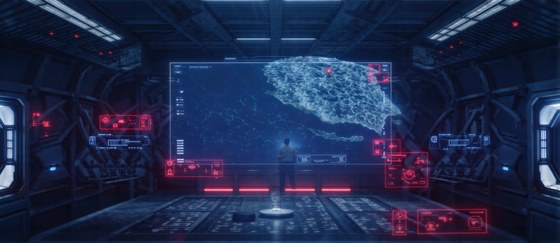}} & \raisebox{-.5\height}{\includegraphics[width=\qualw]{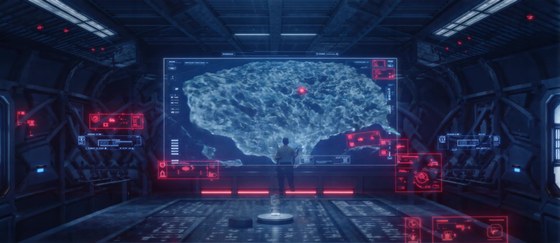}} & \raisebox{-.5\height}{\includegraphics[width=\qualw]{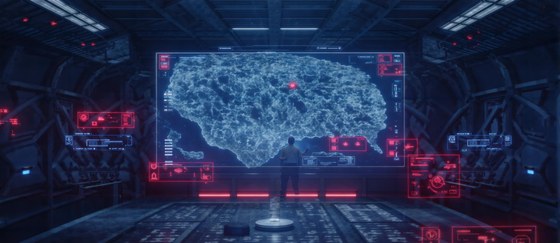}} & \raisebox{-.5\height}{\includegraphics[width=\qualw]{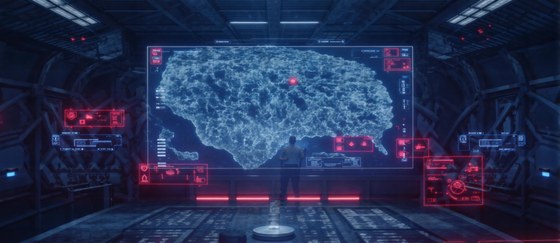}}\\[1pt]
\raisebox{-.5\height}{\rotatebox{90}{\tiny \sys}} & \raisebox{-.5\height}{\includegraphics[width=\qualw]{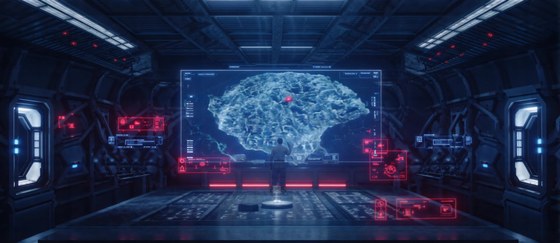}} & \raisebox{-.5\height}{\includegraphics[width=\qualw]{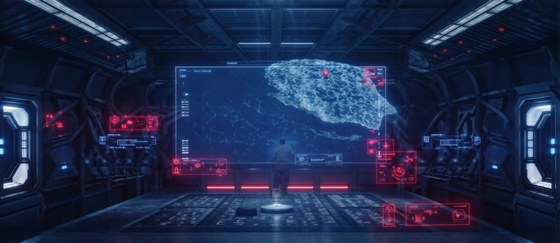}} & \raisebox{-.5\height}{\includegraphics[width=\qualw]{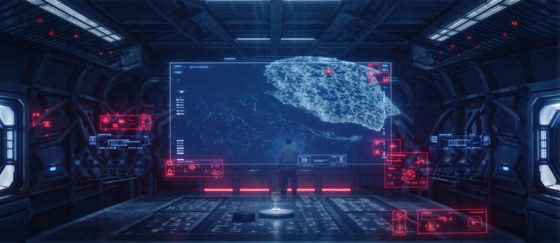}} & \raisebox{-.5\height}{\includegraphics[width=\qualw]{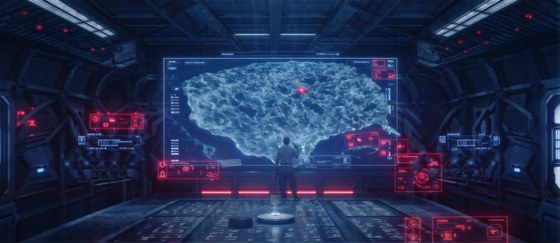}} & \raisebox{-.5\height}{\includegraphics[width=\qualw]{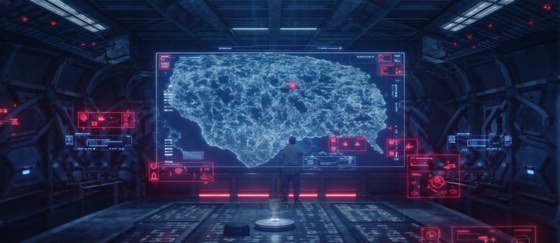}} & \raisebox{-.5\height}{\includegraphics[width=\qualw]{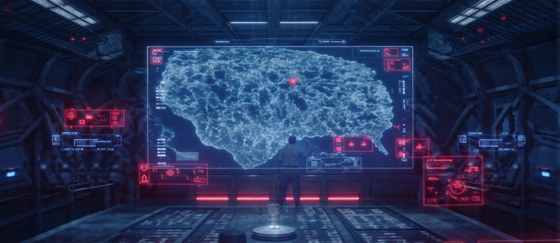}}\\
\end{tabular}}
\par\vspace{1pt}
\makebox[\textwidth][c]{\begin{tabular}{@{}c@{\hspace{1.5pt}}c@{\hspace{1pt}}c@{\hspace{1pt}}c@{\hspace{1pt}}c@{\hspace{1pt}}c@{\hspace{1pt}}c@{}}
\multicolumn{7}{@{}l@{}}{\makebox[\dimexpr6\qualw+5pt\relax][l]{\scriptsize\textbf{4K}, 7\,s\quad\textit{lightning skyscraper}\hfill 349\,s\,$\rightarrow$\,190\,s (1.84\x), 35.5\,dB}}\\[1pt]
\multicolumn{7}{@{}p{\dimexpr6\qualw+5pt\relax}@{}}{\tiny\raggedright\textit{Prompt (seed 1012):} A massive bolt of lightning strikes the spire of a glass skyscraper during a violent night storm, branching across the dark clouds and lighting up the entire city below for an instant. Ultra-detailed, dramatic, cinematic. A deafening thunderclap and pouring rain.}\\[2pt]
 & \tiny $t{=}0.7$\,s & \tiny $t{=}1.8$\,s & \tiny $t{=}3.0$\,s & \tiny $t{=}4.0$\,s & \tiny $t{=}5.2$\,s & \tiny $t{=}6.3$\,s\\[1pt]
\raisebox{-.5\height}{\rotatebox{90}{\tiny Dense}} & \raisebox{-.5\height}{\includegraphics[width=\qualw]{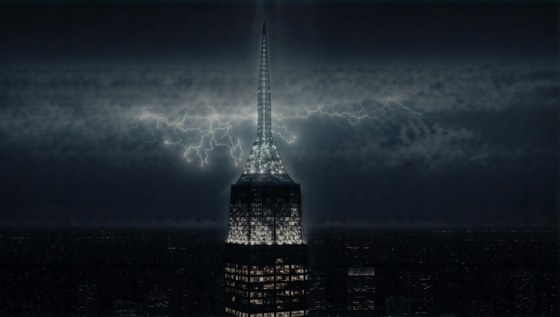}} & \raisebox{-.5\height}{\includegraphics[width=\qualw]{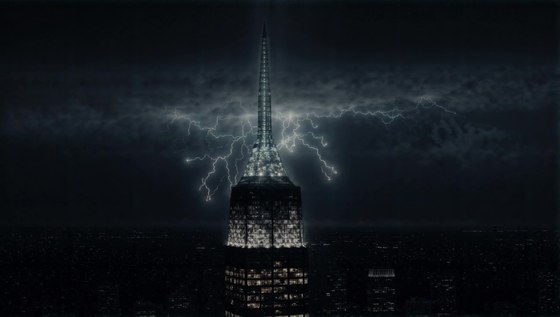}} & \raisebox{-.5\height}{\includegraphics[width=\qualw]{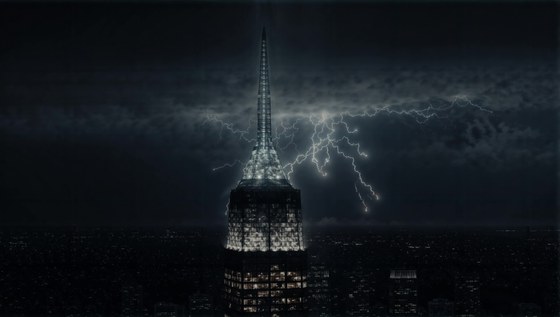}} & \raisebox{-.5\height}{\includegraphics[width=\qualw]{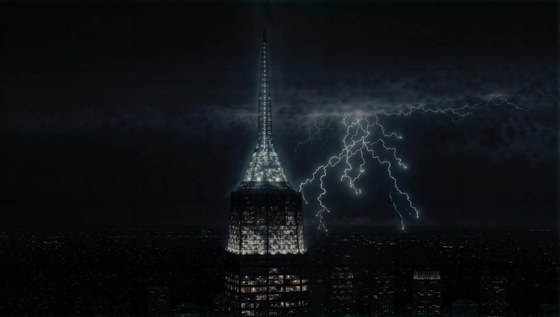}} & \raisebox{-.5\height}{\includegraphics[width=\qualw]{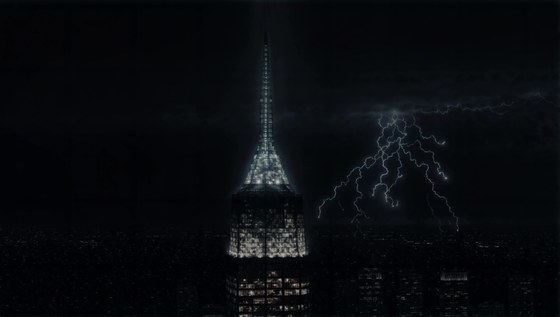}} & \raisebox{-.5\height}{\includegraphics[width=\qualw]{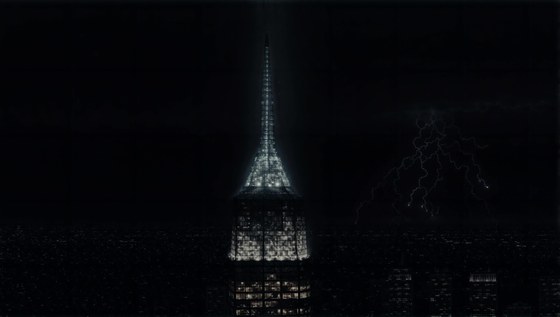}}\\[1pt]
\raisebox{-.5\height}{\rotatebox{90}{\tiny \sys}} & \raisebox{-.5\height}{\includegraphics[width=\qualw]{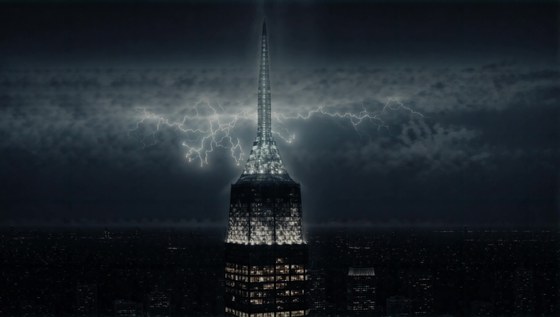}} & \raisebox{-.5\height}{\includegraphics[width=\qualw]{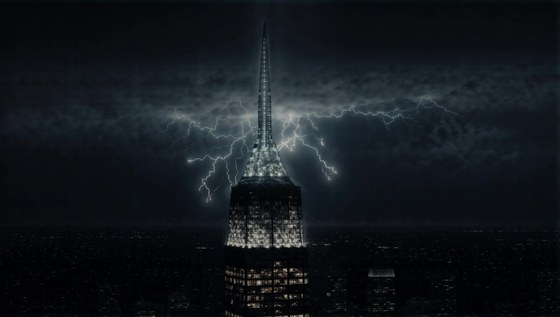}} & \raisebox{-.5\height}{\includegraphics[width=\qualw]{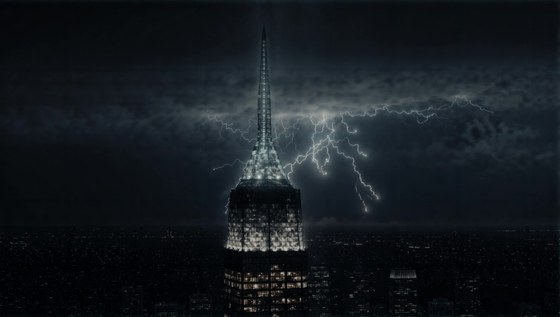}} & \raisebox{-.5\height}{\includegraphics[width=\qualw]{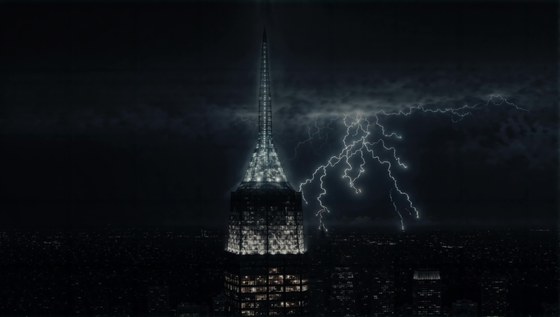}} & \raisebox{-.5\height}{\includegraphics[width=\qualw]{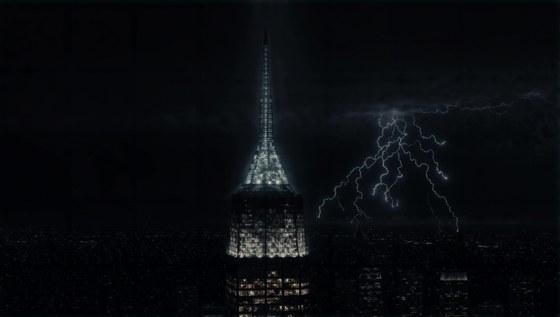}} & \raisebox{-.5\height}{\includegraphics[width=\qualw]{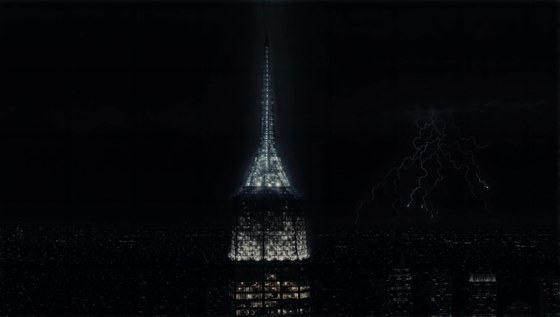}}\\
\end{tabular}}
\par\vspace{1pt}
\makebox[\textwidth][c]{\begin{tabular}{@{}c@{\hspace{1.5pt}}c@{\hspace{1pt}}c@{\hspace{1pt}}c@{\hspace{1pt}}c@{\hspace{1pt}}c@{\hspace{1pt}}c@{}}
\multicolumn{7}{@{}l@{}}{\makebox[\dimexpr6\qualw+5pt\relax][l]{\scriptsize\textbf{4K}, 7\,s\quad\textit{ice cave light}\hfill 349\,s\,$\rightarrow$\,191\,s (1.82\x), 35.5\,dB}}\\[1pt]
\multicolumn{7}{@{}p{\dimexpr6\qualw+5pt\relax}@{}}{\tiny\raggedright\textit{Prompt (seed 1016):} Inside a glowing blue glacier ice cave, sunlight filters through the translucent ice ceiling, casting shifting turquoise light on the frozen walls as tiny droplets fall and sparkle. Slow, calm camera glide, ultra-detailed, serene. Soft dripping water and a hushed ambient tone.}\\[2pt]
 & \tiny $t{=}0.7$\,s & \tiny $t{=}1.8$\,s & \tiny $t{=}3.0$\,s & \tiny $t{=}4.0$\,s & \tiny $t{=}5.2$\,s & \tiny $t{=}6.3$\,s\\[1pt]
\raisebox{-.5\height}{\rotatebox{90}{\tiny Dense}} & \raisebox{-.5\height}{\includegraphics[width=\qualw]{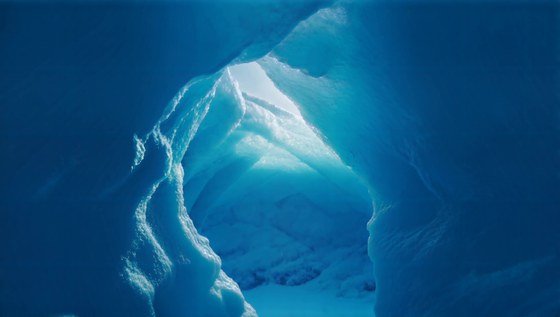}} & \raisebox{-.5\height}{\includegraphics[width=\qualw]{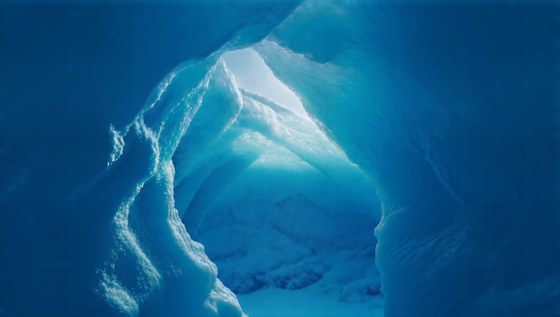}} & \raisebox{-.5\height}{\includegraphics[width=\qualw]{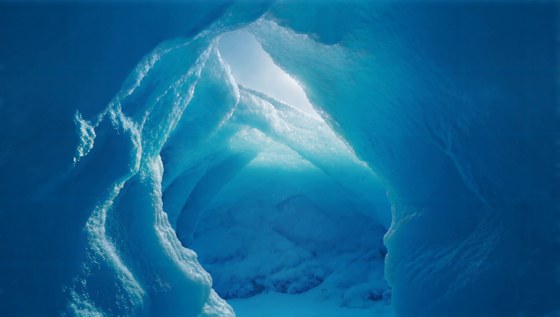}} & \raisebox{-.5\height}{\includegraphics[width=\qualw]{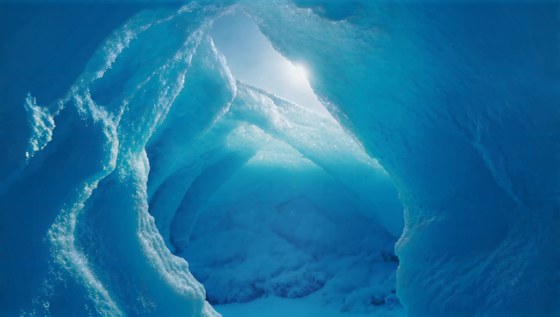}} & \raisebox{-.5\height}{\includegraphics[width=\qualw]{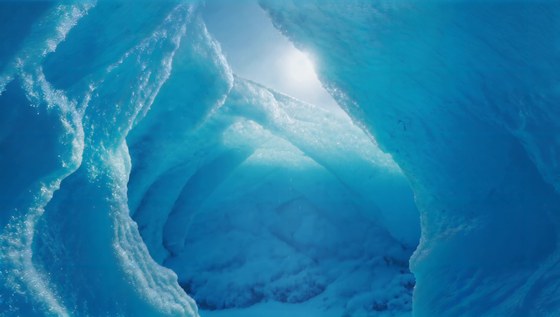}} & \raisebox{-.5\height}{\includegraphics[width=\qualw]{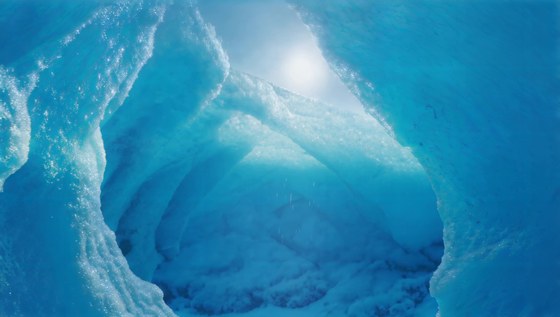}}\\[1pt]
\raisebox{-.5\height}{\rotatebox{90}{\tiny \sys}} & \raisebox{-.5\height}{\includegraphics[width=\qualw]{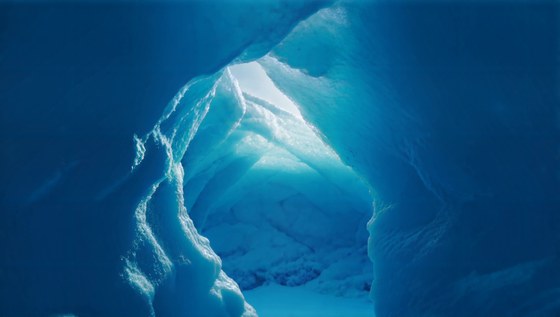}} & \raisebox{-.5\height}{\includegraphics[width=\qualw]{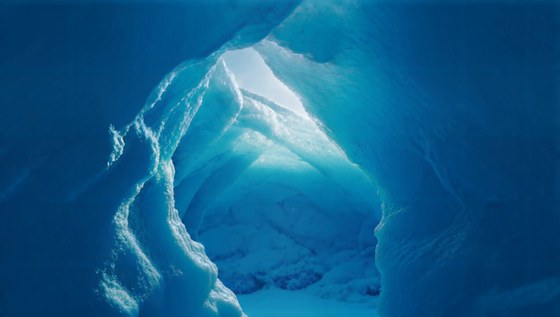}} & \raisebox{-.5\height}{\includegraphics[width=\qualw]{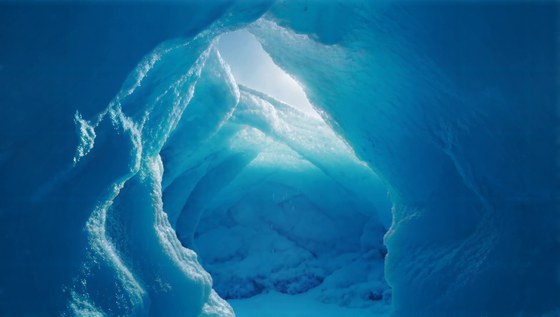}} & \raisebox{-.5\height}{\includegraphics[width=\qualw]{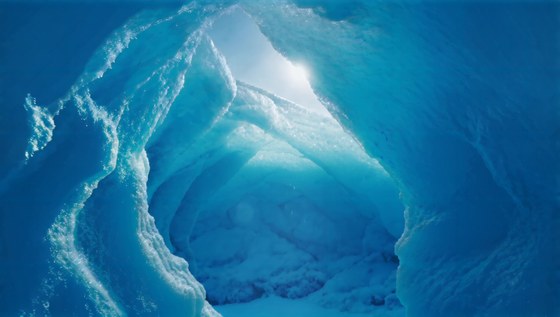}} & \raisebox{-.5\height}{\includegraphics[width=\qualw]{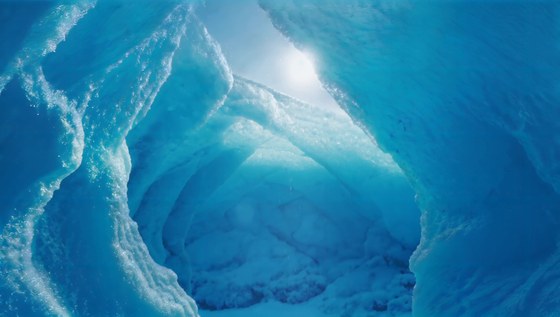}} & \raisebox{-.5\height}{\includegraphics[width=\qualw]{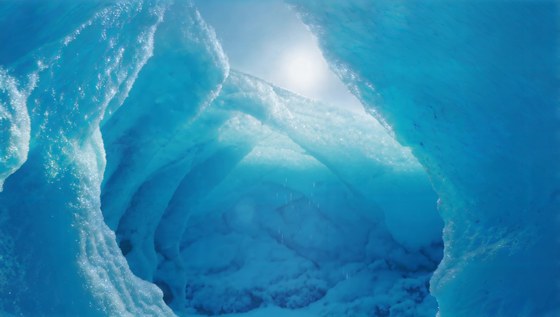}}\\
\end{tabular}}
\caption{LTX-2.5 with dense attention and with \sys, continued from \figref{fig:app-qual-ltx} (3/3).}
\label{fig:app-qual-ltx-3}
\end{figure*}

\fi

\end{document}